\documentclass[10pt,twocolumn,letterpaper]{article}

\PassOptionsToPackage{table}{xcolor}
\usepackage[pagenumbers]{cvpr}
\usepackage{iftex}
\ifPDFTeX
\else
  \usepackage{fontspec}
\fi

\usepackage{tikz}
\usepackage{enumitem}

\usepackage{float,array,longtable,placeins}
\DeclareFontShape{OMX}{cmex}{m}{n}{<-> cmex10}{}
\usepackage{xspace}
\newcommand{\method}{TailorGS\xspace}

\newcommand{\covw}{\ensuremath{w_i}}                  
\newcommand{\bsupp}{\ensuremath{\mu_i}}               
\newcommand{\capsupp}{\ensuremath{S}}                 
\newcommand{\bref}{\ensuremath{B_{\mathrm{ref}}}}     
\newcommand{\util}{\ensuremath{\widehat{U}_g}}        
\newcommand{\lamstar}{\ensuremath{\lambda^{*}}}       

\newcommand{\krho}{\ensuremath{\rho}}                 
\newcommand{\numg}{\ensuremath{N_{\mathrm{G}}}}       
\newcommand{\npre}{\ensuremath{N_{\mathrm{pre}}}}     
\newcommand{\pix}{\ensuremath{X}}                     
\newcommand{\tcap}{\ensuremath{T_{\mathrm{cap}}}}     

\usepackage{booktabs}
\usepackage{multirow}

\definecolor{linkblue}{rgb}{0.21,0.49,0.74}
\usepackage[pagebackref,breaklinks,colorlinks,allcolors=linkblue,pdfusetitle]{hyperref}
\usepackage[capitalize]{cleveref}
\def\confName{CVPR}
\def\confYear{2027}

\title{TailorGS: Made-to-Measure Density Control for 3D Gaussian Splatting}

\author{Afif Boudaoud \qquad Jiayi Liu \qquad Alexandru Calotoiu \qquad Torsten Hoefler\\
ETH Zurich}
\hypersetup{pdfauthor={Afif Boudaoud, Jiayi Liu, Alexandru Calotoiu, Torsten Hoefler}}

\begin{document}
\twocolumn[{%
    \renewcommand\twocolumn[1][]{#1}%
    \maketitle
    \vspace{-14pt}%
    \begingroup
    \centering
    \includegraphics[width=\linewidth]{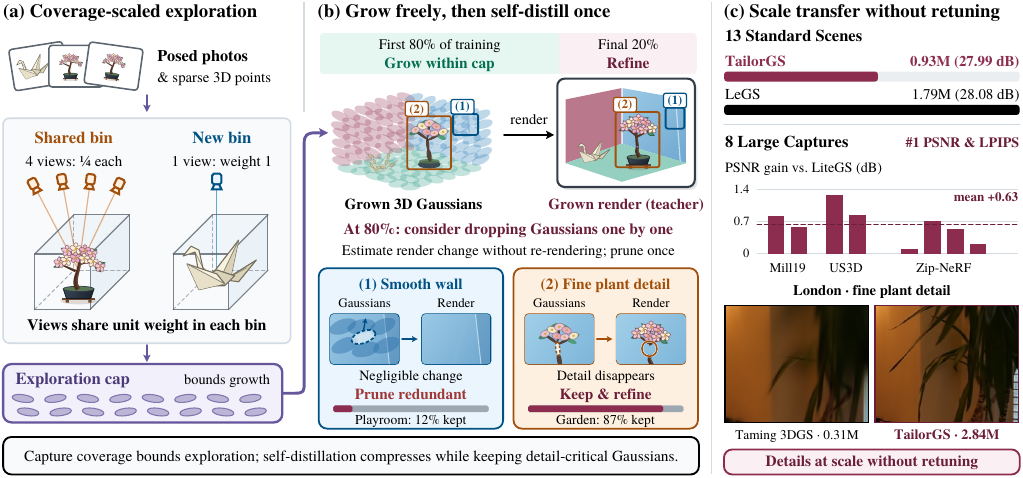}
    \par\vspace{-4pt}
    \captionsetup{hypcap=false}
    \captionof{figure}{\textbf{\method{} overview.} \textbf{(a)}~View coverage weights and image resolution set an exploration cap before training. \textbf{(b)}~After growing within this cap for $80\%$ of training, coverage-weighted self-distillation prunes redundant Gaussians; the retained primitives are refined.
    \textbf{(c)}~The same controller stays compact on standard scenes and scales to large captures without retuning.
    }
    \captionsetup{hypcap=true}
    \label{fig:pipeline}
\endgroup

}]

\begin{abstract}
3D Gaussian Splatting reconstructs a scene from a set of posed photographs called the capture; Gaussian count affects reconstruction quality, storage, and rendering cost.
How to select this number automatically across capture scales remains unresolved: density control techniques effective on standard benchmarks can leave larger captures with too few Gaussians to reconstruct fine details.
We observe that the \emph{surface} to represent, given by capture extent and resolution, is known before training, whereas the capacity needed to reconstruct its \emph{content} depends strongly on the 3D optimization trajectory and becomes apparent during reconstruction.
We introduce \method{}, which unifies coverage-scaled exploration and coverage-weighted self-distillation under a non-redundant 3D scene coverage measure.
Before training, \method{} groups observed surface points into coarse voxels and shares each voxel's weight among views observing it from the same side.
Combined with image resolution, these weights set a generous exploration cap that limits budget inflation from repeated observations.
During the first $80\%$ of training, \method{} grows toward this cap.
The resulting model serves as teacher for a single closed-form self-distillation step, balancing fidelity to its renderings against Gaussian count under the same coverage measure.
The retained primitives are refined over the remaining $20\%$ of training.
On 13 standard scenes, \method{} matches the mean PSNR of the best-performing evaluated baseline, LeGS, with $48\%$ fewer Gaussians.
On eight large captures, the same configuration automatically scales to larger models when necessary, achieving the highest mean PSNR across 13 evaluated baselines: $0.63$\,dB above the runner-up by allocating $2.8\times$ as many Gaussians.
\end{abstract}

\section{Introduction}
\label{sec:intro}

3D Gaussian Splatting~\citep{kerbl3Dgaussians} represents a scene as Gaussians learned from photographs.
During training, \emph{density control} determines the number of Gaussians through \emph{densification} (adding Gaussians) and \emph{pruning} (removing them). Selecting the Gaussian count requires balancing reconstruction quality against storage across a broad range of model sizes. For example, on one large capture, our evaluated models range from $87$K to $7.4$M Gaussians and span $5$\,dB in PSNR.

We group existing density controllers by how they determine Gaussian count:
\emph{Fixed-constant methods} use preset thresholds or pruning fractions, making model size sensitive to gradient distributions that shift with resolution and content~\citep{kerbl3Dgaussians,hanson2025speedysplat}.
\emph{Preset-count methods} provide explicit targets or caps, but suitable values must be chosen per capture and inherited gradient growth thresholds can prevent these targets from being reached~\citep{tamings3dgs,kheradmand2024mcmc,litegs}.
\emph{Automatic-budget methods} let model size emerge from learned decisions, budget estimates, or local rules, yet their final counts still depend on regularization weights, fixed fractions, or local thresholds that are not conditioned on capture scale~\citep{chen2025dashgaussian,lee2024compact,ning2026legs}.
On larger or higher-resolution captures, these unconditioned controls break down, leaving requested budgets largely unfilled or even shrinking models below sparse initialization, as our transfer experiments show (\cref{sec:exp-generalization}).
The challenge is to balance reconstruction quality and model size across capture scales without the repeated full-training runs that manual retuning requires.

Two kinds of information determine how many Gaussians a capture requires.
The first is \emph{surface}: the capture's extent and resolution indicate how much of the scene must be represented and at what level of detail.
These are known before training.
The second is \emph{content complexity}: equally sized captures may depict smooth surfaces or intricate vegetation, but the capacity needed to reconstruct them depends strongly on the 3D optimization trajectory.
As the reconstruction develops, each primitive's rendering contribution across views indicates whether that capacity is needed to preserve scene detail.

Guided by five capacity-scaling principles (\cref{tab:scale-principles}), we introduce \method{}, which unifies \emph{coverage-scaled exploration} with \emph{one-shot coverage-weighted self-distillation} under a shared measure of the observed 3D surface.
To limit inflation of the exploration cap from repeated observations of the same surface, \method{} divides the sparse reconstruction into coarse voxels before training and shares each voxel's weight among views observing it from the same side. Each view’s coverage weight $\covw \in (0, 1]$ is the average of its shares. These weights account for repeated observations; the exploration cap must also reflect the resolution at which the surface is observed.
Smooth surfaces can span many pixels with relatively few Gaussians, while boundaries and fine structures often require denser representations. This motivates combining these weights with linear image scale to set the exploration cap (\cref{fig:pipeline}a).
During the first $80\%$ of training, \method{} grows toward this cap.
The resulting model serves as the teacher for a single closed-form self-distillation step that balances fidelity to its renderings against Gaussian count. The retained primitives are refined over the remaining $20\%$ of training (\cref{fig:pipeline}b). If a different quality--size trade-off is desired, only self-distillation and refinement need to be rerun, without repeating the growth phase.

Our contributions are threefold:
\begin{itemize}
    \item \textbf{Principled 3D scene coverage.} We formalize five capacity-scaling principles covering view duplication, camera bursts, disjoint scenes, SfM point density, and image resolution, and show that the existing capacity signals we examine each violate at least two. We therefore construct a 3D scene coverage measure from training views alone and prove that it satisfies all five principles.

    \item \textbf{Scale-transferable density control.} \method{} uses the same measure for coverage-scaled exploration and a single self-distillation step that adapts model size to scene content without retuning. Calibrated once on standard scenes, our method transfers unchanged to large captures. Beyond automatically adapting capacity to capture scale, our formulation offers a simple way to adjust the quality--size trade-off without retraining the model from scratch.

    \item \textbf{An empirical survey across capture scales.} We organize density controllers by how they set and adapt model size, and rerun 13 released implementations under a common evaluation protocol on 13 standard scenes and eight large captures. We examine how quality--size trade-offs shift from standard to large captures, giving practitioners a common basis for choosing a controller suited to their capture scale and quality--size priorities.
\end{itemize}

On the 13 standard scenes, \method{} matches LeGS's~\citep{ning2026legs} mean PSNR with roughly half as many Gaussians and lies on the quality--size Pareto front for every dataset (\cref{sec:exp-automatic}).
Transferred unchanged to eight large captures, it achieves the highest mean PSNR and lowest mean learned perceptual image patch similarity (LPIPS)~\citep{zhang2018lpips} among evaluated baselines.
Compared with the perceptual-quality runner-up, vanilla 3DGS~\citep{kerbl3Dgaussians}, \method{} uses $1.27\times$ as many Gaussians, achieves $0.70$\,dB higher mean PSNR, and lowers LPIPS on seven of eight captures. At neither scale does an evaluated baseline achieve lower mean LPIPS with fewer Gaussians.
Ablations confirm both components' contributions: using a fixed exploration cap reduces large-capture PSNR by at least $0.46$\,dB, while disabling self-distillation inflates the Gaussian count by at least $2\times$ for negligible PSNR gains.

\section{Density control across capture scales}
\label{sec:related}

We examine density control across a broad range of scene extents, image resolutions, and content, from individual objects and rooms to aerial sites and multi-room interiors, with 125–3,019 photographs per capture.
We use 13 \emph{standard scenes} (Std-13) commonly used to develop and evaluate 3DGS methods~\citep{kerbl3Dgaussians,hanson2025speedysplat}: nine from Mip-NeRF~360~\citep{barron2022mipnerf360}, \emph{truck} / \emph{train} from Tanks~\&~Temples~\citep{knapitsch2017tanks}, and \emph{drjohnson} / \emph{playroom} from Deep Blending~\citep{hedman2018deepblending}.
We also study scaling to eight \emph{large captures} (Large-8): \emph{rubble} / \emph{building} from Mill19~\citep{turki2022meganerf}, \emph{residence} / \emph{sci-art} from UrbanScene3D~\citep{lin2022urbanscene3d}, and \emph{alameda} / \emph{london} / \emph{nyc} / \emph{berlin} from Zip-NeRF~\citep{barron2023zipnerf}.

We group density controllers into three classes: fixed constant, preset count, and automatic budget.
In this section, we describe how each class determines model size and examine the limitations of transferring its controls across capture scales to identify a gap in existing density control approaches that our proposed method addresses.
Additional related work appears in Supplementary~\S G.

\subsection{Fixed constant controllers}
\label{sec:related-fixedconstant}

Vanilla 3DGS densifies Gaussians whose average per-view positional gradient magnitude exceeds a fixed \emph{gradient growth threshold} ($2\times10^{-4}$ by default).
These gradients measure how moving projected Gaussian centers changes the reconstruction loss.
A fixed threshold does not directly specify the final Gaussian count.
More importantly, because growth is tied to the distribution of positional gradients, differences in this distribution can produce different growth rates across captures without indicating whether the resulting models have sufficient capacity.

\begin{figure}[t]
    \centering
    \includegraphics[width=\linewidth]{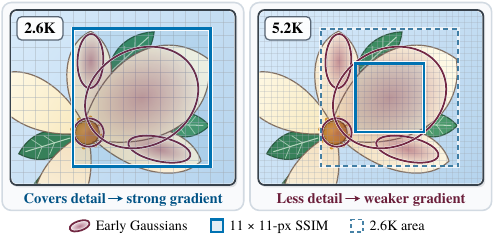}
    \caption{\textbf{SSIM at two resolutions.} A fixed $11 \times 11$-pixel window covers less of the scene at higher resolution.}
    \label{fig:ssim-resolution}
\end{figure}

Mip-NeRF~360's Garden scene illustrates this limitation.
Increasing its input resolution from $2.6$K to $5.2$K reduces vanilla 3DGS's final count from $5.18$M to $4.18$M Gaussians under the default threshold.
This is because the positional gradients are computed from $L_1$ and SSIM~\citep{wang2004ssim}, an image-similarity measure used during training, which compares local regions at a fixed scale in pixels.
At higher resolution, these comparisons cover a smaller part of the scene (\cref{fig:ssim-resolution}).
In Garden, this weakens early gradients, leaving fewer Gaussians above the gradient growth threshold (an effect removed by adapting only the SSIM window to resolution).
To improve quality and allow at least as much capacity at higher resolution, we halve the threshold from $2\times10^{-4}$ to $10^{-4}$.
This produces $3.19\times$ as many Gaussians and improves PSNR by $0.49$\,dB, indicating that the default threshold restricts useful growth.
However, this larger model may still offer an unsuitable quality--size trade-off, leaving users to select a threshold through retraining models from scratch.

Gradient refinements and error-driven densification~\citep{ye2024absgs,zhang2024pixelgs,bulo2024revising} use quality thresholds that may need retuning per capture.
Speedy-Splat~\citep{hanson2025speedysplat}, LightGaussian~\citep{fan2024lightgaussian}, and PUP~3D-GS~\citep{hanson2025pup} prune fixed fractions using importance or sensitivity scores.

\subsection{Preset count controllers}
\label{sec:related-presetcount}

Preset-count methods~\citep{fang2024minisplatting,kheradmand2024mcmc,litegs} require both a suitable, manually supplied target and a growth rule that can reach it.
When the target is reliably reached, this provides a direct way to control model size, but choosing a suitable value for each capture still requires trial and error.
Even a well-chosen target is not always reached: Taming~3DGS~\citep{tamings3dgs} multiplies the initial SfM point count by a user-provided or dataset-specific factor, but limits growth to Gaussians that pass vanilla 3DGS's gradient growth threshold.
On Zip-NeRF's London scene, all three released multipliers ($2\times$, $5\times$, $15\times$) yield approximately $0.30$M Gaussians.

\subsection{Automatic budget controllers}
\label{sec:related-automaticbudget}
\label{sec:method-confounding}

Automatic-budget controllers, including learned methods, determine model size during optimization through growth and retention decisions~\citep{dong2025cdcgs,ren2026fastgs,controlgs2025,zhang2024lp3dgs}.
Compact~3DGS~\citep{lee2024compact} learns binary masks on Gaussian opacity and scale by balancing reconstruction loss against a sparsity penalty.
Although its final Gaussian count emerges from optimization, the quality--size trade-off depends on the penalty weight and the inherited gradient growth threshold.
On London, reducing the default mask penalty to zero increases the Gaussian count from $0.17$M to $0.52$M and improves PSNR by $0.21$\,dB, while also halving that threshold increases the count to $1.28$M and improves PSNR by a further $0.24$\,dB.

DashGaussian~\citep{chen2025dashgaussian} estimates a budget from observed growth, while LeGS~\citep{ning2026legs} learns growth and pruning decisions.
Because both retain fixed gradient growth thresholds, they automate decisions only within candidate limits that may still require retuning across captures.

Across all classes, choosing a quality--size trade-off naturally involves constants; the limitation is that existing constants act on signals that shift with resolution and scene extent, breaking transfer across scales.
For a single calibration to transfer across scales, the signals those constants control must track observed 3D surface and resolved detail.

\section{Method}
\label{sec:method}

\begin{table}[t]
    \centering
    \footnotesize
    \setlength{\tabcolsep}{2.2pt}
    \caption{Capacity signals under transformations.}
    \label{tab:scale-principles}
    \begin{tabular}{@{}lccccc@{}}
        \toprule
        & \textbf{(P1)} & \textbf{(P2)} & \textbf{(P3)} & \textbf{(P4)} & \textbf{(P5)} \\
        Capacity signal & $k\times$ dup. & $+B$ burst & $A \sqcup B$ & $c\times$ pts. & $\alpha\times$ width \\
        \midrule
        Fixed preset ($N_0$) & $1\times$ & $O(1)$ & fixed & $1\times$ & $1\times$ \\
        SfM points ($M$) & $1\times$ & $O(1)$ & add. & $c\times$ & $1\times$ \\
        Raw pixel sum ($\pix$) & $k\times$ & $O(B)$ & add. & $1\times$ & $\alpha^2\times$ \\
        View-avg.\ MSE & $1\times$ & $O(1)$ & diluted & $1\times$ & $1\times$ \\
        \midrule
        \textbf{\method{}} & $\mathbf{1\times}$ & $\mathbf{O(1)}$ & \textbf{add.} & $\mathbf{O(1)}$ & $\boldsymbol{\alpha\times}$ \\
        \bottomrule
    \end{tabular}
\end{table}

Given $N$ posed training images $\{I_i\}_{i=1}^{N}$ of loaded resolution $W_i \times H_i$ and their sparse SfM reconstruction, we want a 3D Gaussian set $\mathcal{G}$ ($\numg = |\mathcal{G}|$) that balances reconstruction quality and model size while accounting for scene extent, image resolution, and content complexity, without scene-specific tuning.
We first formalize five capacity-scaling principles (\cref{sec:method-principles}) and use them to derive coverage-scaled exploration from a pre-training scene coverage measure $\capsupp$ (\cref{sec:method-measure}).
We then motivate selecting capacity after growth within the exploration cap and introduce coverage-weighted self-distillation (\cref{sec:method-consolidation}).

\subsection{Capacity-scaling principles}
\label{sec:method-principles}

The capacity needed to represent a scene should depend on its content and the detail to be resolved, not on how that same content is sampled.
We present five principles that describe how capacity should respond to changes in the scene and how it is captured:
\begin{itemize}[leftmargin=*,itemsep=2pt,topsep=3pt]
    \item \textbf{(P1) Duplication invariance.} Repeating every training view $k$ times adds no 3D information and leaves capacity unchanged ($\times 1$).
    \item \textbf{(P2) Burst boundedness.} Adding $B$ near-duplicate views of an already observed region contributes bounded weight ($O(1)$ as $B \to \infty$) rather than growing with $B$.
    \item \textbf{(P3) Disjoint additivity.} Reconstructing non-overlapping sub-scenes $A \sqcup B$ together allocates the sum of their individual capacities.
    \item \textbf{(P4) SfM point-density boundedness.} Extracting $c\times$ denser SfM points across the same observed surfaces (for example, changing the feature detector, matcher threshold, or local texture density) adds no new surface and keeps capacity bounded ($O(1)$ as $c \to \infty$) rather than scaling with point count.
    \item \textbf{(P5) Boundary resolution scaling.} Scaling image dimensions uniformly ($\alpha W_i \times \alpha H_i$) increases capacity approximately linearly with image dimensions ($\times \alpha$) rather than pixel area ($\times \alpha^2$).
\end{itemize}

\begin{figure}[t]
    \centering
    \includegraphics[width=\linewidth]{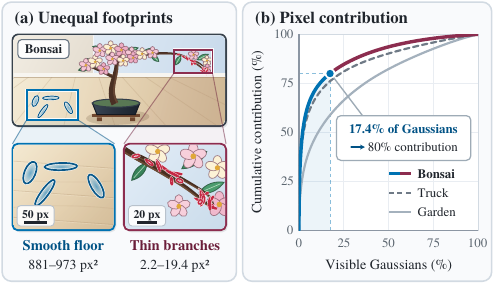}
    \caption{\textbf{Unequal footprints in 3DGS.} \textbf{(a)}~Stylized Bonsai regions with measured $3\sigma$ footprints. \textbf{(b)}~Cumulative rendered pixel contribution, by decreasing footprint.}
    \label{fig:gaussian-footprints}
\end{figure}

Principle~\textbf{(P5)} is motivated by how anisotropic 3D Gaussians partition image space (\cref{fig:gaussian-footprints}): on Bonsai, the largest $17.4\%$ of visible primitives ($23.1\%$ on Truck, $45.5\%$ on Garden) supply $80\%$ of total pixel contribution. As image resolution increases, smooth regions can largely reuse existing Gaussians, while boundaries and thin structures are expected to require more, smaller Gaussians, roughly in proportion to their projected length in pixels.
This motivates approximating resolution-dependent capacity by a length rather than an area; we use $\sqrt{W_iH_i}$, which grows linearly when both image dimensions increase by the same factor.

\Cref{tab:scale-principles} compares candidate capacity signals; each baseline violates at least two principles. To limit the contribution of repeated coverage (\textbf{P2}) while preserving additivity across disjoint scenes (\textbf{P3}), we account for surfaces shared across views, motivating the scene coverage measure below.

\subsection{Coverage-scaled exploration}
\label{sec:method-measure}
\label{sec:method-budget}

\begin{figure*}[t]
    \centering
    \includegraphics[width=\linewidth]{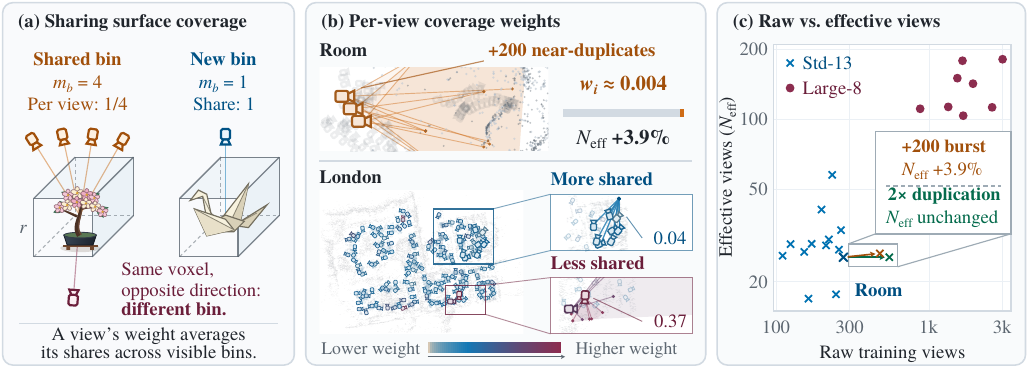}
    \caption{\textbf{Per-view coverage weights.} \textbf{(a)}~Views share unit weight within each coverage bin. \textbf{(b)}~Measured coverage weights on Room and London discount repeated coverage. \textbf{(c)}~Effective versus raw view counts across 21 scenes.}
    \label{fig:method-coverage}
\end{figure*}

\paragraph{Local 3D overlap sharing for (P1)--(P4).}
We construct the scene coverage measure by combining contributions from training views, weighting each view to discount overlap with other views.
Since a 3D Gaussian can represent a surface patch across multiple views, we assign that patch a single unit of weight, shared equally among the $m$ views observing the same side ($1/m$ each; \cref{fig:method-coverage}a).
To obtain a bounded response to SfM point density (\textbf{P4}) while remaining independent of scene coordinate units, we partition training-view SfM points into coarse voxels of width $r = \beta s_{\mathrm{cam}}$ ($\beta = 0.02$), where $s_{\mathrm{cam}}$ is the weighted median camera-to-scene distance normalized by dimensionless focal length $\sqrt{f_{x,i}f_{y,i}/(W_iH_i)}$ (with focal lengths scaled to the training-image resolution).
To preserve rigid-motion invariance and exact independence across disconnected sub-scenes, the voxel frame and $s_{\mathrm{cam}}$ are estimated within each connected component of the camera–point visibility graph using overlap-discounted camera weights computed from the number of training views observing each SfM point, before constructing the voxel grid (exact coordinate formulas in Supplementary~\S B).
We define a \emph{coverage bin} as an occupied voxel paired with the camera-to-point ray's signed dominant axis ($\pm x, \pm y, \pm z$) in this frame, so that opposite faces of thin walls or leaves form separate bins (\cref{fig:method-coverage}a).

Let $\mathcal{B}_i$ denote the set of distinct coverage bins formed by training view $i$'s SfM feature-track observations, and let $m_b = |\{j : b \in \mathcal{B}_j\}|$ be the number of training views observing bin $b$.
For nonempty $\mathcal{B}_i$, view $i$'s coverage weight is the mean share it holds across its visible bins:
\begin{equation}
    \covw = \frac{1}{|\mathcal{B}_i|} \sum_{b \in \mathcal{B}_i} \frac{1}{m_b} \in (0, 1].
    \label{eq:view-weight}
\end{equation}
Empty views receive $\covw=0$.
These weights satisfy \textbf{(P1)}--\textbf{(P4)}.
An isolated view ($m_b = 1$) receives $\covw = 1$.
Duplicating all views $k$ times multiplies every $m_b$ by $k$ and divides every $\covw$ by $k$, keeping effective view count $N_{\mathrm{eff}} = \sum_{i=1}^{N} \covw$ exact (\textbf{P1}; \cref{fig:method-coverage}c).
With fixed binning, adding $B$ burst views over $\mathcal{B}_{\mathrm{burst}}$ increases $m_b$ only within $\mathcal{B}_{\mathrm{burst}}$, keeping total bin mass bounded as $B \to \infty$; bounded capture extent and positive minimum normalized camera distance bound the response when binning is recomputed (\textbf{P2}; $+3.9\%$ on Room for $B = 200$).
Combining visibility-disjoint captures leaves each component’s voxel frame, the number of views observing each bin \(m_b\), and the coverage weights \(\covw\) unchanged (\textbf{P3}); when adjacent rooms share doorway tracks inside one component, $s_{\mathrm{cam}}$ remains a per-camera median normalized distance rather than growing with scene diameter, and $m_b$ stays local to each bin (\cref{sec:exp-proxy-validation}).
$0\leq\covw\leq1$ bounds view coverage weights independently of point count; once the frame, scale, and occupied-bin sets stabilize, denser SfM points leave weights unchanged (\textbf{P4}).
Measured weights show overlap sharing: Room's burst views have $\covw \approx 0.004$, while London's views of less-shared surfaces have higher weights (\cref{fig:method-coverage}b).

\paragraph{Scene coverage for (P5) and exploration cap.}
The coverage-weighted linear image scale $\bsupp = \covw \sqrt{W_i H_i}$ combines each view's coverage weight with image resolution.
Summing over views defines the scene coverage measure $\capsupp$ and sets the exploration cap $\tcap$:
\begin{equation}
    \capsupp = \sum_{i=1}^{N} \covw \sqrt{W_i H_i},
    \qquad
    \tcap = \operatorname{round}(\kappa \capsupp),
    \label{eq:capture-budget}
\end{equation}
where $\kappa > 0$ is the global exploration scaling factor, calibrated once (\cref{sec:method-calibration}).

\noindent\textbf{Proposition 1} (\emph{Scene coverage measure}).
\emph{Computed from training views alone under the conditions above, $\capsupp$ satisfies \textbf{(P1)}--\textbf{(P4)}, linear resolution scaling $\capsupp(\alpha W, \alpha H) = \alpha \capsupp(W, H)$ \textbf{(P5)}, and rigid-motion, similarity-scale, and test-split invariance (proof in Supplementary~\S B).}

\subsection{Exploration and self-distillation}
\label{sec:method-consolidation}
\label{sec:method-controller}

\paragraph{Why does $\tcap$ bound exploration rather than decide final size?}
3DGS reconstruction quality depends strongly on the optimization trajectory.
Densification discovers fine structure through greedy local splitting and cloning.
As primitives move, rotate, and resize away from sparse SfM points, many early splits become occluded or redundant once surfaces settle, but restricting the exploration cap from the start deprives early optimization of the transient headroom needed to bridge gaps and resolve thin geometry. Conversely, continuing periodic pruning until the end of training repeatedly perturbs surviving neighbors after learning rates have decayed.

We therefore decouple exploration from final selection.
During the growth phase (first $80\%$ of training), budgeted densification~\citep{tamings3dgs,litegs} grows toward $\tcap$ by sampling parents using opacity-gradient variance scores, without a fixed positional-gradient threshold.
Together with periodic soft prunes retaining $\krho_0 = 0.60$ of pre-growth primitives~\citep{hanson2025speedysplat}, this produces a teacher model $\mathcal{G}_{\mathrm{pre}}$ of size $\npre = |\mathcal{G}_{\mathrm{pre}}|$.
At $80\%$ of training, we apply a single self-distillation step to $\mathcal{G}_{\mathrm{pre}}$ and refine the retained subset uninterrupted over the remaining $20\%$ refinement phase.

\paragraph{Self-distillation under the same coverage measure.}
We seek a smaller subset of Gaussians that preserves the teacher's rendered images, balancing teacher rendering error against a cost $\lambda$ for each retained Gaussian.
To apply \textbf{(P1)}--\textbf{(P5)} during pruning, we formulate coverage-weighted self-distillation as a rate-distortion objective against the teacher renderings $\hat{I}_i(\mathcal{G}_{\mathrm{pre}})$, weighting each view's mean-squared rendering error by the same coverage-weighted linear image scale $\bsupp = \covw \sqrt{W_i H_i}$:
\begin{equation}
    \min_{\mathcal{G} \subseteq \mathcal{G}_{\mathrm{pre}}} \;
    \sum_{i=1}^{N} \bsupp \cdot \frac{1}{W_i H_i} \bigl\| \hat{I}_i(\mathcal{G}_{\mathrm{pre}}) - \hat{I}_i(\mathcal{G}) \bigr\|_2^2
    \;+\; \lambda \, |\mathcal{G}|.
    \label{eq:rd-objective}
\end{equation}
Using $\bsupp = \covw \sqrt{W_i H_i}$ cancels one $\sqrt{W_i H_i}$ factor, weighting view $i$'s pixel-summed squared error by $\covw / \sqrt{W_i H_i}$.

In front-to-back alpha compositing, removing one primitive $g$ of view color $\mathbf{c}_g(i)$ and pixel opacity $\alpha_g(p)$ changes pixel $p$'s rendered color by $\Delta \hat{\mathbf{C}}_{g,i}(p) = T_g(p)\alpha_g(p)\bigl(\mathbf{c}_g(i) - \hat{\mathbf{C}}_{i,>g}(p)\bigr)$, where $T_g(p)$ is the transmittance reaching $g$ and $\hat{\mathbf{C}}_{i,>g}(p)$ is the color composited strictly behind $g$ with unit incoming transmittance, including the background (Supplementary~\S B).
This leave-one-out color change is exact to all orders in $\alpha_g(p)$ and accumulated across all primitives in one rasterization pass~\citep{hanson2025speedysplat,litegs}.
Letting $u_{g,i} = \sum_{p} \|\Delta \hat{\mathbf{C}}_{g,i}(p)\|_2^2$ denote primitive $g$'s leave-one-out RGB squared error in view $i$, approximating the teacher rendering error in \cref{eq:rd-objective} by the sum of leave-one-out errors gives an additive surrogate minimized by the threshold rule:
\begin{equation}
    \mathcal{G}^{*} = \Bigl\{ g \in \mathcal{G}_{\mathrm{pre}} \;:\;
    \util = \sum_{i=1}^{N} \frac{\covw}{\sqrt{W_i H_i}} \, u_{g,i} > \lamstar \Bigr\}.
    \label{eq:rd-threshold}
\end{equation}
The teacher model is scored once over the training views, and the retained subset is selected by one threshold application; subsequent optimization is confined to the refinement phase.
Because $\util$ weights each view by $\covw / \sqrt{W_i H_i}$ and scores leave-one-out pixel errors rather than SfM points, it inherits \textbf{(P1)}--\textbf{(P4)}.
Under the boundary-scaling approximation (\textbf{P5}), $u_{g,i}$ grows roughly with projected contour length ($u_{g,i} \propto \sqrt{W_i H_i}$), motivating the factor $1/\sqrt{W_i H_i}$ to limit resolution-driven growth under a fixed $\lamstar$.

We validate the additive approximation and compare alternative weighting and scoring rules in Supplementary~\S E.

\subsection{One-time reference calibration}
\label{sec:method-calibration}

\method{} calibrates $(\kappa, \lamstar)$ once on the 13 standard training sets ($\mathcal{S}_{\mathrm{ref}}$) against a reference mean budget $\bref = 0.925$M, using zero test views, quality metrics, or large-capture statistics:
(1)~setting the geometric-mean exploration cap on $\mathcal{S}_{\mathrm{ref}}$ to $2\bref$ (providing $2\times$ exploration headroom) fixes $\kappa = 62.04$;
(2)~bisection on the utility scores $\{\util\}$ across $\mathcal{S}_{\mathrm{ref}}$ to match mean retained count $\bref$ fixes $\lamstar = 1.0861 \times 10^{-5}$.
If the default model does not offer the desired quality--size trade-off, $\lambda$ can be increased for a more compact model or decreased to retain more primitives.
Each alternative reuses the teacher model and requires only self-distillation and the final $20\%$ refinement, allowing different quality--size trade-offs without repeating the growth phase (Supplementary~\S J).
We implement \method{} in LiteGS~\citep{litegs} with unchanged rasterizer, optimizer, and schedules, rounding retained counts to $128$-primitive clusters with a $10\%$ safety floor (inactive in all tested default benchmark runs) for the final $20\%$ refinement phase (Supplementary~\S B).

\section{Experiments}
\label{sec:experiments}

We compare \method{} against 13 controllers on 13 standard scenes and test transfer to eight large captures without retuning.
We then isolate the contributions of exploration and self-distillation and assess the robustness of the proposed scene coverage measure.

\subsection{Setup}
\label{sec:exp-setup}

On the benchmarks introduced in \cref{sec:related}, we rerun 13 released baselines with $30$k requested iterations on standard scenes (official eighth-frame splits) and $60$k on large captures, transferring settings unchanged.
We run all experiments on a single Nvidia GH200 and average \method{} benchmark results over three random seeds.
For evaluation, we compute LPIPS~\citep{zhang2018lpips} using VGG features.
Dataset preparation details appear in Supplementary~\S B--\S D.

\begin{figure*}[t]
    \centering
    \includegraphics[width=\linewidth]{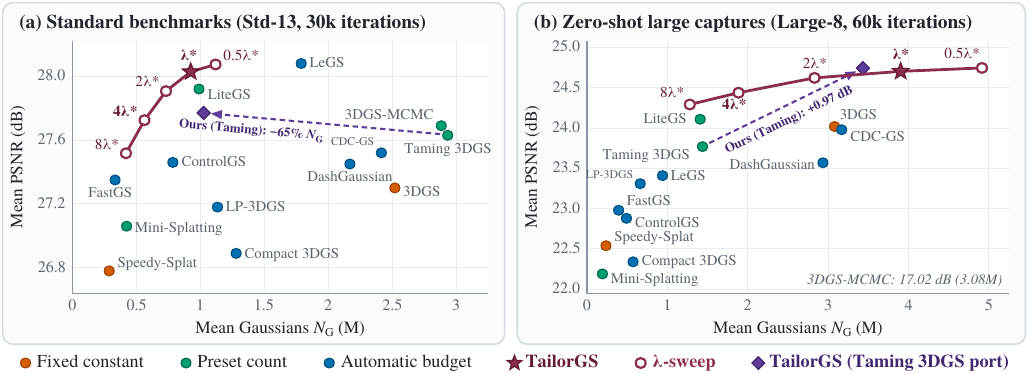}
    \caption{\textbf{Quality--size trade-offs across capture scales.} TailorGS lies on the Pareto front on (a) standard scenes and transfers to (b) large captures without retuning. Dashed arrows show the effect of TailorGS on Taming 3DGS. 3DGS-MCMC diagnosed in Supplementary~\S C.}
    \label{fig:pareto-cross-scale}
\end{figure*}

\subsection{Evaluation on standard benchmarks}
\label{sec:exp-automatic}
\label{sec:exp-fixedconst}
\label{sec:exp-fixedcount}

Even on the benchmarks where existing controllers tune their presets, thresholds, and penalties, \method{} ($0.93$M, $27.99 \pm 0.03$\,dB, $0.195$ LPIPS) lies on the quality--size Pareto front of each dataset while averaging $230$\,MB and $584$\,FPS (\cref{fig:pareto-cross-scale}a; Supplementary~\S C,~\S I).
High-count baselines~\citep{kerbl3Dgaussians,chen2025dashgaussian,dong2025cdcgs,kheradmand2024mcmc,tamings3dgs} spend more than twice as many Gaussians for lower PSNR, while compact controllers~\citep{hanson2025speedysplat,fang2024minisplatting,ren2026fastgs,controlgs2025,zhang2024lp3dgs,lee2024compact} fall below our trade-off curve across sizes; compact $4\lamstar$ alone exceeds 11 of the 13 baselines in PSNR.
Only two baselines approach this curve: LeGS~\citep{ning2026legs} gains under $0.1$\,dB in mean PSNR at nearly double our count (while trailing \method{} in PSNR and LPIPS on Tanks~\&~Temples), and LiteGS~\citep{litegs} lands nearby because it operates near our calibrated reference budget.
Matching the benchmark mean still misallocates capacity across scenes: within the same average budget, \method{} reclaims primitives on smooth interiors (\emph{playroom}: $-68\%$ count, $-0.21$\,dB) to fund foliage-dense scenes (\emph{bicycle}: $+0.89$\,dB, $-0.065$ LPIPS), improving mean PSNR, SSIM, LPIPS, storage, and FPS simultaneously.

\subsection{Transfer, quality--size trade-offs, portability}
\label{sec:exp-generalization}

\paragraph{Transfer to large captures without retuning.}
Transferred unchanged from Std-13, \method{} automatically scales mean capacity from $0.93$M to $3.91$M ($1.59$M--$6.07$M across captures) and ranks first across all 13 baselines in overall ($24.74 \pm 0.03$\,dB), aerial, and Zip-NeRF PSNR, SSIM ($0.821$), and LPIPS ($0.258$), winning seven of eight captures in both PSNR and LPIPS (\cref{fig:pareto-cross-scale}b; Supplementary~\S C,~\S D).
Rather than scaling up, seven of the 13 baselines allocate \emph{fewer} Gaussians on Large-8 than on Std-13: regularized-mask and pruning controllers shrink below $0.7$M, LeGS drops by nearly half (ending below SfM initialization on every aerial scene), and Taming~3DGS drops by half, filling only $7\%$--$12\%$ of its $15\times$ aerial budgets and reaching at most $0.44$M on Zip-NeRF (\cref{sec:related-presetcount}).
Preset LiteGS (PSNR runner-up, trailing by $0.63$\,dB and $0.064$ LPIPS) stalls on all four aerial scenes because SfM initialization ($1.60$M--$2.13$M) already exceeds its $1$M target count.
Even the baselines that grow to $\sim 3$M (vanilla 3DGS, CDC-GS~\citep{dong2025cdcgs}, and DashGaussian) over-grow aerial scenes (up to $7.4$M on Mill19) yet starve multi-room Zip-NeRF interiors ($0.63$M--$1.06$M vs.\ our $1.59$M--$3.80$M), blurring thin architectural detail (\cref{fig:pipeline}c).

\paragraph{Cross-scale quality--size trade-offs.}
To verify that Large-8 gains do not merely reflect higher capacity, we sweep $\lambda \in \{8, 4, 2, 1, 0.5\}\lamstar$ across all 21 scenes from the $80\%$ teacher checkpoints (\cref{fig:pareto-cross-scale}; Supplementary~\S J): unlike presets or sparsity penalties that require rerunning densification for each target size, one growth run provides quality--size trade-offs across both scales.
On Large-8, every swept point surpasses all 13 baselines in both mean PSNR and LPIPS.
Even $8\lamstar$ uses fewer Gaussians than LiteGS and Taming~3DGS while outperforming every baseline, and $4\lamstar$ beats vanilla 3DGS in PSNR and LPIPS ($-0.023$) with $39\%$ fewer Gaussians (Supplementary~\S C).

\begin{figure*}[t]
    \centering
    \includegraphics[width=\linewidth]{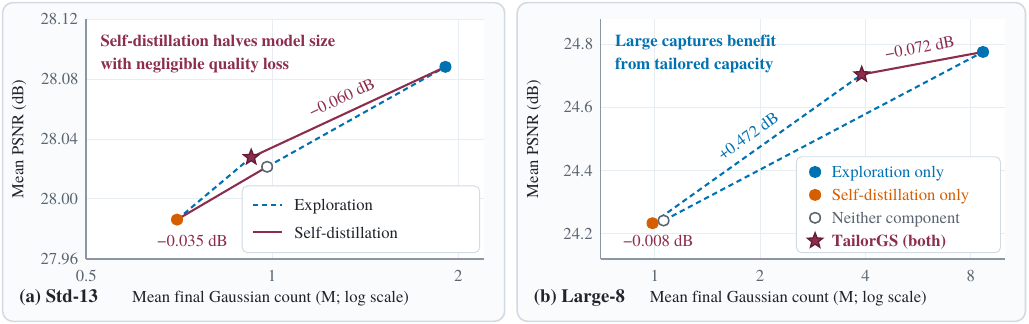}
    \caption{\textbf{Factorial ablation of coverage-scaled exploration and self-distillation.} Mean test PSNR versus final Gaussian count (log scale) on Std-13 and Large-8.}
    \label{fig:controller-components}
\end{figure*}

\paragraph{Controlled manual retuning of baselines.}
To test whether simple adjustments close the large-capture gap without per-scene trial and error, we run controlled retuning sweeps on \emph{nyc}, \emph{residence}, and \emph{london} (Supplementary~\S G).
Sweeping vanilla 3DGS's gradient growth threshold over $\{0.25, 0.5, 1, 2, 4\}\times$ moves optimal settings in opposite directions across captures: default $2\times 10^{-4}$ peaks on \emph{nyc} ($27.19$\,dB), $4\times 10^{-4}$ peaks on \emph{residence} ($21.27$\,dB), and $0.25\times$ triggers severe instability ($-5.91$\,dB on \emph{nyc}; $-3.23$\,dB at $13.1$M on \emph{residence}).
Extending the \emph{london} diagnostics in \cref{sec:related-presetcount,sec:related-automaticbudget}, Taming~3DGS reaches its $15\times$ budget in an 18-run grid only when jointly cutting its gradient growth threshold $8\times$ to $2.5\times 10^{-5}$ ($3.07$M, $26.66$\,dB, within $0.15$\,dB of \method{}), while retuned Compact~3DGS ($25.49$\,dB) trails \method{} by $1.32$\,dB.

\paragraph{Portability to Taming 3DGS.}
We port the controller to Taming~3DGS (\cref{fig:pareto-cross-scale}; Supplementary~\S H).
Controller parameters $(\kappa,\lamstar)$ transfer unchanged to all 21 scenes.
Relative to released Taming~3DGS, \method{} reduces model size on Std-13 while preserving mean reconstruction quality, and expands capacity on Large-8, improving PSNR by $0.09$--$2.08$\,dB and lowering LPIPS on every capture.

\subsection{Ablations: exploration, self-distillation, timing}
\label{sec:exp-contributions}

\paragraph{Exploration and self-distillation.}
\Cref{fig:controller-components} evaluates the $2 \times 2$ factorial matrix across all 21 scenes: a fixed $1$M cap transferred unchanged to Large-8 versus coverage-scaled exploration caps, each with and without self-distillation, under otherwise identical growth and evaluation settings.
To isolate our contributions, all variants share our growth-phase pruning, making ``Neither component'' differ from released LiteGS.
Paired variants share the $80\%$ teacher checkpoint and complete the same remaining training schedule; enabled variants apply the self-distillation threshold once before refinement.
Coverage-scaled exploration preserves Std-13 quality while improving Large-8 mean PSNR by roughly $0.5$\,dB, both with and without self-distillation.
With these exploration caps, self-distillation cuts mean final count by more than half across both scales with under $0.08$\,dB mean PSNR loss.
Using the same $\util > \lamstar$ rule in all enabled variants and scenes, self-distillation retains only $12.3\%$ of Gaussians on \emph{playroom} but up to $86.9\%$ on \emph{garden} under these caps, illustrating scene-dependent retention.

\begin{figure}[t]
    \centering
    \includegraphics[width=\linewidth]{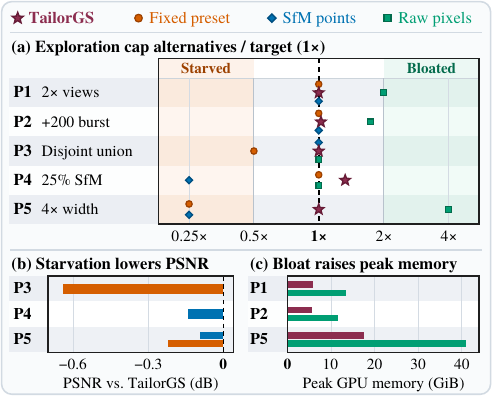}
    \caption{\textbf{(a)}~Exploration caps relative to target scaling (\cref{tab:scale-principles}). \textbf{(b)}~Test PSNR vs.\ \method{} when starved, and \textbf{(c)}~peak training memory (\method{} top, raw pixels bottom) when inflated.}
    \label{fig:capacity-principles}
\end{figure}

\paragraph{Self-distillation timing.}
Varying the self-distillation point $\tau \in \{0.60, 0.70, 0.80, 0.90\}$ across four count-matched standard scenes and two large captures without retuning confirms quality insensitivity (Supplementary~\S E): standard-scene PSNR spans $0.024$\,dB across $\tau \in [0.60, 0.80]$, dipping at $\tau = 0.90$ when only $10\%$ refinement remains.
Applying self-distillation earlier at $\tau = 0.60$ changes mean training wall-clock time by $-20.5\%$ relative to our default $\tau = 0.80$.

\subsection{Empirical test of capacity-scaling principles}
\label{sec:exp-proxy-validation}

Across 36 capture transformations, $\capsupp$ satisfies the predicted invariances and scaling laws within $5 \times 10^{-16}$ relative error (Supplementary~\S F).
Comparing the capacity signals in \cref{tab:scale-principles} as exploration caps calibrated to the same Std-13 geometric mean under identical optimization and self-distillation (\cref{fig:capacity-principles}) reveals two failure modes.
Signals that under-scale restrict exploration before self-distillation can act, degrading reconstruction quality.
Conversely, raw-pixel allocation inflates the teacher model, nearly doubling final count when the $10\%$ safety floor binds and still losing $0.21$\,dB at 5.2K despite self-distillation removing $85\%$ of the teacher.
Post-growth pruning alone cannot rescue a mis-scaled cap: avoiding early starvation and teacher bloat requires satisfying \textbf{(P1)}--\textbf{(P5)} in both phases.

\section{Conclusion}
\label{sec:conclusion}

Existing 3DGS controllers confound scene content with how it is photographed, causing benchmark-tuned settings to starve or bloat larger captures.
\method{} separates observed surface and resolution from content learned during growth: one 3D coverage measure bounds exploration, then distills the teacher to the Gaussians the scene needs.
Calibrated once on standard training views, \method{} matches the best baseline's mean PSNR on standard scenes at nearly half the size, leads all 13 baselines on large captures, and makes model size a post-growth choice, allowing users to explore quality--size trade-offs without retraining from scratch.
Extending this approach beyond photographs to LiDAR-based 3DGS systems is a promising future direction.

\clearpage
\subsection*{LLM Usage Statement}
In this work, we used generative AI tools (GPT-6 Astra and Gemini 4 Carbon) to assist with methodological and experimental design, provide feedback, and implement methods. All AI-generated code was read and tested by the authors. Additionally, we used the same tools to generate tables and refine figures from experiment logs and to edit the text of this paper for readability and grammar.

{\small
\bibliographystyle{ieeenat_fullname}
\bibliography{refs}
}

\onecolumn
\pdfbookmark[0]{Supplementary Material}{supplement}
\begin{center}
{\Large\bfseries Supplementary Material\par}
\end{center}
\setcounter{topnumber}{4}
\setcounter{bottomnumber}{3}
\setcounter{totalnumber}{6}
\renewcommand{\topfraction}{0.95}
\renewcommand{\bottomfraction}{0.9}
\renewcommand{\textfraction}{0.05}
\renewcommand{\floatpagefraction}{0.75}
\raggedbottom
\makeatletter
\@addtoreset{table}{section}
\@addtoreset{figure}{section}
\makeatother
\renewcommand{\thetable}{\thesection.\arabic{table}}
\renewcommand{\thefigure}{\thesection.\arabic{figure}}
\renewcommand{\theHfigure}{supp.\thesection.\arabic{figure}}
\renewcommand{\theHtable}{supp.\thesection.\arabic{table}}
\setcounter{equation}{0}
\renewcommand{\theHequation}{supp.\arabic{equation}}

\noindent\textbf{Contents}
\begin{itemize}[leftmargin=1.6em,itemsep=1pt,topsep=3pt]
    \item \textbf{\Cref{sec:supp-qualitative}}: Qualitative comparisons on large handheld and aerial captures.
    \item \textbf{\Cref{sec:supp-impl}}: Canonical voxel frame, proof of Proposition~1, leave-one-out derivation, LiteGS details, and protocol.
    \item \textbf{\Cref{sec:supp-adaptive}}: Cross-scale benchmark summary, standard-dataset breakdowns, seed stability, and baseline reproduction.
    \item \textbf{\Cref{sec:supp-transfer}}: Complete per-capture comparisons and seed stability on the eight large captures.
    \item \textbf{\Cref{sec:supp-keepratio}}: One-shot versus sequential self-distillation, view-weighting and scoring rules, factorial and oracle-count ablations, resolution scaling, and timing.
    \item \textbf{\Cref{sec:supp-proxy-validation}}: Transformation invariance checks, camera-redundancy tests, per-scene coverage statistics, and SfM studies.
    \item \textbf{\Cref{sec:supp-reported}}: Extended related work, controller taxonomy, and controlled manual retuning sweeps on large captures.
    \item \textbf{\Cref{sec:supp-second-base}}: Portability of \method{} to Taming~3DGS with unchanged $(\kappa, \lamstar)$.
    \item \textbf{\Cref{sec:supp-efficiency}}: Per-capture saved model storage and native rendering throughput across all 21 scenes.
    \item \textbf{\Cref{sec:supp-capacity-objective}}: Post-growth quality--size trade-offs from one teacher checkpoint per scene across all 21 scenes.
\end{itemize}

\appendix

\section{Qualitative Comparisons}
\label{sec:supp-qualitative}

\begin{figure}[!htbp]
\centering
\includegraphics[width=0.83\linewidth]{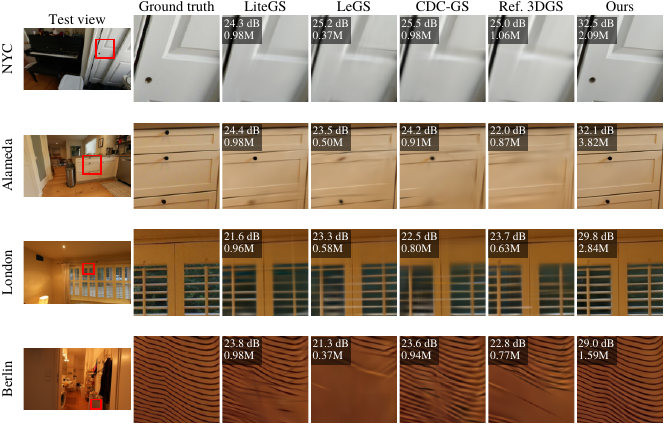}
\caption{\textbf{Qualitative comparison on large handheld captures (Zip-NeRF).} Held-out crops annotated with crop PSNR and final Gaussian count $\numg$. Baselines transferred from standard scenes blur door bevels (\emph{NYC}), cabinet seams (\emph{Alameda}), shutter slats (\emph{London}), and floor wood grain (\emph{Berlin}) that \method{} resolves without retuning.}
\label{fig:supp-selected-handheld}
\end{figure}

\begin{figure}[!htbp]
\centering
\includegraphics[width=0.83\linewidth]{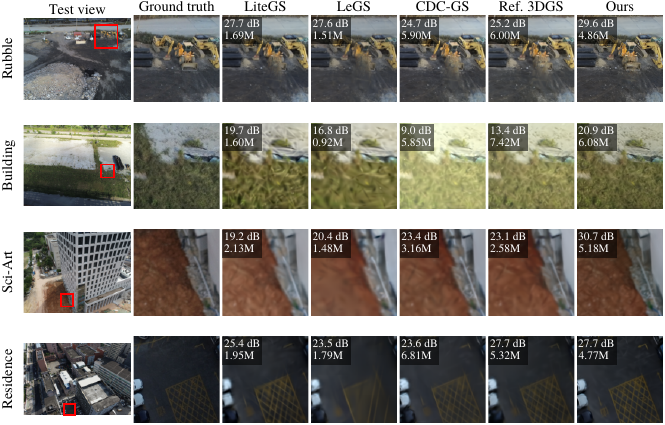}
\caption{\textbf{Qualitative comparison on large aerial captures (Mill19 and UrbanScene3D).} Held-out crops showing construction vehicles (\emph{Rubble}), grass boundaries (\emph{Building}), ground texture (\emph{Sci-Art}), and parking-lot markings (\emph{Residence}). Vanilla 3DGS and CDC-GS allocate up to $1.4\times$ as many Gaussians as \method{} on \emph{Rubble}, \emph{Building}, and \emph{Residence} without recovering these fine structures.}
\label{fig:supp-selected-aerial}
\end{figure}

\Cref{fig:supp-selected-handheld,fig:supp-selected-aerial} compare held-out test renders on all eight large captures, covering four multi-room handheld interiors from Zip-NeRF~\citep{barron2023zipnerf} and four aerial sites from Mill19~\citep{turki2022meganerf} and UrbanScene3D~\citep{lin2022urbanscene3d}.
The leftmost column shows the crop location in the full test view, and each method panel gives the floating-point crop PSNR and final Gaussian count $\numg$.
\emph{Ref.\ 3DGS} refers to released vanilla 3DGS~\citep{kerbl3Dgaussians} trained for $60$k iterations, and \method{} renders use seed 4114.
On the multi-room interiors (\cref{fig:supp-selected-handheld}), the shown baselines allocate too few Gaussians ($0.37$M--$1.06$M) and blur thin architectural detail.
On the aerial captures (\cref{fig:supp-selected-aerial}), vanilla 3DGS and CDC-GS use up to $1.4\times$ as many Gaussians as \method{} while still missing fine ground markings and object boundaries.

\clearpage
\section{Method Details and Experimental Protocol}
\label{sec:supp-impl}

\subsection{Canonical voxel frame for scene coverage}
In the main paper, coverage bins partition training-view SfM points into voxels of width $r = \beta s_{\mathrm{cam}}$ paired with the camera-to-point ray's signed dominant axis.
Constructing axis-aligned voxels directly in world coordinates would tie $\capsupp$ to the arbitrary origin, orientation, and scale of an SfM reconstruction, and would couple disconnected sub-scenes through a shared global scale.
We link each training camera to the SfM points it observes to form the camera--point visibility graph $\mathcal{E}_{\mathrm{train}}$, and give each connected component $q$ its own coordinate frame and voxel width.
For example, if camera A shares a point with B and B shares another with C, all three use the same grid.
A training view with no observed points receives no bins ($\mathcal{B}_i = \emptyset$) and zero coverage weight ($\covw = 0$).

\paragraph{1. Overlap-discounted camera weights.}
When setting the grid origin, orientation, and voxel width, we give each camera $i$ a scalar weight $\pi_i$ to limit the influence of repeated views and camera bursts.
For training camera $i \in q$ observing SfM points $\mathcal{P}_i \neq \emptyset$, let $t_p \ge 1$ be the number of training views observing $p \in \mathcal{P}_i$.
Camera $i$ receives mean point-share weight $\widetilde{a}_i = |\mathcal{P}_i|^{-1} \sum_{p \in \mathcal{P}_i} 1/t_p > 0$, normalized within component $q$ as $\pi_i = \widetilde{a}_i / \sum_{j \in q} \widetilde{a}_j$.

\paragraph{2. Grid origin and orientation.}
Each camera center $\mathbf{c}_i = (x_i,y_i,z_i)^\top$ gives the camera's 3D position. We place the grid origin at the weighted centroid $\boldsymbol{\mu}_q = \sum_{i \in q} \pi_i \mathbf{c}_i$, averaging each coordinate using the same camera weights.
We compute the $3 \times 3$ covariance matrix $\mathbf{C}_q = \sum_{i \in q} \pi_i (\mathbf{c}_i - \boldsymbol{\mu}_q)(\mathbf{c}_i - \boldsymbol{\mu}_q)^\top$. Its eigenvectors, ordered by decreasing eigenvalue, define the three grid axes and form the columns of $\mathbf{U}_q$.
To fix the sign of each eigenvector $\mathbf{u}$ without using world axes, we set $z_i = \mathbf{u}^\top(\mathbf{c}_i - \boldsymbol{\mu}_q)$ and orient $\mathbf{u}$ so that $\sum_{i \in q} \pi_i z_i^3 > 0$, or $\sum_{i \in q} \pi_i \operatorname{sign}(z_i) z_i^2 > 0$ if skewness vanishes.
When eigenvalues are degenerate or both odd moments vanish, $\mathbf{U}_q$ falls back to Gram--Schmidt orthogonalization on the $\boldsymbol{\mu}_q$-centered camera and observed-point vectors in input order, padding missing span directions with zeros if the rank is below $3$.

\paragraph{3. Voxel width.}
We set the voxel width $r_q = \beta s_{\mathrm{cam},q}$ ($\beta = 0.02$) from the distances between cameras and the points they observe.
For each camera, we divide its median point distance by its dimensionless focal length $f_{\mathrm{norm},i} = \sqrt{f_{x,i} f_{y,i} / (W_i H_i)}$, giving $d_i = (\operatorname{median}_{\mathbf{x} \in \mathcal{P}_i} \|\mathbf{x} - \mathbf{c}_i\|_2) / f_{\mathrm{norm},i}$. The component scale $s_{\mathrm{cam},q}$ is the midpoint $\pi_i$-weighted median of $\{d_i\}_{i \in q}$.

\paragraph{4. Coverage bins.}
Each point $\mathbf{x} \in \mathcal{P}_i$ maps to the canonical voxel $\lfloor \mathbf{U}_q^\top (\mathbf{x} - \boldsymbol{\mu}_q) / r_q \rfloor$, paired with the signed dominant coordinate ($\pm x, \pm y, \pm z$) of the ray $\mathbf{U}_q^\top(\mathbf{x} - \mathbf{c}_i)$ and tagged by component $q$.
Ties in coordinate magnitude are broken in $(x,y,z)$ order, taking a positive sign at zero.
Let $\mathcal{B}_i$ be camera $i$'s set of distinct bins and $m_b = |\{j : b \in \mathcal{B}_j\}|$ be the number of training cameras observing bin $b$.
The coverage weight $\covw = |\mathcal{B}_i|^{-1} \sum_{b \in \mathcal{B}_i} 1/m_b$ (with $\covw = 0$ if $\mathcal{B}_i = \emptyset$), scene coverage measure $\capsupp = \sum_{i=1}^{N} \covw \sqrt{W_i H_i}$, and exploration cap $\tcap = \operatorname{round}(\kappa \capsupp)$ then follow Eqs.~(1)--(2) of the main paper.

\subsection{Proof of Proposition 1}
\label{sec:supp-coverage-proof}
We show that the scene coverage measure $\capsupp$ is invariant to duplicating all training views (\textbf{P1}), stays bounded as camera bursts grow (\textbf{P2}), adds across visibility-disjoint captures (\textbf{P3}), stays bounded as SfM point density increases (\textbf{P4}), and scales linearly with image dimensions under uniform resizing (\textbf{P5}).
We also show that $\capsupp$ is independent of the reconstruction's origin, orientation, and uniform scale, and of which held-out views are included.

We assume exact arithmetic, positive component scales $s_{\mathrm{cam},q} > 0$, and fixed training observations apart from the transformation under consideration. Under uniform resolution scaling, the intrinsics $(f_{x,i}, f_{y,i})$ scale by the same factor as $(W_i, H_i)$.

\paragraph{Coordinate and similarity invariance.}
Under any similarity transform $\mathbf{x}' = a\mathbf{Q}\mathbf{x} + \mathbf{t}$ and $\mathbf{c}_i' = a\mathbf{Q}\mathbf{c}_i + \mathbf{t}$ with $a > 0$ and orthogonal $\mathbf{Q}$ ($\mathbf{Q}^\top \mathbf{Q} = \mathbf{I}$, including reflections), the point multiplicities $t_p$ and normalized camera weights $\pi_i$ do not change, while $\boldsymbol{\mu}_q' = a\mathbf{Q}\boldsymbol{\mu}_q + \mathbf{t}$ and $\mathbf{C}_q' = a^2 \mathbf{Q}\mathbf{C}_q\mathbf{Q}^\top$.
The signed PCA axes transform equivariantly as $\mathbf{U}_q' = \mathbf{Q}\mathbf{U}_q$, as do the ordered Gram--Schmidt fallback axes since they use no cross products.
As a result, the canonical point coordinates $\mathbf{U}_q'^\top(\mathbf{x}' - \boldsymbol{\mu}_q') = a \mathbf{U}_q^\top(\mathbf{x} - \boldsymbol{\mu}_q)$, canonical viewing rays, and scale $s_{\mathrm{cam},q}' = a s_{\mathrm{cam},q}$ all multiply by $a$, so the voxel indices, signed dominant ray axes, $\mathcal{B}_i$, $m_b$, $\covw$, and $\capsupp$ remain unchanged.

\paragraph{Duplication invariance and disjoint additivity (P1, P3).}
Duplicating every training view $k$ times sets $t_p' = k t_p$ and $\widetilde{a}_i' = \widetilde{a}_i / k$, so each copy receives weight $\pi_i/k$. The $k$ copies retain total weight $\pi_i$, preserving $\boldsymbol{\mu}_q$, $\mathbf{U}_q$, $s_{\mathrm{cam},q}$, and each view's bin set $\mathcal{B}_i$.
The bin counts become $m_b' = k m_b$, so $\covw' = \covw / k$ and $\capsupp' = k \sum_{i=1}^{N} (\covw / k) \sqrt{W_i H_i} = \capsupp$ (\textbf{P1}).
For visibility-disjoint captures $A \sqcup B$, connected components do not interact and bin keys include the component ID, so $\capsupp(A \sqcup B) = \capsupp(A) + \capsupp(B)$ (\textbf{P3}).

\paragraph{Burst and SfM point-density boundedness (P2, P4).}
Letting $L_i = \sqrt{W_i H_i} \le L_{\max}$ and exchanging summation order over $\mathcal{B} = \bigcup_i \mathcal{B}_i$ gives
\begin{equation}
    \capsupp = \sum_{b \in \mathcal{B}} \frac{1}{m_b} \sum_{i : b \in \mathcal{B}_i} \frac{L_i}{|\mathcal{B}_i|} \le L_{\max} |\mathcal{B}|.
\end{equation}
If cameras and observed points lie within radius $D$ of the origin and normalized camera distances satisfy $d_i \ge s_{\min} > 0$ (so $s_{\mathrm{cam},q} \ge s_{\min}$), then $\|\boldsymbol{\mu}_q\|_2 \le D$ and $\|\mathbf{U}_q^\top(\mathbf{x} - \boldsymbol{\mu}_q)\|_\infty \le 2D$, so each component spans at most $6\bigl(2\lceil 2D / (\beta s_{\min}) \rceil + 1\bigr)^3$ directional voxel bins regardless of burst size $B$.
Because burst views observe already-observed regions and attach to existing components without increasing the component count, summing over components keeps $\capsupp = O(1)$ as $B \to \infty$ even when frames and scales are recomputed (\textbf{P2}).
Similarly, $0 \le \covw \le 1$ gives $\capsupp \le \sum_{i=1}^{N} L_i$ regardless of the SfM point count $M$, and $\capsupp$ becomes constant once $\mathbf{U}_q$, $s_{\mathrm{cam},q}$, and the occupied bin sets $\mathcal{B}_i$ settle (\textbf{P4}).

\paragraph{Resolution scaling and test-split invariance (P5).}
Scaling $(W_i, H_i)$ and $(f_{x,i}, f_{y,i})$ by $\alpha > 0$ keeps $f_{\mathrm{norm},i}$, the canonical bins, and $\covw$ fixed while $L_i' = \alpha L_i$, so $\capsupp(\alpha W, \alpha H) = \alpha \capsupp(W, H)$ (\textbf{P5}).
Finally, because $\mathcal{E}_{\mathrm{train}}$ excludes held-out views, adding or removing test views does not affect $\capsupp$. \hfill$\square$

\subsection{Leave-one-out alpha-compositing derivation}
At pixel $p$ of view $i$, we order the primitives intersecting $p$ front-to-back and split the compositing sum around primitive $g$.
Let $\hat{\mathbf{C}}_{i,<g}(p)$ be the color accumulated strictly in front of $g$, $T_g(p) = \prod_{k < g} (1 - \alpha_k(p))$ the transmittance reaching $g$, and $\hat{\mathbf{C}}_{i,>g}(p)$ the color composited strictly behind $g$ (including the background) under unit incoming transmittance.
The full teacher color and the color rendered with $g$ removed are
\begin{equation}
    \hat{\mathbf{C}}_i(p) = \hat{\mathbf{C}}_{i,<g}(p) + T_g(p)\bigl[\alpha_g(p)\mathbf{c}_g(i) + (1 - \alpha_g(p))\hat{\mathbf{C}}_{i,>g}(p)\bigr],
    \qquad
    \hat{\mathbf{C}}_{i,\setminus g}(p) = \hat{\mathbf{C}}_{i,<g}(p) + T_g(p)\hat{\mathbf{C}}_{i,>g}(p).
\end{equation}
Subtracting the two eliminates $\hat{\mathbf{C}}_{i,<g}(p)$ and gives the exact per-pixel leave-one-out color change to all orders in $\alpha_g(p)$:
\begin{equation}
    \Delta \hat{\mathbf{C}}_{g,i}(p) = \hat{\mathbf{C}}_i(p) - \hat{\mathbf{C}}_{i,\setminus g}(p) = T_g(p)\alpha_g(p)\bigl(\mathbf{c}_g(i) - \hat{\mathbf{C}}_{i,>g}(p)\bigr).
\end{equation}

\subsection{Implementation details in LiteGS}
We build \method{} on LiteGS~\citep{litegs}, which runs density control at whole-epoch boundaries ($\lfloor \text{iterations}/N \rfloor$ epochs).
During the growth phase (first $80\%$ of training), budgeted densification grows toward $\tcap$ with global exploration scaling factor $\kappa = 62.04$, chosen so the Std-13 geometric-mean exploration cap equals $2\bref$ against the reference mean budget $\bref = 0.925$M.
Growth samples parents using opacity-gradient variance scores with a clone-versus-split 3D scale threshold of $10^{-3}$ of scene extent, a learning-rate divisor of $20$ for higher-order spherical harmonics, multiplicative opacity decay of $0.5$ (with a floor of $1/128$) instead of hard resets, and periodic soft prunes retaining $\krho_0 = 0.60$ of pre-growth primitives, producing the teacher model $\mathcal{G}_{\mathrm{pre}}$ of size $\npre = |\mathcal{G}_{\mathrm{pre}}|$.
At the $80\%$ epoch boundary, a single frozen pass over the training views accumulates each primitive's leave-one-out RGB squared error $u_{g,i} = \sum_p \|\Delta \hat{\mathbf{C}}_{g,i}(p)\|_2^2$ and computes utility scores $\util = \sum_{i=1}^{N} (\covw / \sqrt{W_i H_i}) u_{g,i}$.
Let $n = |\{g : \util > \lamstar\}|$ be the number of primitives above $\lamstar = 1.0861 \times 10^{-5}$, which we calibrate once by bisection on seed-4114 Std-13 scores to match the mean aligned count $\bref = 12{,}026{,}752 / 13 \approx 0.925$M.
Because LiteGS groups primitives into $128$-Gaussian CUDA clusters, we round $n$ up to a multiple of $128$ with a $10\%$ safety floor (never triggered in our default benchmark runs):
\begin{equation}
    \numg = \min\!\left(\npre,\; 128\left\lceil \frac{\max(n, 0.10\npre)}{128} \right\rceil\right).
\end{equation}
We keep the top $\numg$ primitives and their Adam states, rebuild the spatial clusters and bounding boxes, and run the final $20\%$ refinement phase without further densification or pruning.

\subsection{Dataset preparation and evaluation protocol}
Standard scenes (Std-13) train for $30$k iterations with every eighth sorted frame held out for testing.
On Mip-NeRF~360~\citep{barron2022mipnerf360}, we use $1/2$-resolution indoor images and $1/4$-resolution outdoor images; Tanks~\&~Temples~\citep{knapitsch2017tanks} and Deep Blending~\citep{hedman2018deepblending} use their default benchmark resolutions.
Large captures (Large-8) train for $60$k iterations with position learning-rate decay scaled to $60$k.
Mill19~\citep{turki2022meganerf} and UrbanScene3D~\citep{lin2022urbanscene3d} use their official validation splits, with training-only COLMAP on \emph{Rubble} (where the original sparse model is unavailable) and the prepared images on \emph{Residence} and \emph{Sci-Art}.
On Zip-NeRF~\citep{barron2023zipnerf}, we hold out every eighth sorted frame at $1/2$ resolution, capped at $1600$\,px width.
All compared methods use the same training and test images and the same sparse reconstructions.

Unless noted otherwise, \method{} results average seeds 4114, 4115, and 4116 for each capture before averaging equally across captures.
Cross-method tables report PSNR, TorchMetrics SSIM, and unnormalized VGG LPIPS from a shared evaluator.
The Taming~3DGS port (\cref{sec:supp-second-base}) uses Taming~3DGS's reference PNG evaluator for parity, and internal LiteGS ablations noted in table captions report LiteGS's native Gaussian-window SSIM.
All 13 baselines run with their released density-control defaults.
We extend their dataset loaders only to read shared \texttt{SIMPLE\_RADIAL}/\texttt{RADIAL} pinhole intrinsics and official \texttt{test.txt} splits; in addition, LP-3DGS uses its restored top-$k$ rasterizer and CDC-GS includes a guard for empty candidate sets.

\clearpage
\section{Standard-Benchmark Results and Reproduction}
\label{sec:supp-adaptive}\label{sec:supp-reproduction}\label{sec:supp-perscene}

\subsection{Cross-scale overview}
\Cref{tab:main-benchmark} reports mean reconstruction quality, final Gaussian count $\numg$, saved storage, and rendering speed on Std-13 and Large-8 for all 13 baselines, \method{}, its compact $4\lamstar$ variant, and our Taming~3DGS port (see \cref{sec:supp-capacity-objective} for the full $\lambda \in \{8, 4, 2, 1, 0.5\}\lamstar$ sweep).
On Large-8, 3DGS-MCMC drops to $17.02$\,dB because its released outdoor $L_1$ opacity penalty ($0.01$) drives most Gaussians transparent over $60$k iterations.
Scaling its position learning-rate schedule alone leaves \emph{NYC} at $14.49$\,dB, whereas lowering the opacity penalty $10\times$ to its Deep Blending preset ($0.001$) narrows the PSNR gap to vanilla 3DGS to $0.36$\,dB on \emph{NYC} and $1.44$\,dB on \emph{Sci-Art}.

\begin{table}[!htbp]
\centering
\caption{\textbf{Quantitative comparison across 13 standard scenes ($\mathbf{30}$k) and 8 large captures without retuning ($\mathbf{60}$k).} All 13 baselines use released configurations on one NVIDIA GH200 (Taming~3DGS uses dataset presets on standard scenes and $15\times$ SfM points on large captures; Aerial averages Mill19 and UrbanScene3D; Zip averages the four Zip-NeRF captures). \textbf{Bold} and \underline{underlining} mark the best and second-best quality among benchmark-evaluator models. We report \method{} ($\lamstar$, three-seed mean), a compact variant ($4\lamstar$, seed 4114), and \method{} (Taming~3DGS)$^{\dagger}$ (controller port, seed 4114, reference PNG evaluator matching Taming~3DGS; \cref{sec:supp-second-base}).}
\label{tab:main-benchmark}
\footnotesize
\setlength{\tabcolsep}{3.1pt}
\begin{tabular}{@{}l rrrrrr rrrrrrrr@{}}
\toprule
 & \multicolumn{6}{c}{\textbf{Standard Benchmarks (Std-13, 13 scenes)}} & \multicolumn{8}{c}{\textbf{Large Captures Without Retuning (Large-8, 8 scenes)}} \\
\cmidrule(lr){2-7}\cmidrule(lr){8-15}
Method & PSNR$\uparrow$ & SSIM$\uparrow$ & LPIPS$\downarrow$ & $\numg$(M)$\downarrow$ & MB$\downarrow$ & FPS$\uparrow$ & Aerial$\uparrow$ & Zip$\uparrow$ & PSNR$\uparrow$ & SSIM$\uparrow$ & LPIPS$\downarrow$ & $\numg$(M)$\downarrow$ & MB$\downarrow$ & FPS$\uparrow$ \\
\midrule
\multicolumn{15}{@{}l}{\emph{Fixed-constant controllers}} \\
vanilla 3DGS~\citep{kerbl3Dgaussians} & 27.30 & 0.836 & 0.209 & 2.52 & 625 & 203 & 22.09 & 25.95 & 24.02 & 0.790 & 0.312 & 3.08 & 764 & 216 \\
Speedy-Splat~\citep{hanson2025speedysplat} & 26.78 & 0.809 & 0.274 & 0.29 & 71 & 1047 & 20.43 & 24.65 & 22.54 & 0.713 & 0.416 & 0.24 & 59 & 1163 \\
\midrule
\multicolumn{15}{@{}l}{\emph{Preset-count controllers}} \\
3DGS-MCMC~\citep{kheradmand2024mcmc} & 27.69 & \underline{0.851} & \textbf{0.186} & 2.88 & 715 & 187 & 17.92 & 16.12 & 17.02 & 0.576 & 0.554 & 3.08 & 764 & 549 \\
Taming 3DGS~\citep{tamings3dgs} & 27.63 & 0.841 & 0.203 & 2.93 & 728 & 220 & 22.07 & 25.48 & 23.77 & 0.764 & 0.346 & 1.44 & 356 & 329 \\
Mini-Splatting~\citep{fang2024minisplatting} & 27.06 & 0.836 & 0.219 & 0.42 & 106 & 806 & 19.91 & 24.48 & 22.19 & 0.693 & 0.434 & 0.19 & 47 & 1074 \\
LiteGS (1M)~\citep{litegs} & 27.92 & 0.842 & 0.203 & 0.99 & 246 & 538 & 22.02 & \underline{26.20} & 24.11 & 0.778 & 0.322 & 1.41 & 350 & 601 \\
\midrule
\multicolumn{15}{@{}l}{\emph{Automatic-budget controllers}} \\
DashGaussian~\citep{chen2025dashgaussian} & 27.45 & 0.838 & 0.210 & 2.17 & 538 & 249 & 21.29 & 25.86 & 23.57 & 0.772 & 0.327 & 2.94 & 729 & 285 \\
Compact 3DGS~\citep{lee2024compact} & 26.89 & 0.822 & 0.234 & 1.28 & 122 & 277 & 20.16 & 24.53 & 22.34 & 0.715 & 0.405 & 0.57 & 83 & 553 \\
LP-3DGS~\citep{zhang2024lp3dgs} & 27.18 & 0.834 & 0.216 & 1.13 & 282 & 467 & 21.34 & 25.28 & 23.31 & 0.754 & 0.360 & 0.66 & 165 & 706 \\
ControlGS~\citep{controlgs2025} & 27.46 & 0.840 & 0.212 & 0.79 & 195 & 161 & 20.79 & 24.97 & 22.88 & 0.724 & 0.391 & 0.49 & 122 & 193 \\
CDC-GS~\citep{dong2025cdcgs} & 27.52 & 0.845 & \underline{0.191} & 2.42 & 600 & 218 & 21.96 & 26.00 & 23.98 & 0.784 & 0.316 & 3.17 & 786 & 225 \\
FastGS~\citep{ren2026fastgs} & 27.35 & 0.821 & 0.251 & 0.33 & 82 & 993 & 21.02 & 24.93 & 22.98 & 0.739 & 0.380 & 0.39 & 98 & 1215 \\
LeGS~\citep{ning2026legs} & \textbf{28.08} & \textbf{0.853} & \textbf{0.186} & 1.79 & 445 & 520 & 21.30 & 25.52 & 23.41 & 0.758 & 0.348 & 0.94 & 233 & 881 \\
\midrule
\multicolumn{15}{@{}l}{\emph{Coverage-scaled exploration + self-distillation (ours)}} \\
\textbf{\method{} (Taming 3DGS)}$^{\dagger}$ & 27.77 & 0.846 & 0.190 & 1.03 & 255 & 250 & 22.78 & 26.71 & 24.74 & 0.824 & 0.255 & 3.43 & 852 & 182 \\
\textbf{\method{} ($4\lamstar$)} & 27.72 & 0.842 & 0.216 & 0.56 & 140 & 699 & \underline{22.67} & \underline{26.20} & \underline{24.44} & \underline{0.804} & \underline{0.289} & 1.89 & 468 & 507 \\
\textbf{\method{}} & \underline{27.99} & 0.850 & 0.195 & 0.93 & 230 & 584 & \textbf{22.90} & \textbf{26.58} & \textbf{24.74} & \textbf{0.821} & \textbf{0.258} & 3.91 & 969 & 366 \\
\bottomrule
\end{tabular}
\end{table}

\FloatBarrier

\subsection{Dataset and per-scene breakdowns}
\begin{figure}[!htbp]
    \centering
    \includegraphics[width=\linewidth]{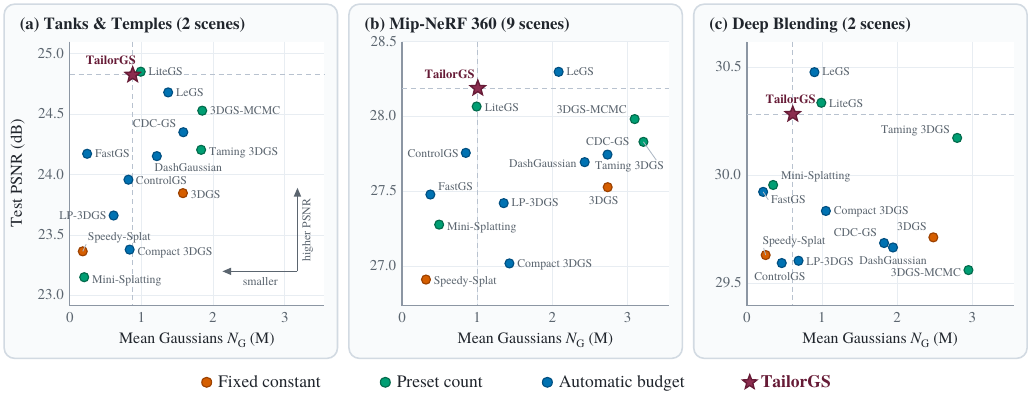}
    \caption{\textbf{Standard-benchmark quality--size trade-offs by dataset.} Dataset means on \textbf{(a)}~Tanks~\&~Temples (2 scenes), \textbf{(b)}~Mip-NeRF~360 (9 scenes), and \textbf{(c)}~Deep Blending (2 scenes), colored by controller class.}
    \label{fig:supp-methods}
\end{figure}

\Cref{fig:supp-methods} and \cref{tab:supp-quality} break down results on the 13 standard scenes across Mip-NeRF~360, Tanks~\&~Temples, and Deep Blending. \method{} lies on the quality--size Pareto front on all three datasets.
\Cref{tab:supp-support-seeds-standard} reports three-seed \method{} results alongside preset LiteGS ($1$M) on each of the 13 standard scenes, while per-scene comparisons against all 13 baselines appear in \cref{sec:supp-efficiency}.

\begin{table}[!htbp]
\centering
\caption{\textbf{Dataset-level comparison across the 13 standard scenes ($\mathbf{30}$k iterations).} Equal-scene means across Mip-NeRF~360 (9 scenes), Tanks~\&~Temples (2 scenes), and Deep Blending (2 scenes); \method{} reports three-seed means. \textbf{Bold} and \underline{underlining} mark best and second-best quality.}
\label{tab:supp-quality}\label{tab:supp-quality-tnt}\label{tab:supp-quality-db}
\footnotesize
\setlength{\tabcolsep}{3.4pt}
\begin{tabular}{@{}l rrrr rrrr rrrr@{}}
\toprule
 & \multicolumn{4}{c}{\textbf{Mip-NeRF 360 (9)}} & \multicolumn{4}{c}{\textbf{Tanks \& Temples (2)}} & \multicolumn{4}{c}{\textbf{Deep Blending (2)}} \\
\cmidrule(lr){2-5}\cmidrule(lr){6-9}\cmidrule(lr){10-13}
Method & PSNR$\uparrow$ & SSIM$\uparrow$ & LPIPS$\downarrow$ & $\numg$(M) & PSNR$\uparrow$ & SSIM$\uparrow$ & LPIPS$\downarrow$ & $\numg$(M) & PSNR$\uparrow$ & SSIM$\uparrow$ & LPIPS$\downarrow$ & $\numg$(M) \\
\midrule
\multicolumn{13}{@{}l}{\emph{Fixed-constant controllers}} \\
vanilla 3DGS~\citep{kerbl3Dgaussians} & 27.53 & 0.816 & 0.211 & 2.74 & 23.85 & 0.853 & 0.168 & 1.58 & 29.71 & 0.907 & 0.239 & 2.48 \\
Speedy-Splat~\citep{hanson2025speedysplat} & 26.91 & 0.786 & 0.283 & 0.32 & 23.37 & 0.819 & 0.241 & 0.18 & 29.63 & 0.903 & 0.268 & 0.25 \\
\midrule
\multicolumn{13}{@{}l}{\emph{Preset-count controllers}} \\
3DGS-MCMC~\citep{kheradmand2024mcmc} & 27.98 & \underline{0.835} & \underline{0.185} & 3.10 & 24.53 & \textbf{0.871} & 0.146 & 1.85 & 29.56 & 0.905 & 0.233 & 2.95 \\
Taming 3DGS~\citep{tamings3dgs} & 27.83 & 0.822 & 0.204 & 3.21 & 24.21 & 0.855 & 0.169 & 1.83 & 30.17 & 0.910 & 0.234 & 2.80 \\
Mini-Splatting~\citep{fang2024minisplatting} & 27.28 & 0.821 & 0.213 & 0.49 & 23.15 & 0.831 & 0.210 & 0.20 & 29.96 & 0.908 & 0.253 & 0.35 \\
LiteGS (1M)~\citep{litegs} & 28.07 & 0.821 & 0.210 & 0.99 & \textbf{24.85} & \underline{0.868} & \underline{0.144} & 0.99 & \underline{30.34} & 0.911 & \underline{0.232} & 0.99 \\
\midrule
\multicolumn{13}{@{}l}{\emph{Automatic-budget controllers}} \\
DashGaussian~\citep{chen2025dashgaussian} & 27.70 & 0.820 & 0.209 & 2.43 & 24.15 & 0.853 & 0.178 & 1.22 & 29.67 & 0.905 & 0.246 & 1.95 \\
Compact 3DGS~\citep{lee2024compact} & 27.02 & 0.800 & 0.238 & 1.43 & 23.38 & 0.835 & 0.199 & 0.84 & 29.84 & 0.905 & 0.254 & 1.05 \\
LP-3DGS~\citep{zhang2024lp3dgs} & 27.42 & 0.815 & 0.216 & 1.35 & 23.66 & 0.845 & 0.185 & 0.61 & 29.61 & 0.906 & 0.248 & 0.69 \\
ControlGS~\citep{controlgs2025} & 27.76 & 0.823 & 0.213 & 0.85 & 23.96 & 0.853 & 0.166 & 0.82 & 29.60 & 0.906 & 0.254 & 0.46 \\
CDC-GS~\citep{dong2025cdcgs} & 27.74 & 0.826 & 0.191 & 2.73 & 24.35 & 0.866 & 0.149 & 1.59 & 29.69 & 0.907 & 0.234 & 1.83 \\
FastGS~\citep{ren2026fastgs} & 27.48 & 0.798 & 0.257 & 0.38 & 24.17 & 0.843 & 0.208 & 0.24 & 29.92 & 0.904 & 0.267 & 0.22 \\
LeGS~\citep{ning2026legs} & \textbf{28.30} & \textbf{0.837} & \textbf{0.183} & 2.08 & 24.68 & 0.866 & 0.155 & 1.37 & \textbf{30.48} & \textbf{0.914} & \textbf{0.229} & 0.90 \\
\midrule
\multicolumn{13}{@{}l}{\emph{Coverage-scaled exploration + self-distillation (ours)}} \\
\textbf{\method{}} & \underline{28.19} & 0.832 & 0.197 & 1.01 & \underline{24.83} & \textbf{0.871} & \textbf{0.143} & 0.88 & 30.28 & \underline{0.912} & 0.235 & 0.61 \\
\bottomrule
\end{tabular}
\end{table}

\begin{table}[!htbp]
\centering
\caption{\textbf{Per-scene standard-benchmark results and three-seed stability.} \method{} reports mean $\pm$ std across seeds 4114, 4115, and 4116 alongside preset LiteGS ($1$M) under identical evaluation metrics; per-scene comparisons across all 13 baselines appear in \cref{sec:supp-efficiency}.}
\label{tab:supp-support-seeds-standard}\label{tab:supp-perscene-main}
\footnotesize
\setlength{\tabcolsep}{4.2pt}
\begin{tabular}{@{}l rrrr rrrr@{}}
\toprule
 & \multicolumn{4}{c}{\textbf{\method{} (3 seeds, mean $\pm$ std)}} & \multicolumn{4}{c}{\textbf{LiteGS (1M preset)}} \\
\cmidrule(lr){2-5}\cmidrule(lr){6-9}
Capture & $\numg$ (M) & PSNR (dB)$\uparrow$ & SSIM$\uparrow$ & LPIPS$\downarrow$ & $\numg$ (M) & PSNR (dB)$\uparrow$ & SSIM$\uparrow$ & LPIPS$\downarrow$ \\
\midrule
Bicycle & 1.42 $\pm$ 0.00 & \textbf{25.73} $\pm$ 0.05 & \textbf{0.788} $\pm$ 0.000 & \textbf{0.198} $\pm$ 0.000 & 0.99 & 24.84 & 0.736 & 0.263 \\
Bonsai & 0.65 $\pm$ 0.01 & 33.00 $\pm$ 0.12 & 0.949 $\pm$ 0.001 & 0.187 $\pm$ 0.000 & 0.99 & \textbf{33.17} & \textbf{0.951} & \textbf{0.172} \\
Counter & 0.65 $\pm$ 0.01 & 29.71 $\pm$ 0.03 & \textbf{0.919} $\pm$ 0.001 & 0.178 $\pm$ 0.000 & 0.99 & \textbf{29.84} & \textbf{0.919} & \textbf{0.170} \\
Flowers & 1.45 $\pm$ 0.00 & \textbf{21.96} $\pm$ 0.02 & \textbf{0.639} $\pm$ 0.000 & \textbf{0.285} $\pm$ 0.000 & 0.99 & 21.71 & 0.620 & 0.315 \\
Garden & 0.93 $\pm$ 0.00 & \textbf{27.65} $\pm$ 0.04 & \textbf{0.860} $\pm$ 0.000 & \textbf{0.131} $\pm$ 0.000 & 0.99 & 27.55 & 0.857 & 0.134 \\
Kitchen & 0.81 $\pm$ 0.01 & \textbf{32.46} $\pm$ 0.11 & \textbf{0.935} $\pm$ 0.001 & 0.117 $\pm$ 0.001 & 0.99 & 32.37 & 0.932 & \textbf{0.114} \\
Room & 0.41 $\pm$ 0.01 & 32.51 $\pm$ 0.06 & 0.926 $\pm$ 0.001 & 0.207 $\pm$ 0.002 & 0.99 & \textbf{32.63} & \textbf{0.931} & \textbf{0.183} \\
Stump & 1.23 $\pm$ 0.00 & \textbf{27.51} $\pm$ 0.01 & \textbf{0.811} $\pm$ 0.000 & \textbf{0.184} $\pm$ 0.000 & 1.00 & 27.13 & 0.791 & 0.214 \\
Treehill & 1.52 $\pm$ 0.00 & 23.17 $\pm$ 0.08 & \textbf{0.660} $\pm$ 0.001 & \textbf{0.289} $\pm$ 0.001 & 0.99 & \textbf{23.37} & 0.651 & 0.323 \\
Truck & 0.69 $\pm$ 0.00 & \textbf{26.60} $\pm$ 0.02 & 0.895 $\pm$ 0.000 & 0.124 $\pm$ 0.000 & 0.99 & 26.57 & \textbf{0.896} & \textbf{0.113} \\
Train & 1.07 $\pm$ 0.00 & 23.05 $\pm$ 0.02 & \textbf{0.848} $\pm$ 0.001 & \textbf{0.162} $\pm$ 0.001 & 0.99 & \textbf{23.14} & 0.839 & 0.174 \\
Dr Johnson & 0.91 $\pm$ 0.00 & \textbf{29.90} $\pm$ 0.06 & \textbf{0.912} $\pm$ 0.000 & \textbf{0.229} $\pm$ 0.001 & 0.99 & 29.79 & 0.909 & 0.237 \\
Playroom & 0.31 $\pm$ 0.00 & 30.67 $\pm$ 0.20 & 0.912 $\pm$ 0.001 & 0.241 $\pm$ 0.000 & 0.99 & \textbf{30.88} & \textbf{0.913} & \textbf{0.226} \\
\midrule
\textbf{Std-13 mean} & 0.93 $\pm$ 0.00 & \textbf{27.99} $\pm$ 0.03 & \textbf{0.850} $\pm$ 0.000 & \textbf{0.195} $\pm$ 0.000 & 0.99 & 27.92 & 0.842 & 0.203 \\
\bottomrule
\end{tabular}
\end{table}

\FloatBarrier

\subsection{Reproduction against published results}
\Cref{tab:supp-reproduction} compares our baseline reruns against the PSNR values reported in the original papers; LiteGS is omitted because its paper reports individual scene scores rather than dataset means across all three benchmarks.
Nine of the twelve baselines match their published dataset means within $0.35$\,dB everywhere they report, including LeGS within $0.13$\,dB under its released evaluation setup.
Our reruns are slightly higher than published numbers on vanilla 3DGS ($+0.30$ to $+0.71$\,dB) and DashGaussian ($+0.38$\,dB on Deep Blending), and $0.39$\,dB below published ControlGS on Tanks~\&~Temples.

\begin{table}[!htbp]
\centering
\caption{\textbf{Baseline reproduction against published PSNR.} $\Delta$ denotes our rerun minus the value reported in each method's paper ($|\Delta|\ge 0.35$\,dB in \textbf{bold}). $^{a}$Excludes \emph{Flowers} and \emph{Treehill} to match the 7-scene split reported by 3DGS-MCMC. $^{b}$CDC-GS reports three-seed means against a $+0.26$\,dB higher Mip-NeRF~360 reference.}
\label{tab:supp-reproduction}
\footnotesize
\setlength{\tabcolsep}{4.5pt}
\begin{tabular}{@{}l rrr rrr rrr@{}}
\toprule
 & \multicolumn{3}{c}{\textbf{Mip-NeRF 360}} & \multicolumn{3}{c}{\textbf{Tanks \& Temples}} & \multicolumn{3}{c}{\textbf{Deep Blending}} \\
\cmidrule(lr){2-4}\cmidrule(lr){5-7}\cmidrule(lr){8-10}
Method & Rerun & Pub. & $\Delta$ & Rerun & Pub. & $\Delta$ & Rerun & Pub. & $\Delta$ \\
\midrule
vanilla 3DGS & 27.53 & 27.21 & +0.32 & 23.85 & 23.14 & \textbf{+0.71} & 29.71 & 29.41 & +0.30 \\
Speedy-Splat & 26.91 & 26.94 & -0.03 & 23.37 & 23.45 & -0.08 & 29.63 & 29.32 & +0.31 \\
3DGS-MCMC$^{a}$ & 29.55 & 29.89 & -0.34 & 24.53 & 24.29 & +0.24 & 29.56 & 29.67 & -0.11 \\
Taming 3DGS & 27.83 & 27.79 & +0.04 & 24.21 & 24.04 & +0.17 & 30.17 & 30.14 & +0.03 \\
Mini-Splatting & 27.28 & 27.34 & -0.06 & 23.15 & 23.18 & -0.03 & 29.96 & 29.98 & -0.02 \\
DashGaussian & 27.70 & 27.81 & -0.11 & 24.15 & 23.83 & +0.32 & 29.67 & 29.29 & \textbf{+0.38} \\
Compact 3DGS & 27.02 & 27.08 & -0.06 & 23.38 & 23.32 & +0.06 & 29.84 & 29.79 & +0.05 \\
LP-3DGS & 27.42 & 27.47 & -0.05 & 23.66 & 23.60 & +0.06 & 29.61 & -- & -- \\
ControlGS & 27.76 & 27.90 & -0.14 & 23.96 & 24.35 & \textbf{-0.39} & 29.60 & 29.81 & -0.21 \\
CDC-GS$^{b}$ & 27.74 & 28.02 & -0.28 & 24.35 & 24.42 & -0.07 & 29.69 & 29.84 & -0.15 \\
FastGS & 27.48 & 27.56 & -0.08 & 24.17 & 24.15 & +0.02 & 29.92 & 30.03 & -0.11 \\
LeGS & 28.30 & 28.30 & -0.00 & 24.68 & 24.74 & -0.06 & 30.48 & 30.35 & +0.13 \\
\bottomrule
\end{tabular}
\end{table}

\clearpage
\section{Large-Capture Results}
\label{sec:supp-transfer}

\begin{figure}[!htbp]
    \centering
    \includegraphics[width=\linewidth]{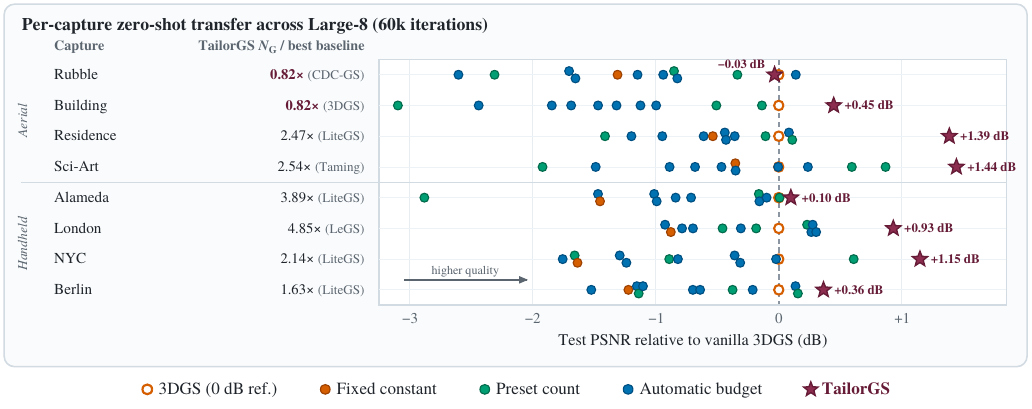}
    \caption{\textbf{Per-capture transfer to large captures without retuning.} Test PSNR relative to vanilla 3DGS ($60$k); left-hand ratios compare \method{}'s Gaussian count to the highest-PSNR baseline on each scene. Speedy-Splat on \emph{Building} ($-4.43$\,dB) and 3DGS-MCMC are omitted for axis scale.}
    \label{fig:supp-transfer}
\end{figure}

\Cref{fig:supp-transfer} summarizes per-capture PSNR gains and model-size ratios on the eight large captures at $60$k iterations without retuning.
\Cref{tab:supp-transfer-mill19,tab:supp-transfer-zipnerf_a} give per-capture metrics for the four aerial captures (Mill19 and UrbanScene3D) and four handheld multi-room interiors (Zip-NeRF), and \cref{tab:supp-support-seeds-large} reports \method{}'s three-seed results on each capture.
With the Std-13 calibration kept unchanged, \method{} ranks first in PSNR and LPIPS on seven of the eight captures; on \emph{Rubble}, it trails CDC-GS and vanilla 3DGS slightly in PSNR while using $18\%$--$19\%$ fewer Gaussians.
Across the three seeds, the standard deviations of its Large-8 mean count and mean PSNR are $0.01$M and $0.03$\,dB.

\begin{table}[!htbp]
\centering
\caption{\textbf{Per-capture comparison on large aerial captures ($\mathbf{60}$k iterations, without retuning).} Mill19 (\emph{Rubble}, \emph{Building}) and UrbanScene3D (\emph{Residence}, \emph{Sci-Art}). Counts $\numg$ are in millions; Taming~3DGS uses its $15\times$ large-capture preset, and \method{} reports three-seed means.}
\label{tab:supp-transfer-mill19}\label{tab:supp-transfer-urban}
\scriptsize
\setlength{\tabcolsep}{2.1pt}
\begin{tabular}{@{}l rrrr rrrr rrrr rrrr@{}}
\toprule
 & \multicolumn{4}{c}{\textbf{Rubble}} & \multicolumn{4}{c}{\textbf{Building}} & \multicolumn{4}{c}{\textbf{Residence}} & \multicolumn{4}{c}{\textbf{Sci-Art}} \\
\cmidrule(lr){2-5}\cmidrule(lr){6-9}\cmidrule(lr){10-13}\cmidrule(lr){14-17}
Method & PSNR$\uparrow$ & SSIM$\uparrow$ & LPIPS$\downarrow$ & $\numg$ & PSNR$\uparrow$ & SSIM$\uparrow$ & LPIPS$\downarrow$ & $\numg$ & PSNR$\uparrow$ & SSIM$\uparrow$ & LPIPS$\downarrow$ & $\numg$ & PSNR$\uparrow$ & SSIM$\uparrow$ & LPIPS$\downarrow$ & $\numg$ \\
\midrule
\multicolumn{17}{@{}l}{\emph{Fixed-constant controllers}} \\
vanilla 3DGS~\citep{kerbl3Dgaussians} & \underline{25.51} & \textbf{0.783} & \textbf{0.268} & 6.00 & \underline{21.20} & \underline{0.744} & \underline{0.277} & 7.42 & 20.70 & 0.724 & 0.293 & 5.32 & 20.96 & 0.784 & 0.268 & 2.58 \\
Speedy-Splat~\citep{hanson2025speedysplat} & 24.20 & 0.683 & 0.404 & 0.58 & 16.77 & 0.516 & 0.531 & 0.19 & 20.16 & 0.668 & 0.375 & 0.53 & 20.61 & 0.741 & 0.345 & 0.31 \\
\midrule
\multicolumn{17}{@{}l}{\emph{Preset-count controllers}} \\
3DGS-MCMC~\citep{kheradmand2024mcmc} & 22.48 & 0.585 & 0.507 & 6.00 & 18.26 & 0.531 & 0.526 & 7.42 & 19.14 & 0.604 & 0.439 & 5.32 & 11.81 & 0.420 & 0.669 & 2.58 \\
Taming 3DGS~\citep{tamings3dgs} & 25.18 & 0.730 & 0.334 & 2.89 & 20.69 & 0.679 & 0.348 & 2.11 & 20.59 & 0.709 & 0.315 & 2.93 & \underline{21.83} & 0.785 & 0.267 & 2.05 \\
Mini-Splatting~\citep{fang2024minisplatting} & 23.20 & 0.620 & 0.468 & 0.23 & 18.10 & 0.507 & 0.555 & 0.09 & 19.28 & 0.608 & 0.445 & 0.15 & 19.04 & 0.666 & 0.432 & 0.13 \\
LiteGS (1M)~\citep{litegs} & 24.66 & 0.718 & 0.349 & 1.69 & 21.06 & 0.698 & 0.333 & 1.60 & \underline{20.81} & 0.707 & 0.316 & 1.95 & 21.56 & 0.785 & \underline{0.258} & 2.13 \\
\midrule
\multicolumn{17}{@{}l}{\emph{Automatic-budget controllers}} \\
DashGaussian~\citep{chen2025dashgaussian} & 24.57 & 0.758 & 0.292 & 6.13 & 19.36 & 0.662 & 0.355 & 4.31 & 20.26 & 0.709 & 0.304 & 6.11 & 20.96 & 0.780 & 0.272 & 3.09 \\
Compact 3DGS~\citep{lee2024compact} & 22.91 & 0.656 & 0.408 & 1.42 & 18.76 & 0.573 & 0.469 & 0.51 & 19.50 & 0.643 & 0.396 & 1.23 & 19.47 & 0.713 & 0.370 & 0.65 \\
LP-3DGS~\citep{zhang2024lp3dgs} & 24.69 & 0.734 & 0.333 & 1.63 & 19.73 & 0.630 & 0.409 & 0.60 & 20.34 & 0.696 & 0.331 & 1.32 & 20.61 & 0.763 & 0.301 & 0.86 \\
ControlGS~\citep{controlgs2025} & 23.81 & 0.661 & 0.411 & 0.81 & 19.51 & 0.603 & 0.430 & 0.45 & 19.75 & 0.641 & 0.391 & 0.82 & 20.07 & 0.725 & 0.349 & 0.65 \\
CDC-GS~\citep{dong2025cdcgs} & \textbf{25.65} & \underline{0.774} & 0.278 & 5.90 & 20.20 & 0.699 & 0.317 & 5.85 & 20.78 & \underline{0.728} & \underline{0.290} & 6.81 & 21.20 & \underline{0.787} & 0.263 & 3.16 \\
FastGS~\citep{ren2026fastgs} & 23.86 & 0.676 & 0.398 & 0.73 & 19.88 & 0.635 & 0.405 & 0.49 & 20.08 & 0.682 & 0.350 & 0.78 & 20.28 & 0.758 & 0.305 & 0.60 \\
LeGS~\citep{ning2026legs} & 24.36 & 0.707 & 0.355 & 1.51 & 20.08 & 0.656 & 0.376 & 0.92 & 20.27 & 0.696 & 0.328 & 1.79 & 20.50 & 0.767 & 0.289 & 1.48 \\
\midrule
\multicolumn{17}{@{}l}{\emph{Coverage-scaled exploration + self-distillation (ours)}} \\
\textbf{\method{}} & 25.48 & 0.770 & \underline{0.269} & 4.86 & \textbf{21.65} & \textbf{0.750} & \textbf{0.255} & 6.07 & \textbf{22.08} & \textbf{0.796} & \textbf{0.216} & 4.82 & \textbf{22.41} & \textbf{0.857} & \textbf{0.181} & 5.19 \\
\bottomrule
\end{tabular}
\end{table}

\begin{table}[!htbp]
\centering
\caption{\textbf{Per-capture comparison on large handheld captures (Zip-NeRF, $\mathbf{60}$k iterations, without retuning).} Counts $\numg$ are in millions. All three released Taming~3DGS multipliers ($2\times, 5\times, 15\times$) stall behind the inherited gradient growth threshold, whereas \method{} scales automatically.}
\label{tab:supp-transfer-zipnerf_a}\label{tab:supp-transfer-zipnerf_b}
\scriptsize
\setlength{\tabcolsep}{2.1pt}
\begin{tabular}{@{}l rrrr rrrr rrrr rrrr@{}}
\toprule
 & \multicolumn{4}{c}{\textbf{Alameda}} & \multicolumn{4}{c}{\textbf{London}} & \multicolumn{4}{c}{\textbf{NYC}} & \multicolumn{4}{c}{\textbf{Berlin}} \\
\cmidrule(lr){2-5}\cmidrule(lr){6-9}\cmidrule(lr){10-13}\cmidrule(lr){14-17}
Method & PSNR$\uparrow$ & SSIM$\uparrow$ & LPIPS$\downarrow$ & $\numg$ & PSNR$\uparrow$ & SSIM$\uparrow$ & LPIPS$\downarrow$ & $\numg$ & PSNR$\uparrow$ & SSIM$\uparrow$ & LPIPS$\downarrow$ & $\numg$ & PSNR$\uparrow$ & SSIM$\uparrow$ & LPIPS$\downarrow$ & $\numg$ \\
\midrule
\multicolumn{17}{@{}l}{\emph{Fixed-constant controllers}} \\
vanilla 3DGS~\citep{kerbl3Dgaussians} & 22.99 & 0.736 & 0.394 & 0.87 & 25.88 & 0.809 & 0.376 & 0.63 & 27.12 & 0.843 & 0.324 & 1.06 & 27.80 & 0.894 & 0.294 & 0.77 \\
Speedy-Splat~\citep{hanson2025speedysplat} & 21.54 & 0.671 & 0.484 & 0.06 & 25.00 & 0.761 & 0.452 & 0.06 & 25.49 & 0.797 & 0.397 & 0.08 & 26.58 & 0.868 & 0.341 & 0.08 \\
\midrule
\multicolumn{17}{@{}l}{\emph{Preset-count controllers}} \\
3DGS-MCMC~\citep{kheradmand2024mcmc} & 14.50 & 0.525 & 0.640 & 0.87 & 17.36 & 0.607 & 0.565 & 0.63 & 15.04 & 0.607 & 0.599 & 1.06 & 17.57 & 0.726 & 0.485 & 0.77 \\
Taming 3DGS ($2\times$)~\citep{tamings3dgs} & 22.87 & 0.719 & 0.421 & 0.44 & 25.44 & 0.790 & 0.407 & 0.30 & 26.32 & 0.819 & 0.363 & 0.37 & 27.45 & 0.885 & 0.313 & 0.42 \\
Taming 3DGS ($5\times$)~\citep{tamings3dgs} & 22.90 & 0.719 & 0.421 & 0.44 & 25.42 & 0.791 & 0.408 & 0.30 & 26.29 & 0.819 & 0.363 & 0.37 & 27.40 & 0.885 & 0.312 & 0.42 \\
Taming 3DGS ($15\times$)~\citep{tamings3dgs} & 22.83 & 0.719 & 0.422 & 0.43 & 25.42 & 0.792 & 0.407 & 0.30 & 26.23 & 0.818 & 0.364 & 0.37 & 27.42 & 0.884 & 0.312 & 0.42 \\
Mini-Splatting~\citep{fang2024minisplatting} & 20.11 & 0.663 & 0.488 & 0.16 & 25.69 & 0.798 & 0.389 & 0.30 & 25.46 & 0.804 & 0.375 & 0.22 & 26.66 & 0.876 & 0.322 & 0.26 \\
LiteGS (1M)~\citep{litegs} & \underline{23.00} & \underline{0.743} & \underline{0.381} & 0.98 & 26.11 & \underline{0.816} & \underline{0.360} & 0.96 & \underline{27.73} & \underline{0.855} & \underline{0.305} & 0.98 & \underline{27.95} & \underline{0.903} & \underline{0.275} & 0.98 \\
\midrule
\multicolumn{17}{@{}l}{\emph{Automatic-budget controllers}} \\
DashGaussian~\citep{chen2025dashgaussian} & 22.89 & 0.731 & 0.397 & 0.99 & 26.14 & 0.815 & 0.366 & 0.85 & 26.81 & 0.832 & 0.333 & 1.03 & 27.59 & 0.892 & 0.296 & 0.98 \\
Compact 3DGS~\citep{lee2024compact} & 21.52 & 0.691 & 0.455 & 0.22 & 24.95 & 0.777 & 0.422 & 0.17 & 25.37 & 0.800 & 0.383 & 0.19 & 26.27 & 0.867 & 0.335 & 0.20 \\
LP-3DGS~\citep{zhang2024lp3dgs} & 22.15 & 0.708 & 0.434 & 0.23 & 25.57 & 0.797 & 0.396 & 0.22 & 26.30 & 0.820 & 0.358 & 0.21 & 27.10 & 0.883 & 0.314 & 0.23 \\
ControlGS~\citep{controlgs2025} & 22.28 & 0.704 & 0.432 & 0.35 & 25.09 & 0.782 & 0.412 & 0.29 & 25.83 & 0.806 & 0.372 & 0.28 & 26.69 & 0.870 & 0.329 & 0.28 \\
CDC-GS~\citep{dong2025cdcgs} & 22.83 & 0.731 & 0.399 & 0.91 & 26.15 & 0.814 & 0.369 & 0.80 & 27.10 & 0.839 & 0.328 & 0.98 & 27.93 & 0.898 & 0.285 & 0.94 \\
FastGS~\citep{ren2026fastgs} & 22.00 & 0.700 & 0.442 & 0.17 & 25.18 & 0.781 & 0.424 & 0.12 & 25.88 & 0.805 & 0.384 & 0.12 & 26.64 & 0.873 & 0.331 & 0.14 \\
LeGS~\citep{ning2026legs} & 21.98 & 0.711 & 0.420 & 0.50 & \underline{26.18} & \underline{0.816} & 0.365 & 0.58 & 26.76 & 0.827 & 0.343 & 0.37 & 27.16 & 0.886 & 0.306 & 0.37 \\
\midrule
\multicolumn{17}{@{}l}{\emph{Coverage-scaled exploration + self-distillation (ours)}} \\
\textbf{\method{}} & \textbf{23.09} & \textbf{0.771} & \textbf{0.325} & 3.80 & \textbf{26.81} & \textbf{0.843} & \textbf{0.300} & 2.81 & \textbf{28.27} & \textbf{0.872} & \textbf{0.270} & 2.11 & \textbf{28.16} & \textbf{0.911} & \textbf{0.250} & 1.59 \\
\bottomrule
\end{tabular}
\end{table}

\begin{table}[!htbp]
\centering
\caption{\textbf{Three-seed stability of \method{} on large captures.} Mean $\pm$ std across seeds 4114, 4115, and 4116 under the unchanged Std-13 calibration.}
\label{tab:supp-support-seeds-large}
\footnotesize
\setlength{\tabcolsep}{5pt}
\begin{tabular}{@{}l rrrr@{}}
\toprule
Capture & $\numg$ (M) & PSNR (dB)$\uparrow$ & SSIM$\uparrow$ & LPIPS$\downarrow$ \\
\midrule
Rubble & 4.86 $\pm$ 0.01 & 25.48 $\pm$ 0.10 & 0.770 $\pm$ 0.001 & 0.269 $\pm$ 0.001 \\
Building & 6.07 $\pm$ 0.01 & 21.65 $\pm$ 0.08 & 0.750 $\pm$ 0.002 & 0.255 $\pm$ 0.001 \\
Residence & 4.82 $\pm$ 0.04 & 22.08 $\pm$ 0.12 & 0.796 $\pm$ 0.002 & 0.216 $\pm$ 0.001 \\
Sci-Art & 5.19 $\pm$ 0.06 & 22.41 $\pm$ 0.22 & 0.857 $\pm$ 0.001 & 0.181 $\pm$ 0.001 \\
Alameda & 3.80 $\pm$ 0.03 & 23.09 $\pm$ 0.04 & 0.771 $\pm$ 0.000 & 0.325 $\pm$ 0.000 \\
Berlin & 1.59 $\pm$ 0.00 & 28.16 $\pm$ 0.07 & 0.911 $\pm$ 0.000 & 0.250 $\pm$ 0.000 \\
London & 2.81 $\pm$ 0.03 & 26.81 $\pm$ 0.03 & 0.843 $\pm$ 0.000 & 0.300 $\pm$ 0.001 \\
NYC & 2.11 $\pm$ 0.02 & 28.27 $\pm$ 0.01 & 0.872 $\pm$ 0.000 & 0.270 $\pm$ 0.000 \\
\midrule
\textbf{Large-8 mean} & \textbf{3.91} $\pm$ 0.01 & \textbf{24.74} $\pm$ 0.03 & \textbf{0.821} $\pm$ 0.000 & \textbf{0.258} $\pm$ 0.000 \\
\bottomrule
\end{tabular}
\end{table}

\clearpage
\section{Controller Ablations}
\label{sec:supp-keepratio}\label{sec:exp-ablations}

\subsection{One-shot versus sequential self-distillation}
\begin{figure}[!htbp]
    \centering
    \includegraphics[width=\linewidth]{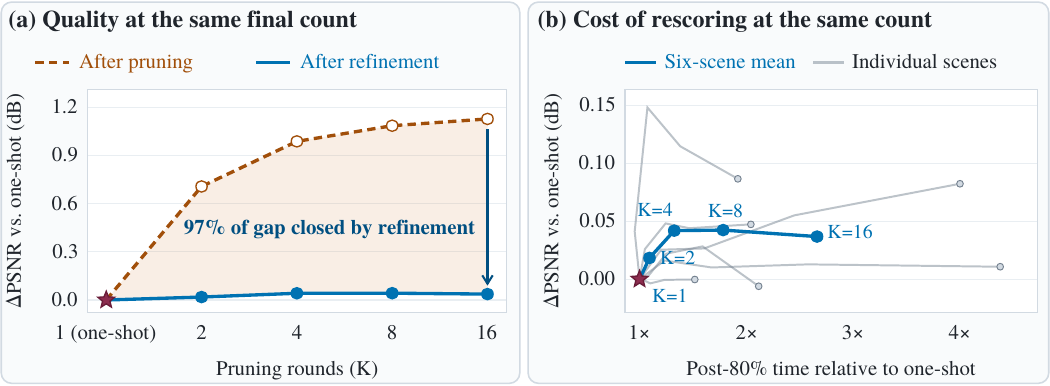}
    \caption{\textbf{One-shot ($K=1$) versus $K$-round sequential self-distillation across six scenes (seed 4114).} \textbf{(a)}~Mean test $\Delta\mathrm{PSNR}$ relative to one-shot self-distillation at the same final Gaussian count, measured immediately after pruning ($80\%$) and after the refinement phase ($100\%$). \textbf{(b)}~Refined $\Delta\mathrm{PSNR}$ versus relative post-$80\%$ time for the same runs; faint curves show individual scenes and the labeled curve shows the six-scene mean, with time normalized to $K=1$ within each scene.}
    \label{fig:supp-sequential-distill}
\end{figure}

\method{} approximates teacher rendering error with the additive leave-one-out surrogate in Eq.~(4) of the main paper, scoring all Gaussians once on the $80\%$ teacher model and pruning those with $\util \le \lamstar$ in one step.
If two Gaussians cover the same surface and removing either alone barely changes the rendered image, both can receive low scores and be dropped together.
Sequential self-distillation instead removes a smaller batch, recomputes $\util$ on the remaining Gaussians, and keeps primitives whose contribution increases once neighbors are removed.
Starting from the same $80\%$ teacher checkpoints on six scenes spanning $12\%$--$87\%$ retention, \cref{fig:supp-sequential-distill} and \cref{tab:supp-sequential-distill} compare one-shot self-distillation ($K = 1$) against $K \in \{2, 4, 8, 16\}$ rounds of pruning and rescoring, either matching \method{}'s final Gaussian count $\numg$ or reapplying the fixed threshold $\lamstar$.
Immediately after pruning at $80\%$, $16$ matched-count rounds raise mean test PSNR by $1.13$\,dB ($+3.22$\,dB on \emph{Playroom}).
Over the remaining $20\%$ refinement phase, however, the retained Gaussians adapt to cover those gaps.
By the end of training, the mean PSNR gain drops to $+0.04$\,dB with a mean LPIPS change of $+0.0003$, while mean time after the $80\%$ checkpoint (\cref{fig:supp-sequential-distill}b) increases from $42$\,s to $139$\,s ($84$\,s to $368$\,s on \emph{London}).
Reapplying $\lamstar$ at each round instead gains $0.09$\,dB at $K = 16$ while keeping $380$K more Gaussians on average ($+34\%$; \cref{tab:supp-sequential-distill}), a quality--size trade-off already reachable in one-shot self-distillation by lowering $\lambda$ (\cref{sec:supp-capacity-objective}).

\begin{table}[!htbp]
\centering
\caption{\textbf{One-shot ($K=1$) versus $K$-round sequential self-distillation from identical $80\%$ teacher checkpoints (seed 4114).} Upper rows report six-scene means across $K \in \{1, 2, 4, 8, 16\}$ when matching the one-shot final count ($\numg(K) = \numg(1)$) or reapplying the fixed threshold $\lamstar$ (dropping $1/(K - r + 1)$ of remaining primitives with $\util \le \lamstar$ in round $r$). Lower rows show the per-scene change from $K=1$ to $K=16$ under matched counts, ordered by one-shot retention $\numg / \npre$, with fixed-$\lamstar$ count increases $\Delta\numg$ at $K=16$ given in parentheses in the last column. Utility scoring (CUDA pass, $>0.99999$ top-$\numg$ Jaccard match to float64 reference) and the $20\%$ refinement phase are timed on one NVIDIA GH200; test metrics use TorchMetrics SSIM and VGG LPIPS.}
\label{tab:supp-sequential-distill}
\scriptsize
\setlength{\tabcolsep}{2.7pt}
\begin{tabular}{@{}l r rr r r rrr rrr@{}}
\toprule
 & & \multicolumn{2}{c}{\textbf{Subset vs.\ $K=1$}} & \textbf{Teacher} & \textbf{After pruning} & \multicolumn{3}{c}{\textbf{$100\%$ refined test quality}} & \multicolumn{3}{c}{\textbf{Post-$80\%$ time (s)}} \\
\cmidrule(lr){3-4}\cmidrule(lr){7-9}\cmidrule(lr){10-12}
Schedule / Capture & $K$ & $\numg$(M) & Jaccard & RGB err.$\downarrow$ & $\Delta$PSNR & PSNR$\uparrow$ & SSIM$\uparrow$ & LPIPS$\downarrow$ & Score & Refine & Total \\
\midrule
\textbf{One-shot (\method{})} & 1 & 1.12 & 1.000 & 192.3 & $0.00$ & 29.82 & 0.900 & 0.220 & 8.8 & 33.2 & 42.0 \\
Matched $\numg$ & 2 & 1.12 & 0.924 & 119.4 & $+0.71$ & 29.84 & 0.900 & 0.219 & 15.1 & 33.1 & 48.2 \\
Matched $\numg$ & 4 & 1.12 & 0.843 & 92.1 & $+0.99$ & 29.87 & 0.901 & 0.219 & 28.2 & 33.2 & 61.4 \\
Matched $\numg$ & 8 & 1.12 & 0.801 & 83.8 & $+1.09$ & 29.87 & 0.901 & 0.220 & 54.8 & 33.1 & 87.8 \\
Matched $\numg$ & 16 & 1.12 & 0.780 & 80.3 & $+1.13$ & 29.86 & 0.901 & 0.220 & 105.8 & 33.2 & 139.0 \\
\midrule
Fixed $\lamstar$ & 2 & 1.19 & 0.916 & 95.1 & $+1.01$ & 29.88 & 0.901 & 0.217 & 15.1 & 34.0 & 49.1 \\
Fixed $\lamstar$ & 4 & 1.36 & 0.802 & 48.6 & $+1.57$ & 29.90 & 0.902 & 0.214 & 27.9 & 36.0 & 63.9 \\
Fixed $\lamstar$ & 8 & 1.46 & 0.754 & 36.5 & $+1.73$ & 29.93 & 0.903 & 0.213 & 54.5 & 37.3 & 91.8 \\
Fixed $\lamstar$ & 16 & 1.50 & 0.737 & 32.9 & $+1.78$ & 29.92 & 0.903 & 0.212 & 106.5 & 38.3 & 144.8 \\
\midrule
\multicolumn{12}{@{}l}{\emph{Per-scene shift from $K=1$ to $K=16$ under matched $\numg$}} \\
\midrule
Playroom (12.3\%) & $1\!\to\!16$ & 0.32 & 0.502 & 382.2$\to$111.5 & $+3.22$ & 30.78 ($+0.09$) & 0.913 & 0.244 & 3.5$\to$21.1 & 15.1 & 18.8$\to$36.2 ($+366$K) \\
Berlin (19.3\%) & $1\!\to\!16$ & 1.59 & 0.750 & 322.6$\to$123.1 & $+0.95$ & 28.19 ($+0.08$) & 0.912 & 0.247 & 17.6$\to$224.0 & 49.7 & 68.3$\to$273.7 ($+669$K) \\
Room (19.8\%) & $1\!\to\!16$ & 0.39 & 0.793 & 47.7$\to$22.8 & $+1.25$ & 32.56 ($-0.01$) & 0.925 & 0.212 & 2.3$\to$30.7 & 23.4 & 25.5$\to$54.1 ($+139$K) \\
London (25.3\%) & $1\!\to\!16$ & 2.84 & 0.820 & 337.8$\to$186.8 & $+0.41$ & 26.84 ($+0.01$) & 0.845 & 0.297 & 25.3$\to$309.1 & 59.0 & 83.9$\to$368.1 ($+907$K) \\
Bonsai (30.0\%) & $1\!\to\!16$ & 0.64 & 0.818 & 62.8$\to$36.7 & $+0.93$ & 33.11 ($+0.05$) & 0.949 & 0.187 & 3.2$\to$38.7 & 31.9 & 34.5$\to$70.6 ($+196$K) \\
Garden (86.9\%) & $1\!\to\!16$ & 0.92 & 0.997 & 0.78$\to$0.76 & $0.00$ & 27.70 ($0.00$) & 0.860 & 0.131 & 0.8$\to$11.4 & 20.1 & 20.7$\to$31.5 ($+2.7$K) \\
\bottomrule
\end{tabular}
\end{table}

\FloatBarrier

\subsection{Alternative view-weighting and scoring rules}
\Cref{tab:supp-weighting-transfer} and \cref{tab:supp-teacher-vs-gt} evaluate alternative view weights and scoring rules starting from the same $80\%$ teacher checkpoints, calibrating each rule once on Std-13 to match the reference mean budget ($\bref = 0.93$M) before transferring unchanged to Large-8.
\Cref{tab:supp-weighting-transfer} replaces our coverage weights $\covw / \sqrt{W_i H_i}$ with pixel-averaged MSE, $(\sum_i W_i H_i)^{-1}\sum_i u_{g,i}$, or a raw pixel sum, $\sum_i u_{g,i}$.
Pixel-averaged MSE dilutes utility scores on larger multi-view captures (\textbf{P3}) and cuts Large-8 retention to $24.0\%$ ($-0.23$\,dB PSNR, $+0.031$ LPIPS), while the raw pixel sum ignores view overlap (\textbf{P1}, \textbf{P2}) and increases Large-8 final counts by $22.1\%$ ($4.77$M vs.\ $3.90$M).
\Cref{tab:supp-teacher-vs-gt} instead scores primitives by opacity sensitivity of the ground-truth residual ($I_i - \hat{I}_i$) rather than teacher renderings ($\hat{I}_i$).
Ground-truth residuals can be small on well-fit surfaces and sensitive to exposure differences. This baseline lowers Std-13 mean PSNR by $0.16$\,dB and increases Large-8 final counts by $35.1\%$ ($5.27$M vs.\ $3.90$M).

\begin{table}[!htbp]
\centering
\caption{\textbf{View-weighting transfer from Std-13 to Large-8 on identical $80\%$ teacher checkpoints (seed 4114).} Each scoring rule is calibrated once on Std-13 to match the reference mean budget ($\bref = 0.93$M; $12{,}026{,}752$ total Gaussians) and transferred unchanged to Large-8 (TorchMetrics SSIM and VGG LPIPS).}
\label{tab:supp-weighting-transfer}
\scriptsize
\setlength{\tabcolsep}{2.9pt}
\begin{tabular}{@{}l rrrr rrrr rrrr@{}}
\toprule
 & \multicolumn{4}{c}{\textbf{\method{} ($\covw / \sqrt{W_i H_i}$)}} & \multicolumn{4}{c}{\textbf{Pixel-averaged MSE}} & \multicolumn{4}{c}{\textbf{Raw pixel sum}} \\
\cmidrule(lr){2-5}\cmidrule(lr){6-9}\cmidrule(lr){10-13}
Capture & $\numg$(M) & Kept & PSNR$\uparrow$ & LPIPS$\downarrow$ & $\numg$(M) & Kept & PSNR$\uparrow$ & LPIPS$\downarrow$ & $\numg$(M) & Kept & PSNR$\uparrow$ & LPIPS$\downarrow$ \\
\midrule
Rubble & 4.86 & 77.9\% & 25.37 & 0.270 & 3.70 & 59.3\% & 25.29 & 0.280 & 5.41 & 86.6\% & 25.38 & 0.267 \\
Building & 6.08 & 71.5\% & 21.56 & 0.256 & 3.47 & 40.8\% & 21.40 & 0.276 & 6.88 & 80.9\% & 21.64 & 0.253 \\
Residence & 4.77 & 63.9\% & 22.21 & 0.216 & 2.81 & 37.6\% & 22.12 & 0.229 & 5.96 & 79.7\% & 22.22 & 0.213 \\
Sci-Art & 5.18 & 46.2\% & 22.16 & 0.180 & 1.27 & 11.3\% & 21.93 & 0.250 & 7.25 & 64.6\% & 22.29 & 0.174 \\
Alameda & 3.82 & 40.3\% & 23.13 & 0.325 & 1.11 & 11.7\% & 22.90 & 0.364 & 4.55 & 48.0\% & 23.14 & 0.322 \\
Berlin & 1.59 & 19.3\% & 28.11 & 0.250 & 0.82 & 10.0\% & 27.86 & 0.270 & 2.45 & 29.8\% & 28.24 & 0.239 \\
London & 2.84 & 25.3\% & 26.83 & 0.300 & 1.12 & 10.0\% & 26.51 & 0.341 & 3.36 & 30.0\% & 26.87 & 0.295 \\
NYC & 2.09 & 29.9\% & 28.26 & 0.270 & 0.80 & 11.4\% & 27.73 & 0.301 & 2.26 & 32.4\% & 28.28 & 0.268 \\
\midrule
\textbf{Large-8 mean} & \textbf{3.90} & \textbf{46.8\%} & \textbf{24.70} & \textbf{0.258} & 1.89 & 24.0\% & 24.47 & 0.289 & 4.77 & 56.5\% & 24.76 & 0.254 \\
\bottomrule
\end{tabular}
\end{table}

\begin{table}[!htbp]
\centering
\caption{\textbf{Teacher self-distillation ($\hat{I}_i$) versus ground-truth $L_2$ residual opacity-sensitivity pruning ($I_i - \hat{I}_i$) across all 21 scenes (seed 4114).} Both share identical $80\%$ teacher checkpoints, view weights $\covw/\sqrt{W_i H_i}$, and Std-13 calibration ($\bref = 0.93$M), transferred unchanged to Large-8 (SSIM uses LiteGS's native Gaussian-window estimator). Jaccard measures top-$k$ subset overlap at \method{}'s retained count $\numg$.}
\label{tab:supp-teacher-vs-gt}
\footnotesize
\setlength{\tabcolsep}{3.8pt}
\begin{tabular}{@{}l rrrr rrrr r@{}}
\toprule
 & \multicolumn{4}{c}{\textbf{Teacher self-distillation ($\hat{I}_i$, ours)}} & \multicolumn{4}{c}{\textbf{GT $L_2$ residual ($I_i - \hat{I}_i$)}} & \\
\cmidrule(lr){2-5}\cmidrule(lr){6-9}
Capture & $\numg$(M) & PSNR$\uparrow$ & SSIM$\uparrow$ & LPIPS$\downarrow$ & $\numg$(M) & PSNR$\uparrow$ & SSIM$\uparrow$ & LPIPS$\downarrow$ & Jaccard \\
\midrule
Bicycle & 1.42 & 25.76 & \textbf{0.791} & \textbf{0.198} & 1.52 & \textbf{25.79} & 0.790 & 0.201 & 0.868 \\
Bonsai & 0.64 & \textbf{33.07} & \textbf{0.950} & \textbf{0.187} & 0.55 & 32.61 & 0.937 & 0.222 & 0.541 \\
Counter & 0.65 & \textbf{29.68} & \textbf{0.920} & \textbf{0.179} & 0.61 & 29.51 & 0.909 & 0.204 & 0.588 \\
Flowers & 1.45 & \textbf{21.94} & \textbf{0.644} & 0.285 & 1.55 & \textbf{21.94} & \textbf{0.644} & \textbf{0.284} & 0.937 \\
Garden & 0.92 & \textbf{27.70} & \textbf{0.862} & \textbf{0.131} & 0.89 & 27.68 & 0.860 & 0.136 & 0.931 \\
Kitchen & 0.81 & \textbf{32.56} & \textbf{0.937} & \textbf{0.117} & 0.54 & 32.27 & 0.928 & 0.136 & 0.772 \\
Room & 0.39 & \textbf{32.55} & \textbf{0.926} & \textbf{0.208} & 0.32 & 32.08 & 0.905 & 0.264 & 0.568 \\
Stump & 1.22 & \textbf{27.53} & \textbf{0.814} & \textbf{0.184} & 1.20 & 27.49 & 0.811 & 0.189 & 0.873 \\
Treehill & 1.52 & \textbf{23.26} & \textbf{0.664} & \textbf{0.290} & 1.63 & 23.24 & 0.663 & 0.292 & 0.874 \\
Truck & 0.69 & \textbf{26.62} & \textbf{0.897} & \textbf{0.124} & 0.85 & 26.59 & 0.895 & 0.132 & 0.727 \\
Train & 1.07 & \textbf{23.05} & 0.849 & 0.161 & 1.28 & 23.04 & \textbf{0.850} & \textbf{0.159} & 0.858 \\
Dr Johnson & 0.91 & \textbf{29.93} & \textbf{0.913} & \textbf{0.230} & 0.78 & 29.77 & 0.908 & 0.249 & 0.578 \\
Playroom & 0.32 & \textbf{30.71} & \textbf{0.913} & \textbf{0.241} & 0.29 & 30.30 & 0.906 & 0.268 & 0.453 \\
\midrule
Rubble & 4.86 & \textbf{25.37} & 0.772 & 0.270 & 5.25 & \textbf{25.37} & \textbf{0.773} & \textbf{0.269} & 0.901 \\
Building & 6.08 & 21.56 & 0.753 & 0.256 & 6.84 & \textbf{21.57} & \textbf{0.754} & \textbf{0.255} & 0.865 \\
Residence & 4.77 & 22.21 & 0.799 & 0.216 & 6.07 & \textbf{22.23} & \textbf{0.801} & \textbf{0.213} & 0.865 \\
Sci-Art & 5.18 & 22.16 & 0.858 & 0.180 & 8.51 & \textbf{22.21} & \textbf{0.862} & \textbf{0.173} & 0.811 \\
Alameda & 3.82 & 23.13 & 0.774 & 0.325 & 6.09 & \textbf{23.15} & \textbf{0.776} & \textbf{0.320} & 0.711 \\
Berlin & 1.59 & \textbf{28.11} & \textbf{0.911} & \textbf{0.250} & 2.33 & 28.08 & 0.910 & 0.255 & 0.607 \\
London & 2.84 & \textbf{26.83} & \textbf{0.844} & \textbf{0.300} & 4.59 & 26.78 & 0.840 & 0.309 & 0.524 \\
NYC & 2.09 & \textbf{28.26} & \textbf{0.874} & \textbf{0.270} & 2.52 & 28.17 & 0.869 & 0.282 & 0.625 \\
\midrule
\textbf{Std-13 mean} & 0.93 & \textbf{28.03} & \textbf{0.852} & \textbf{0.195} & 0.93 & 27.87 & 0.847 & 0.210 & 0.736 \\
\textbf{Large-8 mean} & 3.90 & \textbf{24.70} & \textbf{0.823} & \textbf{0.258} & 5.27 & 24.69 & \textbf{0.823} & 0.259 & 0.739 \\
\textbf{All-21 mean} & 2.06 & \textbf{26.76} & \textbf{0.841} & \textbf{0.219} & 2.58 & 26.66 & 0.838 & 0.229 & 0.737 \\
\bottomrule
\end{tabular}
\end{table}

\FloatBarrier

\subsection{Exploration, self-distillation, and trajectory ablations}
\label{sec:supp-content-adaptation}
\Cref{tab:supp-component-means-standard} gives the per-scene results for the main paper's factorial ablation figure.
\Cref{tab:supp-litegs-selected-count} compares \method{} with released LiteGS given \method{}'s final Gaussian count as the target for each scene; LiteGS finishes $0.5\%$--$7.4\%$ below those targets.
\Cref{tab:supp-garden-frozen21} compares our resolution-normalized score $U_{\mathrm{sqrt}} = \util$ with the unnormalized pixel sum $U_{\mathrm{raw}} = \sum_i \covw u_{g,i}$ across four \emph{Garden} resolutions ($0.65$K--$5.2$K) without retuning.

\begin{table}[!htbp]
\centering
\caption{\textbf{Per-scene $2 \times 2$ factorial ablation of coverage-scaled exploration and self-distillation (seed 4114).} Paired variants share $80\%$ teacher checkpoints and refinement schedules. Counts $\numg$ are in millions.}
\label{tab:supp-component-means-standard}\label{tab:supp-component-means-large}
\scriptsize
\setlength{\tabcolsep}{2.8pt}
\begin{tabular}{@{}l rrr rrr rrr rrr@{}}
\toprule
 & \multicolumn{3}{c}{\textbf{Both (\method{})}} & \multicolumn{3}{c}{\textbf{Exploration only}} & \multicolumn{3}{c}{\textbf{Self-distill.\ only (1M cap)}} & \multicolumn{3}{c}{\textbf{Neither (1M cap)}} \\
\cmidrule(lr){2-4}\cmidrule(lr){5-7}\cmidrule(lr){8-10}\cmidrule(lr){11-13}
Capture & $\numg$(M) & PSNR$\uparrow$ & LPIPS$\downarrow$ & $\numg$(M) & PSNR$\uparrow$ & LPIPS$\downarrow$ & $\numg$(M) & PSNR$\uparrow$ & LPIPS$\downarrow$ & $\numg$(M) & PSNR$\uparrow$ & LPIPS$\downarrow$ \\
\midrule
Bicycle & 1.42 & \underline{25.76} & \underline{0.198} & 1.78 & \textbf{25.79} & \textbf{0.193} & 0.86 & 25.64 & 0.236 & 0.98 & 25.65 & 0.233 \\
Bonsai & 0.64 & \underline{33.07} & 0.187 & 2.15 & \textbf{33.36} & \textbf{0.176} & 0.64 & 32.94 & 0.189 & 0.99 & 33.00 & \underline{0.186} \\
Counter & 0.65 & \underline{29.68} & \underline{0.179} & 2.18 & \textbf{29.76} & \textbf{0.167} & 0.53 & 29.58 & 0.184 & 0.97 & 29.63 & \underline{0.179} \\
Flowers & 1.45 & 21.94 & \underline{0.285} & 1.68 & 21.93 & \textbf{0.282} & 0.91 & \textbf{21.99} & 0.315 & 0.99 & \underline{21.98} & 0.313 \\
Garden & 0.92 & \underline{27.70} & \underline{0.131} & 1.06 & \textbf{27.71} & \textbf{0.129} & 0.86 & 27.62 & 0.136 & 0.97 & 27.64 & 0.134 \\
Kitchen & 0.81 & \underline{32.56} & \underline{0.117} & 1.38 & \textbf{32.65} & \textbf{0.113} & 0.72 & 32.34 & 0.119 & 0.99 & 32.43 & \underline{0.117} \\
Room & 0.39 & 32.55 & 0.208 & 1.99 & \textbf{32.69} & \textbf{0.184} & 0.43 & 32.56 & 0.201 & 0.99 & \underline{32.67} & \underline{0.192} \\
Stump & 1.22 & \textbf{27.53} & \underline{0.184} & 1.61 & \underline{27.51} & \textbf{0.180} & 0.84 & 27.45 & 0.203 & 0.99 & 27.44 & 0.200 \\
Treehill & 1.52 & \textbf{23.26} & \underline{0.290} & 1.82 & 23.23 & \textbf{0.286} & 0.88 & \underline{23.24} & 0.323 & 0.98 & 23.23 & 0.320 \\
Truck & 0.69 & \underline{26.62} & \underline{0.124} & 1.33 & \textbf{26.63} & \textbf{0.119} & 0.60 & 26.56 & 0.134 & 0.97 & 26.58 & 0.131 \\
Train & 1.07 & \underline{23.05} & \underline{0.161} & 1.50 & \textbf{23.09} & \textbf{0.156} & 0.81 & 22.87 & 0.171 & 0.99 & 22.88 & 0.168 \\
Dr Johnson & 0.91 & 29.93 & 0.230 & 3.71 & 29.95 & \textbf{0.216} & 0.66 & \underline{30.06} & 0.231 & 0.96 & \textbf{30.12} & \underline{0.227} \\
Playroom & 0.32 & 30.71 & 0.241 & 2.58 & 30.83 & \textbf{0.214} & 0.38 & \underline{30.98} & 0.238 & 0.99 & \textbf{31.04} & \underline{0.229} \\
\midrule
Rubble & 4.86 & \underline{25.37} & \underline{0.270} & 6.24 & \textbf{25.40} & \textbf{0.265} & 1.07 & 24.27 & 0.373 & 1.10 & 24.27 & 0.373 \\
Building & 6.08 & \textbf{21.56} & \underline{0.256} & 8.50 & \textbf{21.56} & \textbf{0.252} & 1.10 & 20.68 & 0.338 & 1.13 & \underline{20.69} & 0.337 \\
Residence & 4.77 & \underline{22.21} & \underline{0.216} & 7.47 & \textbf{22.22} & \textbf{0.212} & 1.17 & 21.47 & 0.273 & 1.21 & 21.45 & 0.273 \\
Sci-Art & 5.18 & \underline{22.16} & \underline{0.180} & 11.22 & \textbf{22.24} & \textbf{0.171} & 1.10 & 21.68 & 0.231 & 1.14 & 21.70 & 0.231 \\
Alameda & 3.82 & 23.13 & \underline{0.325} & 9.48 & \textbf{23.18} & \textbf{0.313} & 0.92 & \underline{23.15} & 0.364 & 0.98 & 23.14 & 0.363 \\
Berlin & 1.59 & \underline{28.11} & \underline{0.250} & 8.24 & \textbf{28.29} & \textbf{0.230} & 0.80 & 28.01 & 0.268 & 0.98 & 28.03 & 0.267 \\
London & 2.84 & \underline{26.83} & \underline{0.300} & 11.22 & \textbf{26.93} & \textbf{0.281} & 0.86 & 26.57 & 0.342 & 0.97 & 26.58 & 0.341 \\
NYC & 2.09 & \underline{28.26} & \underline{0.270} & 6.98 & \textbf{28.38} & \textbf{0.255} & 0.86 & 28.04 & 0.295 & 0.97 & 28.07 & 0.293 \\
\midrule
\textbf{Std-13 mean} & 0.93 & \underline{28.03} & \underline{0.195} & 1.91 & \textbf{28.09} & \textbf{0.186} & 0.70 & 27.99 & 0.206 & 0.98 & 28.02 & 0.202 \\
\textbf{Large-8 mean} & 3.90 & \underline{24.70} & \underline{0.258} & 8.67 & \textbf{24.78} & \textbf{0.247} & 0.99 & 24.23 & 0.311 & 1.06 & 24.24 & 0.310 \\
\textbf{All-21 mean} & 2.06 & \underline{26.76} & \underline{0.219} & 4.48 & \textbf{26.83} & \textbf{0.209} & 0.81 & 26.56 & 0.246 & 1.01 & 26.58 & 0.243 \\
\bottomrule
\end{tabular}
\end{table}

\begin{table}[!htbp]
\centering
\caption{\textbf{Trajectory ablation: \method{} versus released LiteGS given \method{}'s per-scene final count (seed 4114).} Positive $\Delta\mathrm{PSNR}$ and negative $\Delta\mathrm{LPIPS}$ favor \method{}. Even with oracle per-scene targets, \method{} improves SSIM and LPIPS on all 21 scenes.}
\label{tab:supp-litegs-selected-count}
\footnotesize
\setlength{\tabcolsep}{3.5pt}
\begin{tabular}{@{}l rr rrr rr rrr@{}}
\toprule
 & \multicolumn{2}{c}{\textbf{$\numg$ (M)}} & \multicolumn{3}{c}{\textbf{PSNR (dB)$\uparrow$}} & \multicolumn{2}{c}{\textbf{SSIM$\uparrow$}} & \multicolumn{3}{c}{\textbf{LPIPS$\downarrow$}} \\
\cmidrule(lr){2-3}\cmidrule(lr){4-6}\cmidrule(lr){7-8}\cmidrule(lr){9-11}
Capture & Ours & LiteGS & Ours & LiteGS & $\Delta$ & Ours & LiteGS & Ours & LiteGS & $\Delta$ \\
\midrule
Bicycle & 1.42 & 1.41 & \textbf{25.76} & 25.50 & +0.26 & \textbf{0.788} & 0.774 & \textbf{0.198} & 0.212 & -0.014 \\
Bonsai & 0.64 & 0.64 & \textbf{33.07} & 32.81 & +0.26 & \textbf{0.949} & 0.947 & \textbf{0.187} & 0.193 & -0.006 \\
Counter & 0.65 & 0.64 & 29.68 & \textbf{29.69} & -0.01 & \textbf{0.919} & 0.917 & \textbf{0.179} & 0.185 & -0.007 \\
Flowers & 1.45 & 1.44 & \textbf{21.94} & 21.77 & +0.18 & \textbf{0.639} & 0.630 & \textbf{0.285} & 0.295 & -0.010 \\
Garden & 0.92 & 0.92 & \textbf{27.70} & 27.46 & +0.24 & \textbf{0.860} & 0.853 & \textbf{0.131} & 0.139 & -0.008 \\
Kitchen & 0.81 & 0.80 & \textbf{32.56} & 32.32 & +0.24 & \textbf{0.936} & 0.933 & \textbf{0.117} & 0.120 & -0.003 \\
Room & 0.39 & 0.39 & \textbf{32.55} & 32.35 & +0.20 & \textbf{0.926} & 0.925 & \textbf{0.208} & 0.215 & -0.006 \\
Stump & 1.22 & 1.22 & \textbf{27.53} & 27.22 & +0.31 & \textbf{0.811} & 0.795 & \textbf{0.184} & 0.204 & -0.020 \\
Treehill & 1.52 & 1.52 & \textbf{23.26} & 23.14 & +0.11 & \textbf{0.660} & 0.653 & \textbf{0.290} & 0.300 & -0.010 \\
Truck & 0.69 & 0.68 & \textbf{26.62} & 26.40 & +0.22 & \textbf{0.895} & 0.892 & \textbf{0.124} & 0.125 & -0.001 \\
Train & 1.07 & 1.06 & \textbf{23.05} & 22.96 & +0.09 & \textbf{0.846} & 0.841 & \textbf{0.161} & 0.172 & -0.011 \\
Dr Johnson & 0.91 & 0.90 & \textbf{29.93} & 29.87 & +0.05 & \textbf{0.912} & 0.910 & \textbf{0.230} & 0.239 & -0.009 \\
Playroom & 0.32 & 0.31 & 30.71 & \textbf{30.97} & -0.26 & \textbf{0.912} & 0.911 & \textbf{0.241} & 0.250 & -0.008 \\
\midrule
Rubble & 4.86 & 4.69 & \textbf{25.37} & 24.84 & +0.53 & \textbf{0.769} & 0.735 & \textbf{0.270} & 0.319 & -0.049 \\
Building & 6.08 & 5.83 & \textbf{21.56} & 21.03 & +0.54 & \textbf{0.750} & 0.711 & \textbf{0.256} & 0.304 & -0.048 \\
Residence & 4.77 & 4.42 & \textbf{22.21} & 21.65 & +0.56 & \textbf{0.797} & 0.768 & \textbf{0.216} & 0.256 & -0.040 \\
Sci-Art & 5.18 & 4.80 & 22.16 & \textbf{22.97} & -0.81 & \textbf{0.856} & 0.843 & \textbf{0.180} & 0.207 & -0.026 \\
Alameda & 3.82 & 3.70 & \textbf{23.13} & 23.01 & +0.12 & \textbf{0.771} & 0.745 & \textbf{0.325} & 0.373 & -0.047 \\
Berlin & 1.59 & 1.55 & \textbf{28.11} & 27.84 & +0.27 & \textbf{0.910} & 0.901 & \textbf{0.250} & 0.276 & -0.027 \\
London & 2.84 & 2.69 & \textbf{26.83} & 25.86 & +0.97 & \textbf{0.843} & 0.809 & \textbf{0.300} & 0.368 & -0.068 \\
NYC & 2.09 & 2.05 & \textbf{28.26} & 27.60 & +0.66 & \textbf{0.872} & 0.855 & \textbf{0.270} & 0.301 & -0.030 \\
\midrule
\textbf{Std-13 mean} & 0.93 & 0.92 & \textbf{28.03} & 27.88 & +0.15 & \textbf{0.850} & 0.845 & \textbf{0.195} & 0.204 & -0.009 \\
\textbf{Large-8 mean} & 3.90 & 3.71 & \textbf{24.70} & 24.35 & +0.35 & \textbf{0.821} & 0.796 & \textbf{0.258} & 0.300 & -0.042 \\
\textbf{All-21 mean} & 2.06 & 1.98 & \textbf{26.76} & 26.54 & +0.23 & \textbf{0.839} & 0.826 & \textbf{0.219} & 0.241 & -0.021 \\
\bottomrule
\end{tabular}
\end{table}

\begin{table}[!htbp]
\centering
\caption{\textbf{Multi-resolution scaling on \emph{Garden} ($0.65$K--$5.2$K) with fixed $(\kappa, \lamstar)$ (seed 4114).} Both rules share the $80\%$ teacher at each resolution and are evaluated on the official 24-view Mip-NeRF~360 test split (native LiteGS SSIM) without retuning. $U_{\rm sqrt} = \util$ includes the $1/\sqrt{W_i H_i}$ factor, while $U_{\rm raw} = \sum_i \covw u_{g,i}$ omits it.}
\label{tab:supp-garden-frozen21}
\footnotesize
\setlength{\tabcolsep}{4.2pt}
\begin{tabular}{@{}lll rrrrrr@{}}
\toprule
Scale & Loaded pixels & Selection rule & $\tcap$(M) & $\numg$(M) & Kept & PSNR$\uparrow$ & SSIM$\uparrow$ & LPIPS$\downarrow$ \\
\midrule
0.65K & 648 $\times$ 420 & \method{} ($U_{\rm sqrt}$) & 0.55 & 0.45 & 84.0\% & \textbf{29.52} & \textbf{0.916} & \textbf{0.075} \\
0.65K & 648 $\times$ 420 & Pixel sum ($U_{\rm raw}$) & 0.55 & 0.40 & 75.1\% & 29.50 & 0.915 & 0.077 \\
\midrule
1.3K & 1297 $\times$ 840 & \method{} ($U_{\rm sqrt}$) & 1.10 & 0.92 & 87.0\% & \textbf{27.68} & \textbf{0.862} & \textbf{0.131} \\
1.3K & 1297 $\times$ 840 & Pixel sum ($U_{\rm raw}$) & 1.10 & 0.92 & 86.7\% & \textbf{27.68} & \textbf{0.862} & \textbf{0.131} \\
\midrule
2.6K & 2594 $\times$ 1681 & \method{} ($U_{\rm sqrt}$) & 2.19 & 1.77 & 83.2\% & \textbf{26.98} & \textbf{0.822} & 0.178 \\
2.6K & 2594 $\times$ 1681 & Pixel sum ($U_{\rm raw}$) & 2.19 & 1.88 & 88.8\% & \textbf{26.98} & \textbf{0.822} & \textbf{0.177} \\
\midrule
5.2K & 5187 $\times$ 3361 & \method{} ($U_{\rm sqrt}$) & 4.39 & 3.13 & 73.9\% & 27.03 & 0.820 & 0.224 \\
5.2K & 5187 $\times$ 3361 & Pixel sum ($U_{\rm raw}$) & 4.39 & 3.69 & 87.2\% & \textbf{27.04} & \textbf{0.821} & \textbf{0.221} \\
\bottomrule
\end{tabular}
\end{table}

\FloatBarrier

\subsection{Self-distillation timing}
\Cref{tab:supp-consolidation-timing} varies the self-distillation point $\tau \in \{0.60, 0.70, 0.80, 0.90\}$ on four standard scenes (adjusting $\lambda$ at each $\tau$ to match the same $0.66$M mean final count) and on two large captures with fixed $\lamstar$.
On the standard scenes, PSNR changes by only $0.024$\,dB between $\tau = 0.60$ and $\tau = 0.80$, then dips slightly at $\tau = 0.90$ when only $10\%$ of training remains for refinement.
Distilling earlier at $\tau = 0.60$ reduces mean training time by $20.5\%$ relative to the default $\tau = 0.80$.

\begin{table}[!htbp]
\centering
\caption{\textbf{Self-distillation timing sensitivity ($\tau$, seed 4114).} Evaluated across four count-matched standard scenes (\emph{Garden}, \emph{Room}, \emph{Bonsai}, \emph{Truck}) and two large captures without retuning (\emph{Rubble}, \emph{London}).}
\label{tab:supp-consolidation-timing}
\footnotesize
\setlength{\tabcolsep}{4.5pt}
\begin{tabular}{@{}l rrrr rrrr r@{}}
\toprule
 & \multicolumn{4}{c}{\textbf{4 Standard (count-matched)}} & \multicolumn{4}{c}{\textbf{2 Large (without retuning)}} & \textbf{Mean time} \\
\cmidrule(lr){2-5}\cmidrule(lr){6-9}
Self-distillation $\tau$ & $\numg$(M) & PSNR$\uparrow$ & SSIM$\uparrow$ & LPIPS$\downarrow$ & $\numg$(M) & PSNR$\uparrow$ & SSIM$\uparrow$ & LPIPS$\downarrow$ & (s) \\
\midrule
$0.60$ & 0.66 & 29.961 & \textbf{0.910} & \textbf{0.159} & 3.54 & \textbf{26.228} & \underline{0.807} & 0.290 & 297 \\
$0.70$ & 0.66 & \underline{29.963} & \underline{0.909} & \underline{0.162} & 3.64 & \underline{26.226} & \textbf{0.808} & \underline{0.286} & 289 \\
$0.80$ (default) & 0.66 & \textbf{29.985} & \underline{0.909} & 0.163 & 3.85 & 26.100 & \textbf{0.808} & \textbf{0.285} & 373 \\
$0.90$ & 0.66 & 29.931 & 0.908 & 0.164 & 4.01 & 25.993 & 0.802 & 0.294 & 330 \\
\bottomrule
\end{tabular}
\end{table}

\clearpage
\section{Scene Coverage Studies}
\label{sec:supp-proxy-validation}

\subsection{Transformation invariances and camera redundancy}
\Cref{tab:supp-geometric-invariance} summarizes the 36 transformation checks, and \cref{tab:supp-room-frozen21} reports how duplicate views and a camera burst affect the exploration cap and final Gaussian count on \emph{Room}.

\begin{table}[!htbp]
\centering
\caption{\textbf{Empirical verification of $\capsupp$ across 36 capture transformations and 2 camera-burst stress tests.} All 36 exact invariance and scaling checks hold within $< 5 \times 10^{-16}$ relative error; camera bursts exhibit bounded growth (\textbf{P2}).}
\label{tab:supp-geometric-invariance}
\footnotesize
\setlength{\tabcolsep}{2.5pt}
\begin{tabular}{@{}ll rrr@{}}
\toprule
Transformation family & Tested captures & Cases & Predicted $\capsupp'/\capsupp$ & Measured $\capsupp'/\capsupp$ (max rel.\ err.) \\
\midrule
Exact $2\times$ view duplication (\textbf{P1}) & \emph{Garden}, \emph{Kitchen}, \emph{London} & 3 & $1$ & $1.000000$ ($< 5 \times 10^{-16}$) \\
Disjoint union $A \sqcup B$ (\textbf{P3}) & \emph{Bonsai+Bicycle}, \emph{Room+London}, \emph{Truck+Building} & 3 & $1^{\dagger}$ & $1.000000$ ($< 5 \times 10^{-16}$) \\
Uniform $2\times$ image dimensions (\textbf{P5}) & \emph{Garden}, \emph{Berlin} & 2 & $2$ & $2.000000$ ($< 5 \times 10^{-16}$) \\
Random 3D rotations $\mathbf{Q} \in \mathrm{SO}(3)$ & \emph{Bonsai} ($10$), \emph{Sci-Art} ($10$) & 20 & $1$ & $1.000000$ ($< 5 \times 10^{-16}$) \\
3D coordinate reflections & \emph{Bonsai}, \emph{Sci-Art} & 2 & $1$ & $1.000000$ ($< 5 \times 10^{-16}$) \\
Positive similarity scaling ($a > 0$) & \emph{Bonsai}, \emph{Sci-Art} & 2 & $1$ & $1.000000$ ($< 5 \times 10^{-16}$) \\
Held-out test-split additions / removals & \emph{Garden} ($2$), \emph{Building} ($2$) & 4 & $1$ & $1.000000$ ($< 5 \times 10^{-16}$) \\
\midrule
$+200$-frame camera burst (\textbf{P2}) & \emph{Room} ($+73.5\%$ views), \emph{London} ($+12.2\%$ views) & 2 & $O(1)$ & $1.0388$ ($+3.88\%$), $1.0052$ ($+0.52\%$) \\
\bottomrule
\multicolumn{5}{@{}l}{\scriptsize $^{\dagger}$Ratio $\capsupp(A \sqcup B) / (\capsupp(A) + \capsupp(B))$.} \\
\end{tabular}
\end{table}

\begin{table}[!htbp]
\centering
\caption{\textbf{Camera-redundancy training test on \emph{Room} (seed 4114).} Unweighted ($w_i = 1$) is calibrated to match \emph{Room}'s original exploration cap and retained count, and all runs are evaluated on the same 39 held-out views with fixed $\lamstar$.}
\label{tab:supp-room-frozen21}\label{tab:supp-exact-duplication}
\footnotesize
\setlength{\tabcolsep}{4.5pt}
\begin{tabular}{@{}ll rrrrrr@{}}
\toprule
Capture & Controller & $\tcap$(M) & $\numg$(M) & $\Delta\numg$ & PSNR$\uparrow$ & SSIM$\uparrow$ & LPIPS$\downarrow$ \\
\midrule
Original (272 views) & \method{} (ours) & 2.01 & 0.40 & +0.0\% & 32.52 & 0.926 & 0.208 \\
Original (272 views) & Unweighted ($w_i = 1$) & 2.01 & 0.40 & +0.0\% & \textbf{32.57} & \textbf{0.927} & \textbf{0.206} \\
\midrule
$2\times$ duplication (544 views) & \method{} (ours) & 2.01 & 0.39 & -1.1\% & 32.49 & 0.922 & 0.215 \\
$2\times$ duplication (544 views) & Unweighted ($w_i = 1$) & 4.03 & 0.49 & +22.9\% & \textbf{32.51} & \textbf{0.924} & \textbf{0.210} \\
\midrule
$+200$ burst (472 views) & \method{} (ours) & 2.09 & 0.38 & -4.9\% & 31.95 & \textbf{0.919} & \textbf{0.221} \\
$+200$ burst (472 views) & Unweighted ($w_i = 1$) & 3.49 & 0.41 & +3.5\% & \textbf{31.97} & 0.915 & 0.229 \\
\bottomrule
\end{tabular}
\end{table}

\FloatBarrier

\subsection{Per-scene coverage statistics and voxel-size sensitivity}
\Cref{tab:supp-support-proxy} lists the training view count $N$, effective view count $N_{\mathrm{eff}} = \sum_{i=1}^{N} \covw$, scene coverage measure $\capsupp$, and exploration cap $\tcap$ across all 21 scenes, together with sensitivity to voxel width $\beta \in \{0.015, 0.020, 0.030\}$.
Changing $\beta$ changes the bin size (shown under fixed $\kappa$), which mostly acts as a global scale factor that is absorbed when $\kappa$ is recalibrated on Std-13.
After recalibration, the largest exploration-cap change per scene has a median of $4.2\%$ across the 21 scenes.

\begin{table}[!htbp]
\centering
\caption{\textbf{Per-scene coverage statistics and voxel-size sensitivity across all 21 scenes.} Training views $N$, effective view count $N_{\mathrm{eff}} = \sum_i \covw$, scene coverage measure $\capsupp$, exploration cap $\tcap = \operatorname{round}(\kappa \capsupp)$, and percentage shifts in the unrounded exploration cap $\kappa\capsupp$ under $\beta \in \{0.015, 0.030\}$ relative to default $\beta = 0.020$ (with Std-13 recalibration vs.\ fixed $\kappa$).}
\label{tab:supp-support-proxy}\label{tab:supp-support-points}
\footnotesize
\setlength{\tabcolsep}{4pt}
\begin{tabular}{@{}l rrrr rr rr@{}}
\toprule
 & & & & & \multicolumn{2}{c}{\textbf{Recalibrated $\Delta(\kappa\capsupp)$ (\%)}} & \multicolumn{2}{c}{\textbf{Fixed-$\kappa$ $\Delta(\kappa\capsupp)$ (\%)}} \\
\cmidrule(lr){6-7}\cmidrule(lr){8-9}
Capture & Views $N$ & $N_{\mathrm{eff}}$ & $\capsupp$ ($\times 10^3$) & $\tcap$ (M) & $\beta=0.015$ & $\beta=0.030$ & $\beta=0.015$ & $\beta=0.030$ \\
\midrule
Bicycle & 169 & 29.1 & 29.4 & 1.82 & +2.1 & -1.2 & +9.4 & -9.6 \\
Bonsai & 255 & 27.5 & 35.0 & 2.17 & -4.2 & +3.1 & +2.7 & -5.6 \\
Counter & 210 & 28.7 & 36.5 & 2.26 & +2.0 & -3.5 & +9.4 & -11.7 \\
Flowers & 151 & 26.9 & 27.4 & 1.70 & -1.9 & +0.0 & +5.1 & -8.5 \\
Garden & 161 & 16.9 & 17.7 & 1.10 & +6.0 & -6.8 & +13.6 & -14.7 \\
Kitchen & 244 & 17.7 & 22.5 & 1.40 & +4.8 & -3.3 & +12.3 & -11.5 \\
Room & 272 & 25.5 & 32.4 & 2.01 & +3.9 & -2.4 & +11.4 & -10.7 \\
Stump & 109 & 25.9 & 26.3 & 1.63 & -3.1 & +3.4 & +3.9 & -5.3 \\
Treehill & 123 & 29.1 & 29.8 & 1.85 & -6.5 & +10.4 & +0.3 & +1.0 \\
Truck & 219 & 30.4 & 22.3 & 1.38 & -4.1 & +7.3 & +2.8 & -1.8 \\
Train & 263 & 33.4 & 24.4 & 1.51 & +2.6 & -3.4 & +10.0 & -11.6 \\
Dr Johnson & 230 & 57.7 & 62.3 & 3.86 & +0.1 & -0.6 & +7.3 & -9.0 \\
Playroom & 196 & 40.9 & 42.0 & 2.60 & -0.8 & -1.9 & +6.3 & -10.2 \\
\midrule
Rubble & 1657 & 103.6 & 103.3 & 6.41 & +15.2 & -14.1 & +23.5 & -21.4 \\
Building & 1920 & 142.2 & 141.8 & 8.80 & +6.0 & -1.2 & +13.6 & -9.5 \\
Residence & 2561 & 112.5 & 125.7 & 7.80 & +15.9 & -17.5 & +24.2 & -24.5 \\
Sci-Art & 2998 & 181.1 & 190.7 & 11.83 & +11.2 & -13.9 & +19.2 & -21.2 \\
Alameda & 1517 & 150.2 & 157.9 & 9.79 & +1.7 & -9.4 & +9.0 & -17.1 \\
Berlin & 1322 & 113.0 & 136.5 & 8.47 & +1.7 & -2.8 & +9.0 & -11.0 \\
London & 1639 & 178.4 & 187.5 & 11.63 & +3.6 & -5.9 & +11.0 & -13.9 \\
NYC & 866 & 110.9 & 116.6 & 7.23 & +2.4 & -2.0 & +9.7 & -10.3 \\
\bottomrule
\end{tabular}
\end{table}

\FloatBarrier

\subsection{Alternative pre-training capacity signals and SfM point density}
\Cref{tab:supp-s-proxies} tests replacing $\capsupp$ with SfM point count ($\kappa_M M$) or weighted pixel area ($\kappa_A \sum_i \covw W_i H_i$), each calibrated to the same Std-13 geometric-mean exploration cap ($1.85$M) and paired with fixed $\util > \lamstar$ self-distillation.
Scaling $\tcap$ with $M$ pushes texture-rich \emph{Rubble} to $\tcap = 34.30$M ($49.88$\,GiB peak VRAM vs.\ $13.60$\,GiB for \method{}) while leaving \emph{Garden} $5.2$K at $\tcap = 2.81$M.
Weighted pixel area matches $1.3$K captures closely, but grows quadratically with image dimensions and gives $\tcap = 17.57$M on \emph{Garden} $5.2$K compared with $4.39$M for \method{}.
\Cref{tab:supp-sfm-sparsity} tests the opposite direction by uniformly subsampling SfM points and 2D tracks to $50\%$ and $25\%$ before computing $\capsupp$ and initializing Gaussians.
At $25\%$, an SfM-proportional cap would drop by $75\%$, whereas \method{} raises $\tcap$ moderately (since sparser tracks share fewer bins) and changes four-scene mean PSNR by $-0.04$\,dB ($28.07$ vs.\ $28.11$\,dB).

\begin{table}[!htbp]
\centering
\caption{\textbf{Training runs with alternative pre-training exploration caps (seed 4114).} Each signal is calibrated to the same Std-13 geometric-mean exploration cap ($1.85$M), paired with fixed $\util > \lamstar$ self-distillation, and evaluated with native LiteGS SSIM.}
\label{tab:supp-s-proxies}
\footnotesize
\setlength{\tabcolsep}{4.5pt}
\begin{tabular}{@{}ll rrrrrr@{}}
\toprule
Exploration-cap signal & Capture & $\tcap$(M) & $\numg$(M) & PSNR$\uparrow$ & SSIM$\uparrow$ & LPIPS$\downarrow$ & Peak GiB \\
\midrule
SfM points ($\kappa_M M$) & Bonsai & 4.18 & 0.45 & 32.73 & 0.944 & 0.201 & 7.30 \\
SfM points ($\kappa_M M$) & Garden (1.3K) & 2.81 & 1.59 & 28.03 & 0.878 & 0.101 & 4.24 \\
SfM points ($\kappa_M M$) & Garden (5.2K) & 2.81 & 2.28 & 26.91 & 0.815 & 0.238 & 15.73 \\
SfM points ($\kappa_M M$) & London & 4.14 & 2.35 & 26.97 & 0.846 & 0.304 & 11.41 \\
SfM points ($\kappa_M M$) & Room & 2.28 & 0.38 & 32.40 & 0.925 & 0.212 & 4.85 \\
SfM points ($\kappa_M M$) & Rubble & 34.30 & 6.50 & 25.79 & 0.784 & 0.256 & 49.88 \\
\midrule
Weighted pixel area ($\kappa_A \sum_i \covw W_i H_i$) & Garden (1.3K) & 1.10 & 0.92 & 27.65 & 0.862 & 0.131 & 2.07 \\
Weighted pixel area ($\kappa_A \sum_i \covw W_i H_i$) & London & 11.71 & 2.87 & 26.81 & 0.844 & 0.300 & 21.33 \\
Weighted pixel area ($\kappa_A \sum_i \covw W_i H_i$) & Room & 2.45 & 0.37 & 32.48 & 0.924 & 0.215 & 5.08 \\
Weighted pixel area ($\kappa_A \sum_i \covw W_i H_i$) & Rubble & 6.13 & 4.71 & 25.48 & 0.772 & 0.272 & 13.25 \\
\bottomrule
\end{tabular}
\end{table}

\begin{table}[!htbp]
\centering
\caption{\textbf{Robustness to SfM point density (\textbf{P4}, seed 4114).} We uniformly subsample $100\%$, $50\%$, and $25\%$ of SfM points and 2D tracks before computing $\capsupp$ and initializing Gaussians, keeping $(\kappa, \lamstar)$ fixed (native LiteGS SSIM).}
\label{tab:supp-sfm-sparsity}
\footnotesize
\setlength{\tabcolsep}{4.5pt}
\begin{tabular}{@{}l rrrrrrr@{}}
\toprule
Capture & Kept SfM & 3D points & $\tcap$(M) & $\numg$(M) & PSNR$\uparrow$ & SSIM$\uparrow$ & LPIPS$\downarrow$ \\
\midrule
Garden & 100\% & 138,766 & 1.10 & 0.92 & \underline{27.70} & 0.862 & 0.131 \\
Garden & 50\% & 69,383 & 1.22 & 1.01 & 27.68 & \underline{0.864} & \underline{0.127} \\
Garden & 25\% & 34,692 & 1.42 & 1.14 & \textbf{27.71} & \textbf{0.866} & \textbf{0.121} \\
\midrule
Room & 100\% & 112,627 & 2.01 & 0.39 & \textbf{32.55} & 0.926 & 0.208 \\
Room & 50\% & 56,314 & 2.21 & 0.42 & \underline{32.53} & \textbf{0.927} & \underline{0.207} \\
Room & 25\% & 28,157 & 2.50 & 0.44 & 32.38 & 0.926 & \textbf{0.206} \\
\midrule
Rubble & 100\% & 1,694,315 & 6.41 & 4.86 & 25.37 & 0.772 & 0.270 \\
Rubble & 50\% & 847,158 & 7.78 & 5.69 & \underline{25.56} & \underline{0.779} & \underline{0.263} \\
Rubble & 25\% & 423,579 & 9.78 & 6.82 & \textbf{25.68} & \textbf{0.784} & \textbf{0.253} \\
\midrule
London & 100\% & 204,776 & 11.63 & 2.84 & \textbf{26.83} & \textbf{0.844} & \textbf{0.300} \\
London & 50\% & 102,388 & 12.83 & 3.01 & \underline{26.77} & \underline{0.842} & \textbf{0.300} \\
London & 25\% & 51,194 & 14.49 & 3.16 & 26.51 & 0.840 & 0.303 \\
\bottomrule
\end{tabular}
\end{table}

\clearpage
\section{Extended Related Work and Baseline Retuning}
\label{sec:supp-reported}

\subsection{Additional related work and controller taxonomy}
Several additional methods study capacity scheduling and local densification rules beyond those discussed in the main paper.
Among capacity controllers, Mini-Splatting~\citep{fang2024minisplatting} reorganizes Gaussians during densification and subsamples them with a preset fraction.
LP-3DGS~\citep{zhang2024lp3dgs} learns $L_p$-regularized masks over importance scores, while ControlGS~\citep{controlgs2025} combines uniform splitting with global opacity regularization.
Prune Wisely~\citep{wang2026prunewisely} and Smart Target Point Control~\citep{bisht2026smarttarget} schedule pruning or growth toward a user-supplied target count, and SPARE-GS~\citep{chen2026sparegs} divides a count-based budget into regional quotas.
Other methods adjust the growth schedule itself: SIG~\citep{xue2026sig} ties image resolution and model growth to frequency-domain progress, GS$^2$~\citep{yang2026gs2} stops densifying when added complexity outweighs loss reduction, BlitzGS~\citep{wang2026blitzgs} combines importance pruning with density feedback, and RLGS~\citep{li2025rlgs} learns schedule updates on top of budgeted controllers.

A separate line of work refines \emph{where} primitives are added or pruned within a budget using per-view error maps~\citep{bulo2024revising,conegs2025}, gradient direction or homodirectional aggregation~\citep{ye2024absgs,zhou2026gdags}, second-order curvature~\citep{wang2025steepgs}, perceptual sensitivity~\citep{zhou2025perceptualgs}, cross-view consistency~\citep{ren2026fastgs}, or local image frequency relative to primitive density~\citep{dong2025cdcgs}.
\Cref{tab:taxonomy} summarizes how the 13 baselines we evaluate set capacity; each still depends on gradient growth thresholds, fixed fractions, preset counts, or sparsity penalties whose behavior shifts across capture scales.

For large-scale scenes, VastGaussian~\citep{lin2024vastgaussian} and CityGaussian~\citep{liu2024citygaussian} partition captures into spatial blocks for training, while Grendel~\citep{zhao2025grendel} distributes a single scene model across GPUs by partitioning both the Gaussian parameters and the image-space rendering work.

\begin{table}[!htbp]
\centering
\caption{\textbf{Candidate retuning controls for large captures.} Scale-mechanism groups and primary capacity controls of the 13 evaluated baselines compared with \method{}.}
\label{tab:taxonomy}
\footnotesize
\setlength{\tabcolsep}{3.5pt}
\begin{tabular}{@{}>{\raggedright\arraybackslash}p{0.24\linewidth}>{\raggedright\arraybackslash}p{0.30\linewidth}>{\raggedright\arraybackslash}p{\dimexpr0.46\linewidth-14pt\relax}@{}}
\toprule
Method & Controls to revisit & Calibration issue \\
\midrule
\multicolumn{3}{@{}l}{\emph{Fixed constant}} \\
vanilla 3DGS~\citep{kerbl3Dgaussians} & Gradient growth threshold & Threshold has no direct count or quality interpretation. \\
\addlinespace[3pt]
Speedy-Splat~\citep{hanson2025speedysplat} & Gradient growth threshold; pruning fraction & Retention depends on the Gaussians admitted by growth. \\
\midrule
\multicolumn{3}{@{}l}{\emph{Preset count}} \\
Taming 3DGS~\citep{tamings3dgs} & SfM multiplier; gradient growth threshold & A larger budget need not admit more growth candidates. \\
\addlinespace[3pt]
Mini-Splatting~\citep{fang2024minisplatting} & Count limits; sampling fraction & Capacity before simplification and final retention both need selection. \\
\addlinespace[3pt]
\mbox{3DGS-MCMC}~\citep{kheradmand2024mcmc} & Target count; opacity regularization & Raising the target does not remove the opacity penalty. \\
\addlinespace[3pt]
LiteGS~\citep{litegs} & Target count & Default target can fall below initialization; useful capacity is unknown. \\
\midrule
\multicolumn{3}{@{}l}{\emph{Automatic budget}} \\
DashGaussian~\citep{chen2025dashgaussian} & Gradient growth threshold; budget estimation & Estimated budgets still depend on this threshold. \\
\addlinespace[3pt]
Compact 3DGS~\citep{lee2024compact} & Mask regularization & Penalty has no direct retained-count or quality interpretation. \\
\addlinespace[3pt]
LP-3DGS~\citep{zhang2024lp3dgs} & Mask regularization & Learned retention still depends on the chosen penalty. \\
\addlinespace[3pt]
ControlGS~\citep{controlgs2025} & Opacity regularization & One global control still needs calibration to the capture. \\
\addlinespace[3pt]
CDC-GS~\citep{dong2025cdcgs} & Gradient growth threshold; count fractions & Local adaptation retains threshold and fraction choices. \\
\addlinespace[3pt]
FastGS~\citep{ren2026fastgs} & Opacity and consistency thresholds & Final pruning depends on both threshold choices. \\
\addlinespace[3pt]
LeGS~\citep{ning2026legs} & Gradient growth and opacity thresholds & Policy acts within the candidates passing these thresholds. \\
\midrule
\textbf{\method{} (ours)} & Exploration cap and self-distillation threshold & Global constants; no per-capture selection in our evaluation. \\
\bottomrule
\end{tabular}
\end{table}

\FloatBarrier

\subsection{Controlled manual retuning on large captures}
\Cref{tab:supp-manual-retuning} tests whether simple hyperparameter sweeps close the large-capture gap without per-scene trial and error.
Sweeping the gradient growth threshold of vanilla 3DGS over $\{0.25, 0.5, 1, 2, 4\}\times$ on held-out validation splits gives the highest PSNR at $1\times$ on \emph{NYC} ($27.19$\,dB) and $2\times$ on \emph{Residence} ($21.27$\,dB). The $0.25\times$ setting reduces PSNR on both ($21.28$\,dB on \emph{NYC}; $17.89$\,dB at $13.13$M on \emph{Residence}).
On \emph{London}, Taming~3DGS stays at $0.32$M across all three budget multipliers ($2\times, 5\times, 15\times$) unless its gradient growth threshold is also lowered $8\times$ to $2.5 \times 10^{-5}$ ($3.07$M, $26.66$\,dB).
Compact~3DGS reaches $25.49$\,dB on \emph{London} even after setting its mask penalty to zero and halving its gradient growth threshold, still below \method{} without retuning ($26.81$\,dB).

\begin{table}[!htbp]
\centering
\caption{\textbf{Controlled manual retuning sweeps on large captures ($\mathbf{60}$k iterations).} Top: vanilla 3DGS validation-split gradient growth threshold sweep on \emph{NYC} and \emph{Residence} (seed 4114). Middle: 18-run Taming~3DGS grid on \emph{London} ($3$ seeds $\times$ $3$ multipliers $\times$ $2$ thresholds, $30$k position-LR horizon). Bottom: Compact~3DGS mask-penalty and gradient growth threshold sweep on \emph{London} (seed 4114). Both London sweeps report normalized VGG LPIPS. Without retuning, \method{} reaches $28.27$\,dB (\emph{NYC}), $22.08$\,dB (\emph{Residence}), and $26.81$\,dB (\emph{London}).}
\label{tab:supp-manual-retuning}
\footnotesize
\setlength{\tabcolsep}{4.5pt}
\begin{tabular}{@{}lll rrr@{}}
\toprule
Method & Capture & Intervention & $\numg$ (M) & PSNR$\uparrow$ & LPIPS$\downarrow$ \\
\midrule
vanilla 3DGS & NYC & grad.\ thresh.\ $5\times 10^{-5}$ ($0.25\times$) & 2.89 & 21.28 & 0.454 \\
vanilla 3DGS & NYC & grad.\ thresh.\ $10^{-4}$ ($0.5\times$) & 2.35 & \underline{27.01} & \textbf{0.365} \\
vanilla 3DGS & NYC & grad.\ thresh.\ $2\times 10^{-4}$ (default) & 1.12 & \textbf{27.19} & \underline{0.376} \\
vanilla 3DGS & NYC & grad.\ thresh.\ $4\times 10^{-4}$ ($2\times$) & 0.49 & 26.73 & 0.402 \\
vanilla 3DGS & NYC & grad.\ thresh.\ $8\times 10^{-4}$ ($4\times$) & 0.25 & 26.18 & 0.427 \\
vanilla 3DGS & Residence & grad.\ thresh.\ $5\times 10^{-5}$ ($0.25\times$) & 13.13 & 17.89 & 0.364 \\
vanilla 3DGS & Residence & grad.\ thresh.\ $10^{-4}$ ($0.5\times$) & 15.09 & 20.82 & \textbf{0.261} \\
vanilla 3DGS & Residence & grad.\ thresh.\ $2\times 10^{-4}$ (default) & 8.05 & 21.12 & \underline{0.268} \\
vanilla 3DGS & Residence & grad.\ thresh.\ $4\times 10^{-4}$ ($2\times$) & 3.57 & \textbf{21.27} & 0.290 \\
vanilla 3DGS & Residence & grad.\ thresh.\ $8\times 10^{-4}$ ($4\times$) & 1.84 & \underline{21.15} & 0.313 \\
\midrule
Taming 3DGS (3-seed) & London & $2\times$ budget, default $2\!\times\!10^{-4}$ & 0.32 & 25.40 & 0.468 \\
Taming 3DGS (3-seed) & London & $2\times$ budget, lowered $2.5\!\times\!10^{-5}$ & 0.41 & 25.67 & 0.458 \\
Taming 3DGS (3-seed) & London & $5\times$ budget, default $2\!\times\!10^{-4}$ & 0.32 & 25.40 & 0.468 \\
Taming 3DGS (3-seed) & London & $5\times$ budget, lowered $2.5\!\times\!10^{-5}$ & 1.02 & \underline{26.25} & \underline{0.425} \\
Taming 3DGS (3-seed) & London & $15\times$ budget, default $2\!\times\!10^{-4}$ & 0.32 & 25.42 & 0.468 \\
Taming 3DGS (3-seed) & London & $15\times$ budget, lowered $2.5\!\times\!10^{-5}$ & 3.07 & \textbf{26.66} & \textbf{0.396} \\
\midrule
Compact 3DGS & London & $\lambda_{\rm mask}=5\times 10^{-4}$, default grad.\ thresh. & 0.17 & 25.04 & 0.485 \\
Compact 3DGS & London & $\lambda_{\rm mask}=5\times 10^{-5}$, default grad.\ thresh. & 0.27 & 25.20 & 0.474 \\
Compact 3DGS & London & $\lambda_{\rm mask}=0$, default grad.\ thresh. & 0.52 & \underline{25.26} & \underline{0.470} \\
Compact 3DGS & London & $\lambda_{\rm mask}=0$, grad.\ thresh.\ $10^{-4}$ & 1.28 & \textbf{25.49} & \textbf{0.448} \\
\bottomrule
\end{tabular}
\end{table}

\clearpage
\section{Portability to Taming 3DGS}
\label{sec:supp-second-base}

To test portability beyond LiteGS, we integrate \method{} into Taming~3DGS~\citep{tamings3dgs} while keeping the LiteGS-calibrated parameters $(\kappa = 62.04,\; \lamstar = 1.0861 \times 10^{-5})$ unchanged across all 21 scenes.
Over the first $80\%$ of training, the port keeps Taming~3DGS's quadratic growth ramp, sampling parents by opacity-gradient variance scores with retained split parents, opacity decay, and sigmoid opacity.
At $80\%$, it applies one step of coverage-weighted self-distillation ($\util > \lamstar$), without $128$-cluster rounding since Taming~3DGS does not group primitives into clusters.
It then refines the surviving Gaussians over the final $20\%$ of training, and we evaluate with Taming~3DGS's reference PNG evaluator.

\Cref{tab:supp-second-base} shows that \method{} cuts the Std-13 mean Gaussian count by $65.0\%$ ($1.03$M vs.\ $2.93$M) while slightly improving mean PSNR ($27.77$ vs.\ $27.63$\,dB).
On Large-8, it raises mean capacity to $3.43$M, improving PSNR by $+0.97$\,dB and lowering LPIPS by $0.091$ with gains on all eight captures.

\begin{table}[!htbp]
\centering
\caption{\textbf{Portability of \method{} to Taming~3DGS across all 21 scenes (seed 4114).} Unchanged LiteGS parameters ($\kappa = 62.04$, $\lamstar = 1.0861 \times 10^{-5}$) evaluated with Taming~3DGS's reference PNG evaluator against released Taming~3DGS (dataset presets on Std-13; $15\times$ SfM points on Large-8).}
\label{tab:supp-second-base}
\footnotesize
\setlength{\tabcolsep}{4pt}
\begin{tabular}{@{}l rrrr rrrr@{}}
\toprule
 & \multicolumn{4}{c}{\textbf{Released Taming 3DGS}} & \multicolumn{4}{c}{\textbf{Taming 3DGS + \method{} (ours)}} \\
\cmidrule(lr){2-5}\cmidrule(lr){6-9}
Dataset / Capture & PSNR$\uparrow$ & SSIM$\uparrow$ & LPIPS$\downarrow$ & $\numg$(M) & PSNR$\uparrow$ & SSIM$\uparrow$ & LPIPS$\downarrow$ & $\numg$(M) \\
\midrule
Mip-NeRF 360 & 27.83 & 0.822 & 0.204 & 3.21 & \textbf{28.00} & \textbf{0.827} & \textbf{0.193} & 1.09 \\
Tanks \& Temples & \textbf{24.21} & 0.855 & 0.169 & 1.83 & 24.12 & \textbf{0.865} & \textbf{0.140} & 0.93 \\
Deep Blending & 30.17 & 0.910 & 0.234 & 2.80 & \textbf{30.40} & \textbf{0.911} & \textbf{0.228} & 0.85 \\
\midrule
Rubble & 25.18 & 0.730 & 0.334 & 2.89 & \textbf{25.27} & \textbf{0.771} & \textbf{0.270} & 4.25 \\
Building & 20.69 & 0.679 & 0.348 & 2.11 & \textbf{21.42} & \textbf{0.745} & \textbf{0.262} & 4.68 \\
Residence & 20.59 & 0.709 & 0.315 & 2.93 & \textbf{21.91} & \textbf{0.792} & \textbf{0.219} & 3.87 \\
Sci-Art & 21.83 & 0.785 & 0.267 & 2.05 & \textbf{22.52} & \textbf{0.864} & \textbf{0.173} & 4.33 \\
Alameda & 22.83 & 0.719 & 0.422 & 0.43 & \textbf{23.07} & \textbf{0.773} & \textbf{0.326} & 3.53 \\
London & 25.42 & 0.792 & 0.407 & 0.30 & \textbf{27.04} & \textbf{0.855} & \textbf{0.286} & 2.98 \\
NYC & 26.23 & 0.818 & 0.364 & 0.37 & \textbf{28.31} & \textbf{0.874} & \textbf{0.265} & 2.13 \\
Berlin & 27.42 & 0.884 & 0.312 & 0.42 & \textbf{28.42} & \textbf{0.918} & \textbf{0.241} & 1.70 \\
\midrule
\textbf{Std-13 mean} & 27.63 & 0.841 & 0.203 & 2.93 & \textbf{27.77} & \textbf{0.846} & \textbf{0.190} & 1.03 \\
\textbf{Large-8 mean} & 23.77 & 0.764 & 0.346 & 1.44 & \textbf{24.74} & \textbf{0.824} & \textbf{0.255} & 3.43 \\
\bottomrule
\end{tabular}
\end{table}

\clearpage
\section{Model Storage and Rendering Throughput}
\label{sec:supp-efficiency}

We report on-disk model size in decimal MB ($10^6$ bytes), measured from the exported PLY point cloud; for Compact~3DGS~\citep{lee2024compact}, this includes its learned color MLP, and we list its estimated quantized size below each table.
Rendering speed (FPS) is measured in each method's native rasterizer on one NVIDIA GH200 at evaluation resolution.
After two warm-up passes, we time ten full passes over the held-out views with CUDA synchronization (excluding disk I/O and metric computation) and divide the view count by the median pass duration.
\Cref{tab:supp-efficiency-standard_a,tab:supp-efficiency-standard_d,tab:supp-efficiency-standard_f} report per-scene PSNR, LPIPS, storage, and FPS on all 13 standard scenes.
\Cref{tab:supp-efficiency} reports storage and FPS on the eight large captures, with corresponding quality metrics and Gaussian counts in \cref{sec:supp-transfer} and cross-scale means in \cref{tab:main-benchmark}.

\begin{table}[!htbp]
\centering
\caption{\textbf{Per-scene quality, saved storage, and rendering throughput: Mip-NeRF~360 outdoor scenes ($\mathbf{30}$k iterations).} Saved model size in decimal MB ($10^6$ bytes) and native rasterizer FPS on one NVIDIA GH200; \method{} reports three-seed means.}
\label{tab:supp-efficiency-standard_a}\label{tab:supp-efficiency-standard_b}\label{tab:supp-efficiency-standard_c}
\scriptsize
\setlength{\tabcolsep}{1.4pt}
\begin{tabular}{@{}l rrrr rrrr rrrr rrrr rrrr@{}}
\toprule
 & \multicolumn{4}{c}{\textbf{Bicycle}} & \multicolumn{4}{c}{\textbf{Flowers}} & \multicolumn{4}{c}{\textbf{Garden}} & \multicolumn{4}{c}{\textbf{Stump}} & \multicolumn{4}{c}{\textbf{Treehill}} \\
\cmidrule(lr){2-5}\cmidrule(lr){6-9}\cmidrule(lr){10-13}\cmidrule(lr){14-17}\cmidrule(lr){18-21}
Method & PSNR$\uparrow$ & LPIPS$\downarrow$ & MB$\downarrow$ & FPS$\uparrow$ & PSNR$\uparrow$ & LPIPS$\downarrow$ & MB$\downarrow$ & FPS$\uparrow$ & PSNR$\uparrow$ & LPIPS$\downarrow$ & MB$\downarrow$ & FPS$\uparrow$ & PSNR$\uparrow$ & LPIPS$\downarrow$ & MB$\downarrow$ & FPS$\uparrow$ & PSNR$\uparrow$ & LPIPS$\downarrow$ & MB$\downarrow$ & FPS$\uparrow$ \\
\midrule
\multicolumn{21}{@{}l}{\emph{Fixed-constant controllers}} \\
vanilla 3DGS~\citep{kerbl3Dgaussians} & 25.20 & 0.208 & 1231 & 116 & 21.57 & 0.335 & 718 & 242 & 27.49 & 0.106 & 1026 & 162 & 26.64 & 0.215 & 1063 & 200 & 22.51 & 0.324 & 824 & 203 \\
Speedy-Splat~\citep{hanson2025speedysplat} & 24.85 & 0.304 & 152 & 842 & 21.42 & 0.394 & 91 & 1024 & 26.84 & 0.183 & 133 & 931 & 26.69 & 0.259 & 127 & 897 & 22.52 & 0.442 & 92 & 1176 \\
\midrule
\multicolumn{21}{@{}l}{\emph{Preset-count controllers}} \\
3DGS-MCMC~\citep{kheradmand2024mcmc} & \underline{25.69} & \underline{0.169} & 1463 & 107 & \textbf{22.05} & 0.292 & 718 & 151 & 27.81 & 0.096 & 1290 & 140 & 27.31 & \textbf{0.171} & 1178 & 118 & 22.93 & \textbf{0.275} & 824 & 162 \\
Taming 3DGS~\citep{tamings3dgs} & 25.50 & 0.195 & 1485 & 144 & 21.82 & 0.334 & 897 & 220 & 27.78 & 0.100 & 1273 & 191 & 26.97 & 0.201 & 1207 & 209 & 23.06 & 0.313 & 935 & 201 \\
Mini-Splatting~\citep{fang2024minisplatting} & 25.21 & 0.225 & 131 & 793 & 21.58 & 0.327 & 142 & 735 & 26.88 & 0.150 & 140 & 815 & 27.22 & 0.199 & 151 & 782 & 22.67 & 0.314 & 142 & 751 \\
LiteGS (1M)~\citep{litegs} & 24.84 & 0.263 & 246 & 673 & 21.71 & 0.315 & 246 & 476 & 27.55 & 0.134 & 246 & 786 & 27.13 & 0.214 & 247 & 612 & \textbf{23.37} & 0.323 & 247 & 600 \\
\midrule
\multicolumn{21}{@{}l}{\emph{Automatic-budget controllers}} \\
DashGaussian~\citep{chen2025dashgaussian} & 25.41 & 0.196 & 1166 & 185 & 21.94 & 0.322 & 664 & 270 & 27.67 & 0.111 & 814 & 254 & 27.09 & 0.205 & 890 & 274 & 23.05 & 0.308 & 873 & 235 \\
Compact 3DGS$^{\ast}$~\citep{lee2024compact} & 24.87 & 0.252 & 180 & 159 & 21.06 & 0.377 & 136 & 284 & 26.90 & 0.142 & 176 & 188 & 26.29 & 0.256 & 152 & 252 & 22.58 & 0.342 & 169 & 206 \\
LP-3DGS~\citep{zhang2024lp3dgs} & 25.17 & 0.216 & 583 & 277 & 21.50 & 0.343 & 387 & 413 & 27.38 & 0.110 & 598 & 308 & 26.61 & 0.219 & 499 & 375 & 22.42 & 0.331 & 420 & 380 \\
ControlGS~\citep{controlgs2025} & 25.37 & 0.234 & 237 & 149 & 21.76 & 0.309 & 359 & 133 & 27.17 & 0.156 & 225 & 186 & 26.62 & 0.210 & 262 & 168 & 22.53 & 0.321 & 240 & 148 \\
CDC-GS~\citep{dong2025cdcgs} & 25.48 & 0.170 & 1277 & 153 & 21.64 & \textbf{0.281} & 815 & 208 & \underline{27.82} & \underline{0.094} & 994 & 178 & 26.85 & 0.207 & 1032 & 218 & 22.46 & 0.295 & 1014 & 198 \\
FastGS~\citep{ren2026fastgs} & 24.81 & 0.296 & 119 & 1046 & 21.35 & 0.389 & 111 & 930 & 27.29 & 0.157 & 164 & 932 & 26.52 & 0.282 & 86 & 1026 & 22.85 & 0.416 & 90 & 1062 \\
LeGS~\citep{ning2026legs} & \textbf{25.73} & \textbf{0.167} & 840 & 446 & \underline{22.04} & \underline{0.282} & 651 & 477 & \textbf{28.09} & \textbf{0.089} & 836 & 400 & \underline{27.40} & \underline{0.177} & 569 & 505 & 23.00 & 0.303 & 629 & 529 \\
\midrule
\multicolumn{21}{@{}l}{\emph{Coverage-scaled exploration + self-distillation (ours)}} \\
\textbf{\method{}} & \textbf{25.73} & 0.198 & 353 & 576 & 21.96 & 0.285 & 360 & 403 & 27.65 & 0.131 & 229 & 859 & \textbf{27.51} & 0.184 & 304 & 602 & \underline{23.17} & \underline{0.289} & 378 & 585 \\
\bottomrule
\end{tabular}
\par\vspace{2pt}{\scriptsize $^{\ast}$Compact~3DGS reports saved checkpoint MB above; estimated quantized payload: Bicycle: 64 MB, Flowers: 51 MB, Garden: 63 MB, Stump: 56 MB, Treehill: 61 MB.}
\end{table}

\begin{table}[!htbp]
\centering
\caption{\textbf{Per-scene quality, saved storage, and rendering throughput: Mip-NeRF~360 indoor scenes ($\mathbf{30}$k iterations).}}
\label{tab:supp-efficiency-standard_d}\label{tab:supp-efficiency-standard_e}
\scriptsize
\setlength{\tabcolsep}{2.3pt}
\begin{tabular}{@{}l rrrr rrrr rrrr rrrr@{}}
\toprule
 & \multicolumn{4}{c}{\textbf{Room}} & \multicolumn{4}{c}{\textbf{Counter}} & \multicolumn{4}{c}{\textbf{Kitchen}} & \multicolumn{4}{c}{\textbf{Bonsai}} \\
\cmidrule(lr){2-5}\cmidrule(lr){6-9}\cmidrule(lr){10-13}\cmidrule(lr){14-17}
Method & PSNR$\uparrow$ & LPIPS$\downarrow$ & MB$\downarrow$ & FPS$\uparrow$ & PSNR$\uparrow$ & LPIPS$\downarrow$ & MB$\downarrow$ & FPS$\uparrow$ & PSNR$\uparrow$ & LPIPS$\downarrow$ & MB$\downarrow$ & FPS$\uparrow$ & PSNR$\uparrow$ & LPIPS$\downarrow$ & MB$\downarrow$ & FPS$\uparrow$ \\
\midrule
\multicolumn{17}{@{}l}{\emph{Fixed-constant controllers}} \\
vanilla 3DGS~\citep{kerbl3Dgaussians} & 31.54 & 0.205 & 323 & 198 & 29.07 & 0.191 & 268 & 208 & 31.31 & 0.122 & 392 & 178 & 32.43 & 0.190 & 264 & 288 \\
Speedy-Splat~\citep{hanson2025speedysplat} & 30.75 & 0.267 & 29 & 971 & 28.20 & 0.265 & 25 & 1072 & 29.84 & 0.198 & 29 & 1100 & 31.09 & 0.237 & 32 & 1122 \\
\midrule
\multicolumn{17}{@{}l}{\emph{Preset-count controllers}} \\
3DGS-MCMC~\citep{kheradmand2024mcmc} & 32.24 & 0.187 & 372 & 199 & 29.34 & \underline{0.178} & 298 & 183 & 31.80 & 0.118 & 446 & 168 & 32.67 & 0.178 & 322 & 213 \\
Taming 3DGS~\citep{tamings3dgs} & 32.05 & 0.199 & 384 & 206 & 29.04 & 0.190 & 295 & 218 & 31.83 & 0.118 & 401 & 202 & 32.43 & 0.188 & 294 & 246 \\
Mini-Splatting~\citep{fang2024minisplatting} & 31.05 & 0.201 & 97 & 661 & 28.40 & 0.191 & 102 & 509 & 31.19 & 0.126 & 106 & 552 & 31.30 & 0.186 & 90 & 615 \\
LiteGS (1M)~\citep{litegs} & \textbf{32.63} & \underline{0.183} & 245 & 516 & \textbf{29.84} & \textbf{0.170} & 246 & 459 & 32.37 & \underline{0.114} & 246 & 521 & \underline{33.17} & \underline{0.172} & 246 & 441 \\
\midrule
\multicolumn{17}{@{}l}{\emph{Automatic-budget controllers}} \\
DashGaussian~\citep{chen2025dashgaussian} & 31.68 & 0.212 & 289 & 217 & 28.97 & 0.200 & 210 & 232 & 31.40 & 0.134 & 310 & 186 & 32.04 & 0.191 & 210 & 259 \\
Compact 3DGS$^{\ast}$~\citep{lee2024compact} & 30.63 & 0.218 & 80 & 349 & 28.59 & 0.213 & 81 & 274 & 30.52 & 0.135 & 116 & 223 & 31.74 & 0.203 & 84 & 372 \\
LP-3DGS~\citep{zhang2024lp3dgs} & 31.42 & 0.212 & 110 & 504 & 28.88 & 0.197 & 109 & 456 & 31.38 & 0.125 & 199 & 381 & 32.03 & 0.196 & 116 & 544 \\
ControlGS~\citep{controlgs2025} & 32.11 & 0.200 & 121 & 131 & 29.74 & 0.179 & 165 & 98 & 31.81 & 0.121 & 159 & 126 & 32.70 & 0.187 & 125 & 140 \\
CDC-GS~\citep{dong2025cdcgs} & 32.12 & 0.185 & 277 & 200 & 29.26 & 0.181 & 210 & 213 & 31.92 & 0.118 & 239 & 206 & 32.15 & 0.184 & 245 & 254 \\
FastGS~\citep{ren2026fastgs} & 31.70 & 0.227 & 53 & 1086 & 29.07 & 0.212 & 53 & 926 & 31.65 & 0.132 & 97 & 789 & 32.07 & 0.200 & 69 & 979 \\
LeGS~\citep{ning2026legs} & \underline{32.62} & \textbf{0.181} & 259 & 577 & \underline{29.82} & \textbf{0.170} & 207 & 493 & \textbf{32.58} & \textbf{0.112} & 354 & 387 & \textbf{33.41} & \textbf{0.164} & 308 & 508 \\
\midrule
\multicolumn{17}{@{}l}{\emph{Coverage-scaled exploration + self-distillation (ours)}} \\
\textbf{\method{}} & 32.51 & 0.207 & 102 & 588 & 29.71 & \underline{0.178} & 162 & 605 & \underline{32.46} & 0.117 & 201 & 578 & 33.00 & 0.187 & 161 & 458 \\
\bottomrule
\end{tabular}
\par\vspace{2pt}{\scriptsize $^{\ast}$Compact~3DGS reports saved checkpoint MB above; estimated quantized payload: Room: 34 MB, Counter: 34 MB, Kitchen: 45 MB, Bonsai: 35 MB.}
\end{table}

\begin{table}[!htbp]
\centering
\caption{\textbf{Per-scene quality, saved storage, and rendering throughput: Tanks~\&~Temples and Deep Blending ($\mathbf{30}$k iterations).}}
\label{tab:supp-efficiency-standard_f}\label{tab:supp-efficiency-standard_g}
\scriptsize
\setlength{\tabcolsep}{2.3pt}
\begin{tabular}{@{}l rrrr rrrr rrrr rrrr@{}}
\toprule
 & \multicolumn{4}{c}{\textbf{Truck}} & \multicolumn{4}{c}{\textbf{Train}} & \multicolumn{4}{c}{\textbf{Dr Johnson}} & \multicolumn{4}{c}{\textbf{Playroom}} \\
\cmidrule(lr){2-5}\cmidrule(lr){6-9}\cmidrule(lr){10-13}\cmidrule(lr){14-17}
Method & PSNR$\uparrow$ & LPIPS$\downarrow$ & MB$\downarrow$ & FPS$\uparrow$ & PSNR$\uparrow$ & LPIPS$\downarrow$ & MB$\downarrow$ & FPS$\uparrow$ & PSNR$\uparrow$ & LPIPS$\downarrow$ & MB$\downarrow$ & FPS$\uparrow$ & PSNR$\uparrow$ & LPIPS$\downarrow$ & MB$\downarrow$ & FPS$\uparrow$ \\
\midrule
\multicolumn{17}{@{}l}{\emph{Fixed-constant controllers}} \\
vanilla 3DGS~\citep{kerbl3Dgaussians} & 25.46 & 0.142 & 512 & 231 & 22.24 & 0.195 & 272 & 225 & 29.35 & 0.236 & 775 & 159 & 30.07 & 0.241 & 456 & 235 \\
Speedy-Splat~\citep{hanson2025speedysplat} & 25.14 & 0.190 & 63 & 1100 & 21.59 & 0.292 & 26 & 1022 & 29.13 & 0.267 & 78 & 1130 & 30.14 & 0.269 & 46 & 1217 \\
\midrule
\multicolumn{17}{@{}l}{\emph{Preset-count controllers}} \\
3DGS-MCMC~\citep{kheradmand2024mcmc} & 26.45 & \textbf{0.110} & 645 & 159 & 22.61 & 0.182 & 273 & 280 & 29.42 & 0.234 & 843 & 240 & 29.70 & 0.232 & 620 & 310 \\
Taming 3DGS~\citep{tamings3dgs} & 25.91 & 0.129 & 641 & 245 & 22.50 & 0.208 & 269 & 308 & 29.78 & 0.234 & 812 & 214 & 30.57 & 0.234 & 577 & 251 \\
Mini-Splatting~\citep{fang2024minisplatting} & 25.00 & 0.163 & 54 & 1045 & 21.30 & 0.257 & 46 & 1084 & 29.38 & 0.257 & 94 & 1008 & 30.53 & 0.250 & 79 & 1129 \\
LiteGS (1M)~\citep{litegs} & 26.57 & \underline{0.113} & 246 & 483 & \textbf{23.14} & \underline{0.174} & 246 & 382 & 29.79 & 0.237 & 246 & 522 & \underline{30.88} & \underline{0.226} & 246 & 529 \\
\midrule
\multicolumn{17}{@{}l}{\emph{Automatic-budget controllers}} \\
DashGaussian~\citep{chen2025dashgaussian} & 25.90 & 0.149 & 354 & 318 & 22.41 & 0.207 & 248 & 266 & 29.20 & 0.249 & 590 & 238 & 30.13 & 0.243 & 375 & 303 \\
Compact 3DGS$^{\ast}$~\citep{lee2024compact} & 25.06 & 0.162 & 104 & 317 & 21.70 & 0.236 & 90 & 313 & 29.23 & 0.254 & 126 & 280 & 30.45 & 0.254 & 93 & 391 \\
LP-3DGS~\citep{zhang2024lp3dgs} & 25.37 & 0.152 & 198 & 537 & 21.95 & 0.218 & 106 & 621 & 29.07 & 0.249 & 202 & 563 & 30.14 & 0.246 & 139 & 717 \\
ControlGS~\citep{controlgs2025} & 25.32 & 0.141 & 179 & 186 & 22.59 & 0.191 & 227 & 175 & 29.26 & 0.258 & 116 & 226 & 29.93 & 0.250 & 115 & 229 \\
CDC-GS~\citep{dong2025cdcgs} & 26.22 & 0.118 & 512 & 271 & 22.48 & 0.180 & 274 & 264 & 29.47 & 0.233 & 577 & 190 & 29.90 & 0.236 & 329 & 275 \\
FastGS~\citep{ren2026fastgs} & 25.78 & 0.177 & 63 & 1006 & 22.56 & 0.239 & 57 & 875 & 29.52 & 0.272 & 63 & 1089 & 30.33 & 0.263 & 45 & 1163 \\
LeGS~\citep{ning2026legs} & \underline{26.58} & 0.128 & 472 & 449 & 22.78 & 0.182 & 209 & 491 & \textbf{29.96} & \underline{0.232} & 244 & 713 & \textbf{30.99} & \textbf{0.225} & 202 & 790 \\
\midrule
\multicolumn{17}{@{}l}{\emph{Coverage-scaled exploration + self-distillation (ours)}} \\
\textbf{\method{}} & \textbf{26.60} & 0.124 & 171 & 601 & \underline{23.05} & \textbf{0.162} & 265 & 365 & \underline{29.90} & \textbf{0.229} & 226 & 648 & 30.67 & 0.241 & 77 & 723 \\
\bottomrule
\end{tabular}
\par\vspace{2pt}{\scriptsize $^{\ast}$Compact~3DGS reports saved checkpoint MB above; estimated quantized payload: Truck: 42 MB, Train: 37 MB, Dr Johnson: 48 MB, Playroom: 38 MB.}
\end{table}

\begin{table}[!htbp]
\centering
\caption{\textbf{Per-capture saved model storage (MB) and native rendering throughput (FPS) on all eight large captures ($\mathbf{60}$k iterations).} Corresponding quality metrics and Gaussian counts appear in \cref{tab:supp-transfer-mill19,tab:supp-transfer-zipnerf_a}.}
\label{tab:supp-efficiency}\label{tab:supp-efficiency-aerial}\label{tab:supp-efficiency-handheld}
\scriptsize
\setlength{\tabcolsep}{2.3pt}
\begin{tabular}{@{}l rr rr rr rr rr rr rr rr@{}}
\toprule
 & \multicolumn{2}{c}{\textbf{Rubble}} & \multicolumn{2}{c}{\textbf{Building}} & \multicolumn{2}{c}{\textbf{Residence}} & \multicolumn{2}{c}{\textbf{Sci-Art}} & \multicolumn{2}{c}{\textbf{Alameda}} & \multicolumn{2}{c}{\textbf{London}} & \multicolumn{2}{c}{\textbf{NYC}} & \multicolumn{2}{c}{\textbf{Berlin}} \\
\cmidrule(lr){2-3}\cmidrule(lr){4-5}\cmidrule(lr){6-7}\cmidrule(lr){8-9}\cmidrule(lr){10-11}\cmidrule(lr){12-13}\cmidrule(lr){14-15}\cmidrule(lr){16-17}
Method & MB$\downarrow$ & FPS$\uparrow$ & MB$\downarrow$ & FPS$\uparrow$ & MB$\downarrow$ & FPS$\uparrow$ & MB$\downarrow$ & FPS$\uparrow$ & MB$\downarrow$ & FPS$\uparrow$ & MB$\downarrow$ & FPS$\uparrow$ & MB$\downarrow$ & FPS$\uparrow$ & MB$\downarrow$ & FPS$\uparrow$ \\
\midrule
\multicolumn{17}{@{}l}{\emph{Fixed-constant controllers}} \\
vanilla 3DGS~\citep{kerbl3Dgaussians} & 1488 & 152 & 1841 & 108 & 1320 & 131 & 640 & 195 & 215 & 283 & 155 & 362 & 263 & 260 & 190 & 237 \\
Speedy-Splat~\citep{hanson2025speedysplat} & 143 & 948 & \underline{48} & 1120 & 130 & 890 & \underline{77} & \underline{1016} & \textbf{16} & \underline{1341} & \textbf{14} & \underline{1427} & \textbf{21} & \underline{1374} & \textbf{21} & \underline{1184} \\
\midrule
\multicolumn{17}{@{}l}{\emph{Preset-count controllers}} \\
3DGS-MCMC~\citep{kheradmand2024mcmc} & 1488 & 241 & 1841 & 175 & 1320 & 257 & 640 & 236 & 215 & 851 & 155 & 1042 & 263 & 788 & 190 & 804 \\
Taming 3DGS ($15\times$)~\citep{tamings3dgs} & 716 & 294 & 524 & 246 & 727 & 232 & 507 & 249 & 108 & 387 & 74 & 468 & 92 & 429 & 104 & 323 \\
Mini-Splatting~\citep{fang2024minisplatting} & \textbf{56} & \textbf{1217} & \textbf{22} & \textbf{1465} & \textbf{37} & \textbf{1045} & \textbf{31} & \textbf{1166} & \underline{39} & 1077 & 74 & 908 & 54 & 980 & 65 & 734 \\
LiteGS (1M)~\citep{litegs} & 420 & 666 & 398 & 555 & 484 & 672 & 529 & 575 & 242 & 581 & 237 & 720 & 244 & 515 & 242 & 526 \\
\midrule
\multicolumn{17}{@{}l}{\emph{Automatic-budget controllers}} \\
DashGaussian~\citep{chen2025dashgaussian} & 1521 & 242 & 1070 & 269 & 1516 & 196 & 766 & 247 & 245 & 335 & 211 & 398 & 257 & 325 & 242 & 267 \\
Compact 3DGS$^{\ast}$~\citep{lee2024compact} & \underline{130} & 345 & 79 & 522 & \underline{119} & 307 & 87 & 450 & 63 & 683 & 60 & 813 & 61 & 747 & 61 & 559 \\
LP-3DGS~\citep{zhang2024lp3dgs} & 404 & 470 & 150 & 729 & 327 & 433 & 214 & 570 & 56 & 884 & 55 & 968 & 53 & 907 & 58 & 688 \\
ControlGS~\citep{controlgs2025} & 202 & 197 & 110 & 219 & 204 & 132 & 161 & 183 & 86 & 208 & 73 & 239 & 70 & 230 & 70 & 136 \\
CDC-GS~\citep{dong2025cdcgs} & 1464 & 152 & 1450 & 140 & 1689 & 117 & 785 & 198 & 227 & 309 & 198 & 361 & 242 & 279 & 234 & 248 \\
FastGS~\citep{ren2026fastgs} & 182 & \underline{1037} & 121 & \underline{1179} & 194 & \underline{1010} & 150 & 996 & 42 & \textbf{1355} & \underline{30} & \textbf{1483} & \underline{30} & \textbf{1430} & \underline{34} & \textbf{1234} \\
LeGS~\citep{ning2026legs} & 374 & 765 & 228 & 953 & 444 & 638 & 367 & 736 & 124 & 981 & 144 & 993 & 93 & 1028 & 91 & 957 \\
\midrule
\multicolumn{17}{@{}l}{\emph{Coverage-scaled exploration + self-distillation (ours)}} \\
\textbf{\method{}} & 1206 & 353 & 1505 & 364 & 1194 & 357 & 1287 & 267 & 942 & 331 & 697 & 485 & 523 & 383 & 394 & 386 \\
\bottomrule
\end{tabular}
\par\vspace{2pt}{\scriptsize $^{\ast}$Compact~3DGS reports saved checkpoint MB above; estimated quantized payload: Rubble: 49 MB, Building: 34 MB, Residence: 46 MB, Sci-Art: 36 MB, Alameda: 29 MB, London: 28 MB, NYC: 28 MB, Berlin: 29 MB.}
\end{table}

\clearpage
\section{Post-Growth Quality--Size Trade-Offs}
\label{sec:supp-capacity-objective}

Because \method{} separates the $80\%$ growth phase from final selection, changing the quality--size trade-off does not require repeating densification or retuning growth thresholds.
Starting from a single $80\%$ teacher checkpoint $\mathcal{G}_{\mathrm{pre}}$ and its utility scores $\{\util\}$, varying $\lambda \in \{8, 4, 2, 1, 0.5\}\lamstar$ applies one threshold step followed by the final $20\%$ refinement phase.
For $\lambda \ge 2\lamstar$, scenes with large teacher headroom, mainly smooth indoor captures and \emph{Sci-Art} at $8\lamstar$, reach the $0.10\npre$ safety floor of \cref{sec:supp-impl}, which sets their minimum retained count.
\Cref{tab:supp-capacity-pareto,tab:supp-capacity-pareto-perscene} report aggregate and per-scene results across all 21 scenes ($105$ models in total).
On Large-8 without retuning, every operating point from $8\lamstar$ ($1.28$M, $24.29$\,dB, $555$\,FPS) to $0.5\lamstar$ ($4.92$M, $24.75$\,dB, $322$\,FPS) outperforms all 13 baselines in both mean PSNR and LPIPS.

\begin{table}[!htbp]
\centering
\caption{\textbf{Aggregate post-growth quality--size trade-offs across capture scales (seed 4114).} All five operating points reuse the single $80\%$ teacher checkpoint and score vector $\{\util\}$ per capture, requiring only threshold selection and the final $20\%$ refinement without repeating growth.}
\label{tab:supp-capacity-pareto}
\footnotesize
\setlength{\tabcolsep}{4.5pt}
\begin{tabular}{@{}l rrrrr rrrrr rr@{}}
\toprule
 & \multicolumn{5}{c}{\textbf{Standard Scenes (Std-13)}} & \multicolumn{5}{c}{\textbf{Large Captures Without Retuning (Large-8)}} & \multicolumn{2}{c}{\textbf{All-21}} \\
\cmidrule(lr){2-6}\cmidrule(lr){7-11}\cmidrule(lr){12-13}
Threshold & $\numg$(M) & PSNR$\uparrow$ & SSIM$\uparrow$ & LPIPS$\downarrow$ & FPS$\uparrow$ & $\numg$(M) & PSNR$\uparrow$ & SSIM$\uparrow$ & LPIPS$\downarrow$ & FPS$\uparrow$ & $\numg$(M) & PSNR$\uparrow$ \\
\midrule
$8\lamstar$ & 0.42 & 27.52 & 0.832 & 0.234 & 764 & 1.28 & 24.29 & 0.790 & 0.310 & 555 & 0.75 & 26.29 \\
$4\lamstar$ & 0.56 & 27.72 & 0.842 & 0.216 & 699 & 1.89 & 24.44 & 0.804 & 0.289 & 507 & 1.07 & 26.47 \\
$2\lamstar$ & 0.73 & 27.91 & 0.847 & 0.204 & 633 & 2.83 & 24.62 & 0.815 & 0.270 & 431 & 1.53 & 26.66 \\
$\lamstar$ (default) & 0.93 & \underline{28.03} & \underline{0.850} & \underline{0.195} & 580 & 3.90 & \underline{24.70} & \underline{0.821} & \underline{0.258} & 366 & 2.06 & \underline{26.76} \\
$0.5\lamstar$ & 1.12 & \textbf{28.07} & \textbf{0.852} & \textbf{0.190} & 546 & 4.92 & \textbf{24.75} & \textbf{0.824} & \textbf{0.253} & 322 & 2.57 & \textbf{26.81} \\
\bottomrule
\end{tabular}
\end{table}

\begin{table}[!htbp]
\centering
\caption{\textbf{Per-scene post-growth quality--size trade-offs across all 21 scenes (seed 4114).} Retained count $\numg$ (M), test PSNR (dB), and LPIPS obtained from each scene's shared $80\%$ teacher checkpoint across $\lambda \in \{8, 4, 2, 1, 0.5\}\lamstar$.}
\label{tab:supp-capacity-pareto-perscene}
\scriptsize
\setlength{\tabcolsep}{2.2pt}
\begin{tabular}{@{}l rrr rrr rrr rrr rrr@{}}
\toprule
 & \multicolumn{3}{c}{\textbf{$8\lamstar$}} & \multicolumn{3}{c}{\textbf{$4\lamstar$}} & \multicolumn{3}{c}{\textbf{$2\lamstar$}} & \multicolumn{3}{c}{\textbf{$\lamstar$ (default)}} & \multicolumn{3}{c}{\textbf{$0.5\lamstar$}} \\
\cmidrule(lr){2-4}\cmidrule(lr){5-7}\cmidrule(lr){8-10}\cmidrule(lr){11-13}\cmidrule(lr){14-16}
Capture & $\numg$ & PSNR & LPIPS & $\numg$ & PSNR & LPIPS & $\numg$ & PSNR & LPIPS & $\numg$ & PSNR & LPIPS & $\numg$ & PSNR & LPIPS \\
\midrule
Bicycle & 0.66 & 25.44 & 0.256 & 0.97 & 25.68 & 0.218 & 1.23 & \underline{25.74} & 0.203 & 1.42 & \textbf{25.76} & \underline{0.198} & 1.54 & \textbf{25.76} & \textbf{0.196} \\
Bonsai & 0.22 & 31.80 & 0.225 & 0.22 & 31.80 & 0.225 & 0.36 & 32.59 & 0.202 & 0.64 & \underline{33.07} & \underline{0.187} & 0.98 & \textbf{33.24} & \textbf{0.181} \\
Counter & 0.22 & 28.93 & 0.225 & 0.22 & 28.97 & 0.223 & 0.41 & 29.48 & 0.193 & 0.65 & \underline{29.68} & \underline{0.179} & 0.91 & \textbf{29.74} & \textbf{0.173} \\
Flowers & 0.94 & \textbf{21.97} & 0.302 & 1.18 & \textbf{21.97} & 0.292 & 1.34 & \underline{21.96} & 0.288 & 1.45 & 21.94 & \underline{0.285} & 1.53 & 21.94 & \textbf{0.284} \\
Garden & 0.52 & 27.40 & 0.166 & 0.71 & 27.60 & 0.142 & 0.84 & 27.67 & 0.134 & 0.92 & \underline{27.70} & \underline{0.131} & 0.97 & \textbf{27.71} & \textbf{0.130} \\
Kitchen & 0.14 & 30.42 & 0.181 & 0.30 & 31.87 & 0.137 & 0.55 & 32.39 & 0.122 & 0.81 & \underline{32.56} & \underline{0.117} & 1.01 & \textbf{32.62} & \textbf{0.115} \\
Room & 0.20 & 32.06 & 0.245 & 0.20 & 32.05 & 0.245 & 0.22 & 32.09 & 0.239 & 0.39 & \underline{32.55} & \underline{0.208} & 0.63 & \textbf{32.70} & \textbf{0.194} \\
Stump & 0.51 & 27.30 & 0.230 & 0.80 & \underline{27.50} & 0.198 & 1.05 & \textbf{27.53} & 0.187 & 1.22 & \textbf{27.53} & \underline{0.184} & 1.34 & \textbf{27.53} & \textbf{0.182} \\
Treehill & 0.87 & \textbf{23.29} & 0.323 & 1.16 & \textbf{23.29} & 0.301 & 1.38 & \underline{23.27} & 0.293 & 1.52 & 23.26 & \underline{0.290} & 1.62 & 23.24 & \textbf{0.289} \\
Truck & 0.20 & 26.02 & 0.174 & 0.35 & 26.44 & 0.143 & 0.51 & \underline{26.58} & 0.130 & 0.69 & \textbf{26.62} & \underline{0.124} & 0.85 & \textbf{26.62} & \textbf{0.122} \\
Train & 0.34 & 22.84 & 0.215 & 0.59 & 22.98 & 0.183 & 0.85 & \textbf{23.09} & 0.168 & 1.07 & 23.05 & \underline{0.161} & 1.22 & \underline{23.06} & \textbf{0.159} \\
Dr Johnson & 0.37 & 29.65 & 0.253 & 0.37 & 29.67 & 0.253 & 0.50 & 29.79 & 0.243 & 0.91 & \underline{29.93} & \underline{0.230} & 1.40 & \textbf{29.99} & \textbf{0.223} \\
Playroom & 0.26 & 30.62 & 0.247 & 0.26 & 30.60 & 0.247 & 0.26 & 30.61 & 0.247 & 0.32 & \underline{30.71} & \underline{0.241} & 0.56 & \textbf{30.82} & \textbf{0.229} \\
\midrule
Rubble & 1.95 & 24.86 & 0.329 & 3.14 & 25.19 & 0.290 & 4.15 & 25.32 & 0.275 & 4.86 & \underline{25.37} & \underline{0.270} & 5.32 & \textbf{25.41} & \textbf{0.267} \\
Building & 1.92 & 21.11 & 0.317 & 3.34 & 21.37 & 0.277 & 4.86 & \underline{21.58} & 0.261 & 6.08 & 21.56 & \underline{0.256} & 6.87 & \textbf{21.59} & \textbf{0.253} \\
Residence & 1.65 & 21.92 & 0.265 & 2.70 & 22.10 & 0.235 & 3.81 & 22.17 & 0.221 & 4.77 & \underline{22.21} & \underline{0.216} & 5.51 & \textbf{22.22} & \textbf{0.214} \\
Sci-Art & 1.12 & 21.89 & 0.265 & 1.88 & 22.04 & 0.227 & 3.38 & \underline{22.17} & 0.195 & 5.18 & 22.16 & \underline{0.180} & 6.87 & \textbf{22.18} & \textbf{0.175} \\
Alameda & 0.95 & 22.76 & 0.374 & 1.37 & 22.96 & 0.357 & 2.44 & 23.09 & 0.337 & 3.82 & \underline{23.13} & \underline{0.325} & 5.27 & \textbf{23.15} & \textbf{0.320} \\
Berlin & 0.82 & 27.79 & 0.273 & 0.82 & 27.80 & 0.272 & 0.96 & 27.88 & 0.267 & 1.59 & \underline{28.11} & \underline{0.250} & 2.39 & \textbf{28.22} & \textbf{0.240} \\
London & 1.12 & 26.49 & 0.342 & 1.12 & 26.50 & 0.342 & 1.75 & 26.71 & 0.319 & 2.84 & \underline{26.83} & \underline{0.300} & 4.12 & \textbf{26.88} & \textbf{0.290} \\
NYC & 0.70 & 27.53 & 0.313 & 0.72 & 27.56 & 0.312 & 1.30 & 28.08 & 0.285 & 2.09 & \underline{28.26} & \underline{0.270} & 2.99 & \textbf{28.32} & \textbf{0.263} \\
\bottomrule
\end{tabular}
\end{table}

\FloatBarrier

\end{document}